\documentclass[a4paper,12pt,numbered,print]{uathesis}
\usepackage[english,ngerman]{babel} 
\addto\captionsngerman{}
\usepackage{graphicx}
\usepackage{algorithm}
\usepackage{algorithmic}
\usepackage{cmap}
\usepackage{wrapfig}
\usepackage{amssymb}
\usepackage{moreverb}
\usepackage{setspace}
\usepackage{booktabs}
\usepackage{url}
\usepackage{commath}
\usepackage{amssymb}
\usepackage{amsmath}
\usepackage{adjustbox}
\usepackage{tabularx}
\usepackage{tabularray}
\usepackage{multicol}
\usepackage{listings}
\usepackage{subcaption}
\usepackage{tabularx}
\usepackage[inline]{enumitem}
\usepackage[table,dvipsnames]{xcolor}
\usepackage{cleveref}
\usepackage{rotating}
\usepackage{makecell}
\usepackage[acronym]{glossaries}
\usepackage{glossary-mcols}
\usepackage{multirow}
\usepackage{fancyhdr}
\usepackage{rotating}
\usepackage{pdfpages}
\usepackage[Lenny]{fncychap}
\ChNameVar{\fontsize{18}{20}\usefont{OT1}{ptm}{m}{n}\selectfont}
\ChTitleVar{\huge\bfseries\fontfamily{ptm}\selectfont}

\usepackage{titlesec}
\titleformat{\section}{\large\scshape\bfseries}{\thesection}{1em}{}
\titleformat{\subsection}{\normalsize\scshape\bfseries}{\thesubsection}{1em}{}
\titleformat{\subsubsection}{\normalsize\scshape\bfseries}{}{0.5em}{}

\newcolumntype{Y}{>{\centering\arraybackslash}X}

\definecolor{light-gray}{gray}{0.95}

\newlist{questions}{enumerate}{2}
\setlist[questions,1]{label=\textit{RQ\arabic*.},ref=\textit{RQ\arabic*}}
\setlist[questions,2]{label=(\alph*),ref=\thequestionsi(\alph*)}

\makeglossaries
\newacronym{ais}{AIS}{Automatic Identification System}
\newacronym{dlr}{DLR}{German Aerospace Center}
\newacronym[\glslongpluralkey={Convolutional Neural Networks},
\glsshortpluralkey={CNNs}]{cnn}{CNN}{Convolutional Neural Network}
\newacronym[\glslongpluralkey={Generative Adversarial Networks},
\glsshortpluralkey={GANs}]{gan}{GAN}{Generative Adversarial Network}
\newacronym{ai}{AI}{Artificial Intelligence}
\newacronym{map}{mAP}{mean Average Precision}
\newacronym{ap}{AP}{Average Precision}
\newacronym[\glslongpluralkey={Graphics Processing Units},
\glsshortpluralkey={GPUs}]{gpu}{GPU}{Graphics Processing Unit}
\newacronym{smd}{SMD}{Singapore Maritime Dataset}
\newacronym{arm}{ARM}{Advanced Reduced instruction set computer Machine}
\newacronym{dlt}{DLT}{Direct Linear Transformation}
\newacronym{ransac}{RANSAC}{Random Sample Consensus}
\newacronym{ls}{LS}{Least Squares}
\newacronym{gis}{GIS}{Geographic Information System}
\newacronym{utm}{UTM}{Universal Transverse Mercator}
\newacronym{gde}{GDE}{Georeferencing Distance Error}
\newacronym{h}{H}{Homography}
\newacronym{relu}{ReLU}{Rectified Linear Unit}
\newacronym{cbam}{CBAM}{Convolutional Block Attention Module}
\newacronym{iou}{IoU}{Intersection over Union}
\newacronym{dtcwt}{DTCWT}{Dual-Tree Complex Wavelet Transform}
\newacronym{cuda}{CUDA}{Compute Unified Device Architecture}
\newacronym{sahi}{SAHI}{Slicing Aided Hyper Inference}
\newacronym{nms}{NMS}{Non-Maximum Suppression}
\newacronym{onnx}{ONNX}{Open Neural Network Exchange}
\newacronym{vts}{VTS}{Vessel Traffic Services}
\newacronym{imo}{IMO}{International Maritime Organization}
\newacronym{sgd}{SGD}{Stochastic Gradient Descent}
\newacronym{nlp}{NLP}{Natural Language Processing}
\newacronym{mlp}{MLP}{Multi-layer Perceptron}
\newacronym{csp}{CSP}{Cross Stage Partial}
\newacronym{silu}{SiLU}{Sigmoid-Weighted Linear Unit}

\begin{document}
\title{Real-time Ship Recognition and Georeferencing for the Improvement of Maritime Situational Awareness}
\author{Borja Carrillo Perez}
\subjectarea{\footnotesize Computer Science}
\crest{\includegraphics[width=0.5\textwidth]{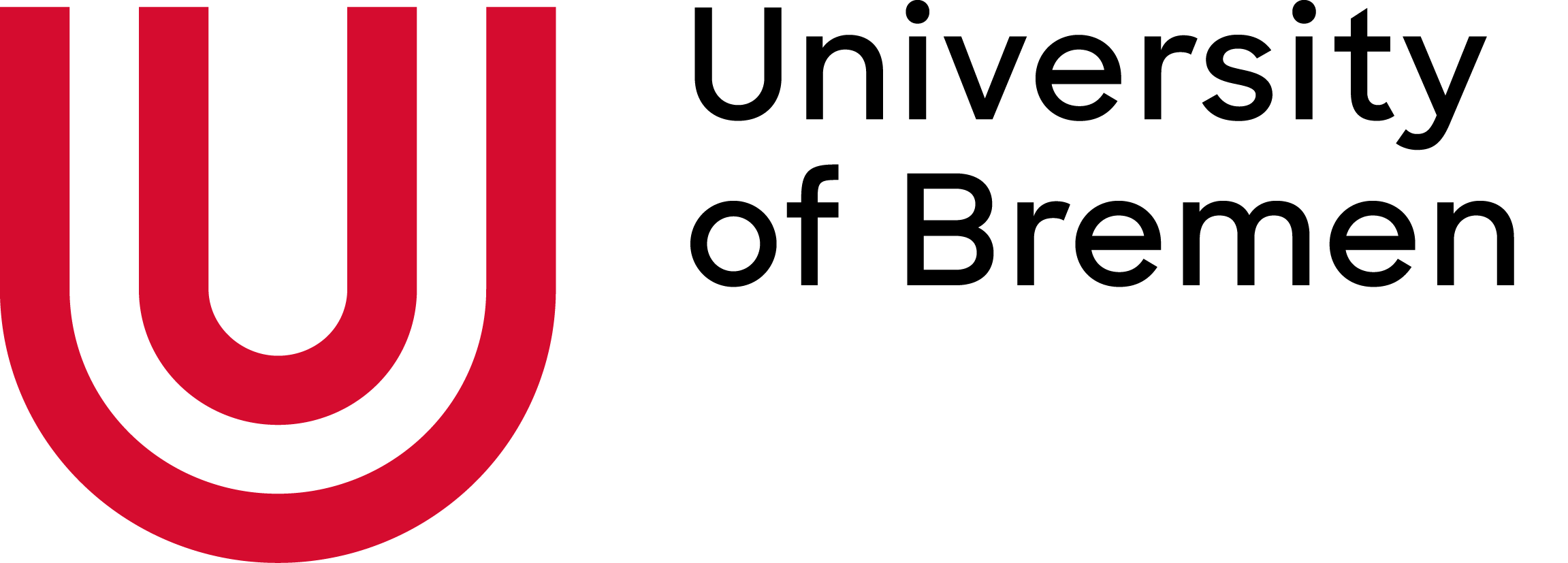}}
\degreetitle{\footnotesize Doctor}
\degreedate{2024} 

\maketitle

\thispagestyle{empty}
\cleardoublepage
\thispagestyle{empty}
This doctoral thesis is a compilation of six articles that I authored and co-authored between 2021 and 2024. The research was conducted at the Institute for the Protection of Maritime Infrastructures of the German Aerospace Center (DLR), in Bremerhaven, Germany. 
\vspace*{\fill} 

\begin{center} 
\textbf{Date of the Defense:}  09.08.2024

\vspace{10mm} 
\textbf{Examiners:}

\vspace{4mm} 
Prof. Dr.-Ing. Udo Frese (University of Bremen)

\vspace{2mm} 
Prof. Dr.-Ing. Tobias Meisen (University of Wuppertal)

\vspace{10mm} 
\textbf{Further members of the examination committee:}

\vspace{4mm} 
Prof. Dr. Ralf Bachmayer (University of Bremen)

\vspace{2mm} 
Dr. rer. nat. Enno Peters (German Aerospace Center)

\vspace{2mm} 
Arne Hasselbring (University of Bremen)

\vspace{2mm} 
Ayleen Lührsen (University of Bremen)
\vspace*{\fill} 
\end{center}
\cleardoublepage

\thispagestyle{empty}
\cleardoublepage
\thispagestyle{empty}
\textbf{Declaration}
\\
\par
I, Borja Carrillo Perez, declare that this thesis, titled ``Real-time Ship Recognition and Georeferencing for the Improvement of Maritime Situational Awareness", is my own work and has not been submitted for any other degree or qualification at this or any other institution. All sources are acknowledged and referenced.\\
\par

\underline{Borja Carrillo Perez}\\\par
June 2024\par

\cleardoublepage
\selectlanguage{english}
\begin{abstract}

In an era where maritime infrastructures are paramount supports of societies, the need for advanced maritime situational awareness solutions has become increasingly important.
Existing ship monitoring procedures, such as the~\gls{ais}, have limitations, suffer from delayed updates and are vulnerable to cyberattacks. 
Other technologies, such as satellite imagery and radar, face challenges in real-time applications due to delays in acquiring and processing data. The use of optical camera systems and image processing  can improve situational awareness, allowing real-time usage of maritime infrastructure footage. However, the number of video streams available poses a challenge for maritime operators, who could be helped by summarized spatial information of recognized ships, irrespective of their size and type, and presented on a map in real-time. This motivates the development of automated ship recognition and georeferencing technologies. Moreover, the deployment of such camera systems, equipped with an embedded device, allows for local data processing on the edge to minimize network demand, energy usage, decrease latency, cut costs, and enhance data protection.

This thesis, integrating six of my publications, presents a comprehensive investigation into leveraging deep learning and computer vision to advance the research in real-time ship recognition and georeferencing for the improvement of maritime situational awareness. I present a novel dataset for ship recognition and georeferencing, ShipSG, which facilitates the development and validation of recognition and georeferencing methodologies. The dataset contains 3505 images and 11625 ship masks with their corresponding class, geographic position and length. Through a series of studies of state-of-the-art deep-learning-based object recognition algorithms, I introduce a custom real-time segmentation architecture, ScatYOLOv8+CBAM. This architecture was created and optimized for the NVIDIA Jetson AGX Xavier as embedded system. ScatYOLOv8+CBAM incorporates the 2D scattering transform, a novel addition that enhances YOLOv8 in real-world applications such as ship segmentation. Additionally, the performance is further improved with the integration of attention mechanisms. 
The proposed architecture exceeds in more than 5\% the performance of state-of-the-art methods, achieving a \gls{map} of 75.46\%. The inference speed, once the customized architecture is deployed on the embedded system using TensorRT, is of 25.3 ms per frame. 
Furthermore, I address the need for precision in recognizing small and distant ships and their real-time processing of full-resolution images on embedded systems, with an enhanced slicing mechanism that performs batch inference and merges predictions, achieving \gls{map} improvements ranging from 8\% to 11\%.
The recognized ships are georeferenced using my proposed method, which automatically calculates the georeferencing pixel of the recognized ships, and uses homographies to provide the geographic position of ships from single images, without prior camera knowledge. 
In the quantitative analysis, the georeferencing method achieved a positioning error of $18~m~\pm~10~m$ for ranges inside the port basin (up to 400 $m$) and $44~m~\pm~27~m$ outside (from 400 $m$ to 1200 $m$).
The main findings reveal significant advancements in maritime situational awareness with the practical demonstration of the applicability of the methodologies in real-world scenarios, such as the detection of abnormal ship behaviour, camera integrity assessment and 3D reconstruction. The approach not only outperforms existing methods in terms of accuracy and processing speed but also provides a framework for seamlessly integrating recognized and georeferenced ships into real-time systems, enhancing operational effectiveness and decision-making for maritime authorities. The integration of these methodologies into embedded systems represents a pivotal advancement in the domain, offering a scalable and efficient solution for improving maritime situational awareness and response capabilities.
This thesis contributes to the maritime computer vision field by establishing a benchmark for ship segmentation and georeferencing research, demonstrating the viability of deep-learning-based recognition and georeferencing methods for real-time maritime monitoring.        

\end{abstract}

\selectlanguage{ngerman}
\glsresetall
\begin{abstract}

In einer Ära, in der maritime Infrastrukturen von größter Bedeutung für Gesellschaften sind, ist der Bedarf an fortschrittlichen Lösungen zur maritimen Lageerkennung zunehmend wichtiger geworden. 
Bestehende Verfahrung zur Schiffsbeobachtung wie das~\gls{ais} haben Einschränkungen, leiden unter verzögerten Aktualisierungen und sind anfällig für Cyberangriffe. 
Andere Technologien, wie Satellitenbilder und Radar, haben Schwierigkeiten bei Echtzeitanwendungen aufgrund von Verzögerungen bei der Erfassung und Verarbeitung von Daten. Der Einsatz von optischen Kamerasystemen und Computer Vision kann das Situationsbewusstsein verbessern, indem sie die Echtzeitnutzung von Aufnahmen direkt an marititmen Infrastrukturen ermöglichen. Die große Anzahl verfügbarer Videostreams stellt jedoch eine Herausforderung für maritime Betreiber dar, die durch zusammengefasste, räumliche Informationen über erkannte Schiffe, unabhängig von ihrer Größe und Art, auf einer Karte in Echtzeit unterstützt werden könnten. Dies motiviert die Entwicklung automatisierter Technologien zur Schiffsidentifikation und Georeferenzierung. Darüber hinaus ermöglicht der Einsatz von Kamerasystemen zusammen mit Embedded Systems, die lokale Datenverarbeitung in situ, um den Netzwerkbedarf zu minimieren, den Energieverbrauch zu senken, die Latenz zu verringern, die Kosten zu senken und den Datenschutz zu verbessern.

Diese Dissertation, die sechs meiner Veröffentlichungen bündelt, präsentiert eine umfassende Untersuchung zur Nutzung von Deep Learning und Computer Vision, um die Forschung zur Echtzeit-Schiffsidentifikation und Georeferenzierung zur Verbesserung des maritimen Situationsbewusstseins voranzutreiben. Ich präsentiere einen neuartigen Datensatz für die Schiffsidentifikation und Georeferenzierung, ShipSG, der die Entwicklung und Validierung von Identifikations- und Georeferenzierungsmethoden erleichtert. Der Datensatz enthält 3505 Bilder und 11625 Schiffsmasken mit entsprechenden Klassen, geografischer Position und Länge. Durch eine Reihe von Studien zu den neuesten Objekterkennungsalgorithmen basierend auf Deep-Learning stelle ich eine neuartige Echtzeit-Segmentierungsarchitektur vor, ScatYOLOv8+CBAM. Diese Architektur wurde speziell für den NVIDIA Jetson AGX Xavier als eingebettetes System entwickelt und optimiert. ScatYOLOv8+CBAM integriert die 2D-Streutransformation, eine neuartige Ergänzung, die YOLOv8 in realen Anwendungen wie der Schiffsegmentierung verbessert. Zudem wird die Leistung durch die Integration von Attention-Mechanismen weiter gesteigert.
Die vorgeschlagene Architektur übertrifft neueste Methoden um mehr als 5\% und erreicht eine Mean-Average-Precision (mAP) von 75.46\%. Die Inferenzlaufzeit beträgt 25.3 ms pro Frame, sobald die angepasste Architektur auf dem eingebetteten System mit TensorRT bereitgestellt ist.
Darüber hinaus gehe ich auf die Notwendigkeit der Präzision bei der Erkennung kleiner und entfernter Schiffe und ihrer Echtzeitverarbeitung von Bildern in voller Auflösung auf eingebetteten Systemen ein. Hierzu betrachte ich einen verbesserten Slicing-Mechanismus, der Batch-Inferenz durchführt und Vorhersagen zusammenführt, was letztlich in \gls{map}-Verbesserungen von 8\% bis 11\% resultiert.
Die erkannten Schiffe werden mittels meiner vorgeschlagenen Methode georeferenziert.  Dazu wird automatisch das Georeferenzierungspixel der erkannten Schiffe berechnet und Homographien verwendet, um die geografische Position der Schiffe aus Einzelbildern ohne vorherige Kamerakenntnisse zu bestimmen.
In der quantitativen Analyse erreichte die Georeferenzierungsmethode einen Positionierungsfehler von $18~m~\pm~10~m$ für Entfernungen innerhalb des betrachtetetn Hafenbeckens (bis zu 400 $m$) und $44~m~\pm~27~m$ außerhalb (von 400 $m$ bis 1200 $m$).
Die Hauptresultate zeigen erhebliche Fortschritte im maritimen Situationsbewusstsein, welche anhand von praktischen Beispielen der Anwendbarkeit der Methoden in realen Szenarien demonstriert werden. Zu diesen Beispielen gehören die Erkennung von abnormalem Schiffsverhalten, die Bewertung der Kameraintegrität als auch die 3D-Rekonstruktion von Schiffen. Der Ansatz übertrifft nicht nur bestehende Methoden in Bezug auf Genauigkeit und Laufzeit, sondern bietet auch ein Framework für die nahtlose Integration erkannter und georeferenzierter Schiffe in Echtzeitsysteme, wodurch die operative Effizienz und Entscheidungsfindung für maritime Behörden verbessert werden kann. Die Integration dieser Methoden in eingebettete Systeme stellt einen entscheidenden Fortschritt in diesem Anwendungsbereich dar und bietet eine skalierbare und effiziente Lösung zur Verbesserung des maritimen Situationsbewusstseins und der Reaktionsfähigkeiten.
Diese Dissertation trägt zum Bereich der maritimen Computer Vision bei, indem sie eine Benchmark für die Forschung zur Schiffssegmentierung und Georeferenzierung etabliert und die Machbarkeit von auf Deep-Learning basierenden Erkennungs- und Georeferenzierungsmethoden für die Echtzeitüberwachung maritimer Umgebungen demonstriert.
\end{abstract}
\selectlanguage{english}

\begin{acknowledgements}
This academic and personal achievement would not have been possible without the support I have received over these past years. I am profoundly grateful to everyone who has contributed.

Firstly, I extend my deepest thanks to Dr. Sarah Barnes and Dr. Maurice Stephan from the German Aerospace Center for their invaluable guidance, insightful discussions, commitment, and time. I feel very lucky to have had the opportunity to learn from them, and their mentorship has profoundly shaped my professional and personal growth.

I am also grateful to Prof. Dr.-Ing. Udo Frese from the University of Bremen for his guidance and advice, especially during the final stages of this work, which significantly elevated the quality of my research. I appreciate Prof. Dr.-Ing. Tobias Meisen from the University of Wuppertal for his time and willingness to review this thesis, and I thank the committee members for their participation in the defense.

I would like to thank Jens-Michael Schlüter and Michael Busack from the Alfred Wegener Institute for Polar and Marine Research, as well as Marco Gawehns and Tino Flenker from the German Aerospace Center, for their technical support in the data acquisition process. Additionally, I am grateful to Dr. Michael Stadermann and Dr. Frank Sill Torres for their support in the publication of the data.

My sincere thanks to my colleagues at the Institute for the Protection of Maritime Infrastructures of the German Aerospace Center for their incredible support, scientific contributions, and the joy they brought to our work environment. Special thanks go to Dr. Angel Bueno, Dr. Edgardo Solano, Felix Sattler and the Methods and Processing group members. I am also grateful to the Machine Learning group of MI and PI for their valuable input and discussions. And big thanks to those who read the articles and thesis and provided such helpful feedback.

I am grateful to the friends I made in Bremen along these years for their encouragement and support. They have always found kind words that made the journey lighter, even when it felt heavier.

Me gustaría agradecer a mis amigos de España por su constante apoyo. 

Muchas gracias a Paloma, por haber sido mi apoyo durante años, por enseñarme tanto y por inspirarme querer alcanzar este objetivo. 

Y por último a mi familia, por su amor, por hacerme quien soy y empujarme a luchar por superarme.


\end{acknowledgements}

\setcounter{tocdepth}{2}
\frontmatter
\tableofcontents
\listoffigures
\addtocontents{lof}{\vspace{-4em}}
\listoftables
\addtocontents{lot}{\vspace{-4em}}
\printglossary[type=\acronymtype, style=mcolindex, nonumberlist]

\mainmatter
\chapter{Introduction}
\label{chap:intro}
\glsresetall

Maritime infrastructures are an essential component towards the support of societal needs, economic activities, mobility, and the advancement of renewable energy sources~\cite{engler2018resiliencen}. This highlights the reason why their security, integrity, and operational safety are crucial. In response, maritime research is aiming at developing, testing, and validating systems to thoroughly assess and operate these infrastructures~\cite{torres2020indicator}. 
Such initiatives aim to cultivate a proactive and informed understanding of maritime contexts, essential for accurately determining the protection status of infrastructures in real-time and enabling prompt action against various threats, including major accidents, natural disasters, and organized crime~\cite{engler2018resiliencen}. Maritime situational awareness, facilitated by advanced technologies and data integration, is critical for a proactive and informed understanding of maritime environments~\cite{cetin2013increasing}. It encompasses real-time monitoring and drives innovative solutions to enhance the security, safety, structural integrity, and operational reliability of infrastructures against various threats~\cite{ventikos2022risk}.

In the improvement of maritime situational awareness the introduction of advanced instruments and sensors plays a key role, which should be designed not only to recognize elements of interest but also to suggest practical measures to both users, for operational decisions, and authorities, for regulatory compliance and emergency response~\cite{wang2019maritime}. Enhancing maritime situational awareness with technology represents a significant advancement for smart ports, exemplifying the potential for improved maritime operations~\cite{belmoukari2023smart}.

The \gls{imo} mandates that vessels exceeding 300 gross tonnage are equipped with \gls{ais} transceivers, which broadcast crucial data including identification numbers, type, position, course, and speed through encoded radio messages~\cite{imo2015resolutionA110629}. This system is pivotal for \gls{vts} and nearby vessels, facilitating marine traffic awareness, critical operations such as collision avoidance, and search and rescue missions. \gls{ais} transmissions occur at intervals ranging from 2 to 10 seconds when ships are underway, which can be extended by up to 6 minutes when stationary\footnote{\url{https://www.navcen.uscg.gov/ais-messages}}. Such intervals may leave gaps in real-time monitoring, highlighting the need for systems capable of analyzing situations with a significantly shorter interval to aid in the prevention and response to potential maritime complications~\cite{kim2019adaptive}. Furthermore, the open standards employed by the \gls{ais} exposes it to various cyber threats, including spoofing, hijacking attacks, and denial of service, underscoring the vulnerability of the system~\cite{jakovlev2020analysis, struck2021backwards, wimpenny2018public, alincourt2016methodology, balduzzi2014security}. Therefore, despite significant efforts, real-time ship monitoring for improved maritime situational awareness only using \gls{ais} continues to pose a challenge for \gls{vts}~\cite{yan2020exploring}. 

Other available sources for the improvement of maritime situational awareness are satellite imagery and radar systems~\cite{reggiannini2024remote}. However, their processing for real-time maritime situational awareness faces challenges due to the time-sensitive nature of data acquisition, periodic satellite overpasses, and processing delays ($\sim$15 minutes per data cycle), impacting the immediacy and utility of the information~\cite{schwarz2015near}. Moreover, revisit times of satellites can range from hours to days.

Optical camera systems, on the other hand, due to their accessibility, cost-efficiency, and ease of deployment, are key in rapidly assessing ship traffic, enhancing maritime situational awareness through views of the infrastructure from strategic positions \cite{prasad2017video}. The vast number of video streams available can present a challenge for operators~\cite{li2020causal}.
The efficiency of real-time recognition is significantly boosted by image processing technologies applied to optical monitoring~\cite{prasad2017video}.
This motivates the use of computer vision and deep learning to automatically recognize and locate geographically (georeference) ships on a map, irrespective of their type or size. This process can support the operational decision-making procedures of maritime authorities by providing them with spatial information in a timely manner~\cite{flenker2021marlin}. 

Early detection of potential threats and the prevention of accidents are significantly enhanced by employing optical cameras when used at full-resolution, ensuring detailed and precise imagery for monitoring purposes~\cite{chen2020deep, rekavandi2022guide}. Therefore, a key challenge in maritime monitoring is the recognition of small and distant ships, which is crucial for safety and security at maritime infrastructures, as it helps in early threat detection and accident prevention~\cite{chen2020deep}.

\begin{figure}[h]
\includegraphics[width=14 cm]{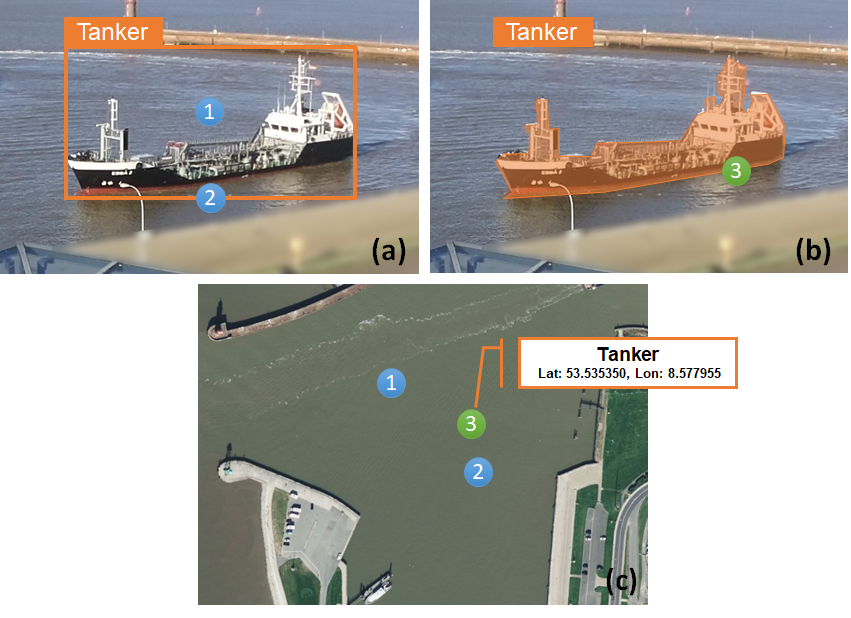}
\centering
\caption[Conceptual representation of ship detection, segmentation and georeferencing from maritime footage.]{Conceptual representation of ship detection, segmentation and georeferencing from maritime footage. (a) Tanker being detected. Point 1 represents the bounding box center, and 2, the bottom-center point. (b) Tanker being segmented. Point 3 represents the intersection of the navigation antenna with the water. (c) Representation of the goereferenced tanker displayed on a map. The georeference from the mask, 3, provides the most accurate ship location of the three points.}\label{fig:motivation_georef_mask}
\end{figure}

Object georeferencing involves linking physical objects to specific locations on the Earth's surface for spatial integration and analysis, a process that can extend to objects captured within an image~\cite{Hastings2009}.
For the improvement of maritime situational awareness, once ships are recognized in real-time from monitoring images, the display of their geographical positions on maps using georeferencing enables a better spatial understanding of the situation~\cite{flenker2021marlin}. 

Ship detection methods which use monitoring footage to present a bounding box surrounding the detected ship,  can be used to improve maritime situational awareness~\cite{wawrzyniak2019vessel, helgesen2020low}. However, ship segmentation provides a more accurate georeference for the ships using the segmented masks, as shown in Figure~\ref{fig:motivation_georef_mask}. The georeferenced pixel can be better inferred from the segmented mask of an object than from a surrounding bounding box, which usually contains unnecessary background. The center and bottom-center of the bounding box, given the perspective of the image, provide a more erroneous georeference compared to the point that lies at the intersection point between the ship hull and the water below the bridge or wheelhouse, where the navigation antenna is located. This rationale motivates the exploration of ship segmentation methods beyond bounding box detection, and paves the way for the development of a method to automatically identify the pixel for georeferencing within this thesis.

Furthermore, the processing using embedded devices powered with a \gls{gpu}, equipped with monitoring cameras and placed at maritime infrastructures, represents a step forward in maritime situational awareness~\cite{mittal2019survey,zhao2019embedded}. These systems can enable on-site deep-learning-based ship recognition, offering significant advantages such as reduced network bandwidth and energy usage, minimized latency, and enhanced security \cite{ning2020heterogeneous}.
The local processing of images using embedded systems directly at the infrastructure facilitates the spatial understanding of the maritime situation~\cite{sattler2023maritime}. 
Recognized and georeferenced ships using an embedded system can then be seamlessly integrated into web services, allowing their display (e.g. on maps) within the situational awareness system~\cite{flenker2021marlin}. This enhances real-time visualization and enriches the overall situational awareness by providing operators with accurate and timely spatial information \cite{flenker2021marlin}. 

Real-time and accurate ship recognition, classification, and georeferencing are essential, not just for improved spatial visualization. Beyond visualization, its combination with other data sources, such as \gls{ais}, satellite imagery and radar systems can further enhance the overall situational understanding~\cite{flenker2021marlin}. Therefore, the faster the image processing occurs, the better it supports the creation of a comprehensive real-time situational picture by fusing with additional maritime data, thereby elevating the operational effectiveness of maritime situational awareness efforts~\cite{flenker2021marlin}.

We have seen, that in the context of enhancing maritime situational awareness, optimized real-time processing is paramount. The objective, therefore should be to ensure that the developed ship recognition and georeferencing system operates with the highest possible accuracy and the shortest inference times on embedded systems. This dual focus on speed and accuracy is critical for facilitating the fusion of the developed methodologies with other sensor data and services, thereby enabling safer, more secure, and more efficient maritime operations. 

This thesis presents a compilation of explorations, methods and results proposed in the publications shown in Chapter~\ref{chap:publications}, which will be referenced throughout the manuscript from~\ref{paper:1} to~\ref{paper:6}. The goals and contributions of this thesis, achieved within these publications, are summarized as follows:

\begin{itemize}[leftmargin=*]
    \item Production of a real-world maritime dataset for ship recognition and georeferencing, advancing the research field of maritime situational awareness.
    \subitem \textbf{Contribution:} Creation and publication of the ShipSG dataset, which provides a comprehensive set of annotated images for ship recognition and georeferencing~\ref{paper:2}.
    \item Investigation and development of a ship recognition architecture that seeks for enhanced real-time ship recognition, able to run in real-time embedded systems.
    \subitem \textbf{Contribution:} In-depth study of state-of-the-art methods for ship recognition, using ShipSG and other datasets~\ref{paper:1}\ref{paper:2}\ref{paper:3}\ref{paper:4}.
    \subitem \textbf{Contribution:} Introduced ScatYOLOv8+CBAM, an innovative ship recognition architecture optimized for real-time processing on embedded systems~\ref{paper:5}.
    \item Proposal of an efficient solution for processing full-resolution images on embedded systems, allowing the recognition of small and distant ships.
    \subitem \textbf{Contribution:} Introduced an improved slicing method that enables the processing of full-resolution images for the recognition of small and distant ships on embedded systems~\ref{paper:6}.
    \item Innovation in the field of ship georeferencing using monocular images by developing a methodology that does not rely on prior camera knowledge.
    \subitem \textbf{Contribution:} Developed a novel ship georeferencing methodology using homographies that operates without requiring prior camera calibration~\ref{paper:1}\ref{paper:2}\ref{paper:5}.
    \item Optimization of real-time ship recognition and georeferencing methodologies for their deployment on embedded systems, balancing performance with computational efficiency.
    \subitem \textbf{Contribution:} Further improvement of the ScatYOLOv8+CBAM architecture for efficient deployment on embedded systems, balancing computational efficiency with high performance~\ref{paper:6}.
    \item Demonstration of the practical application of the methodologies by integrating them into systems for improved maritime situational awareness in a variety of applications.
    \subitem \textbf{Contribution:} Successfully integrated the developed methodologies into applications such as ship georeferencing displays including map-based visualization, abnormal ship behavior detection, camera integrity assessment, and 3D ship reconstruction, showcasing their effectiveness in enhancing maritime situational awareness~\ref{paper:1}\ref{paper:2}\ref{paper:3}\ref{paper:4}\ref{paper:5}.
\end{itemize}

The remaining chapters of this thesis are organized as follows:

\begin{description}
    
    \item[\textbf{\Cref{chap:fundamentals} \nameref{chap:fundamentals}}] \hfill \\
    This chapter dives into the technical background of modern object recognition, introducing key concepts. It explores how deep learning has transformed computer vision for object recognition and how it can be leveraged.
    \item[\textbf{\Cref{chap:sota} \nameref{chap:sota}}] \hfill \\
    This chapter focuses on relevant state-of-the-art, highlighting areas in ship recognition and georeferencing where current research falls short. The chapter presents maritime datasets and object recognition methods essential for this thesis as well as potential improvements, prior ship georeferencing research and deployment on embedded systems. 
    \item[\textbf{\Cref{chap:shipsg} \nameref{chap:shipsg}}] \hfill \\
    This chapter presents the creation of ShipSG\footnote{\url{https://dlr.de/mi/shipsg}}, a novel dataset for ship recognition and georeferencing. ShipSG provides the foundation for this thesis, enabling the development and evaluation of the methods presented in the subsequent chapters.
    \item[\textbf{\Cref{chap:ship_rec} \nameref{chap:ship_rec}}] \hfill \\
    This chapter shows the initial exploration of deep-learning-based methods for ship detection and segmentation, revealing their potential applications, such as ship georeferencing, abnormal ship behavior detection,
    camera integrity assessment, and 3D ship reconstruction. The study of state-of-the-art instance segmentation methods sets the stage for the custom developments and analysis proposed in subsequent chapters.
    \item[\textbf{\Cref{chap:adv_ship_rec} \nameref{chap:adv_ship_rec}}] \hfill \\
    This chapter addresses the need for fast and accurate algorithms on embedded systems for real-world use. While ship detection was proven to perform well, deploying instance segmentation (better for ship georeferencing) on embedded systems was shown to be a significant challenge. This chapter addresses this gap by proposing a customized real-time segmentation method (ScatYOLOv8+CBAM), deployed on an embedded system. It also proposes a method to improve the segmentation accuracy for small and distant ships by processing full-resolution images, crucial for better maritime situational awareness.
    \item[\textbf{\Cref{chap:ship_geo} \nameref{chap:ship_geo}}] \hfill \\
    This chapter focuses on the georeferencing of the ships recognized using monocular cameras to improve maritime situational awareness. This involves the development of a method to present ships on a global scale using only single images without prior camera knowledge. The chapter first explains homographies and then details the proposed method for georeferencing ship bounding boxes, along with the calculation of ship heading direction from optical flow. Finally, the chapter quantitatively analyses how this monocular ship georeferencing improves maritime situational awareness.
    \item[\textbf{\Cref{chap:conclusion} \nameref{chap:conclusion}}] \hfill \\
    This chapter summarizes the contributions and key findings of this thesis, and concludes the outcome of the produced results.
    \item[\textbf{\Cref{chap:future} \nameref{chap:future}}] \hfill \\
    This chapter presents the challenges encountered throughout the thesis and proposes new research lines to approach future work.
    \item[\textbf{\Cref{chap:publications} \nameref{chap:publications}}] \hfill \\
    This chapter presents the list of publications used in this compilation thesis and includes a short summary of my contributions to each publication.
    
\end{description}

\chapter{Fundamentals of Modern Object Recognition}
\label{chap:fundamentals}

As motivated in Chapter~\ref{chap:intro}, the combination of computer vision and deep learning offers a potent solution for automatic ship recognition using optical monitoring cameras. This chapter provides a technical overview of concepts to understand modern object recognition, starting with the role of supervised learning in computer vision, and followed by the use of deep learning for the two computer vision tasks of interest in this thesis, object detection and instance segmentation.

\section{Supervised Learning in Computer Vision}

Computer vision is a discipline within \gls{ai} that allows machines to process and interpret visual data. 
By harnessing algorithms and data, computer vision systems can identify and classify objects, and make decisions based on visual inputs similar to the way humans do~\cite{szeliski2022computer}. 
The field of computer vision has significantly advanced with deep learning, a subfield of machine learning, particularly through the use of \glspl{cnn}~\cite{szeliski2022computer}.
Preceding computer vision approaches relied on hand-engineered feature extraction. Deep learning, on the other hand, utilizes vast amounts of visual data to train hierarchical structures of neurons that excel at identifying patterns to therefore perform automatic feature extraction~\cite{goodfellow2016deep}. Thanks to the use of \glspl{gpu}, deep learning with \glspl{cnn} have significantly surpassed the performance of traditional algorithms in tasks like image classification, object detection, and instance and semantic segmentation~\cite{talaei2023deep}. The computational power of \glspl{gpu}, due to the parallel processing capabilities, enabled the training of deep networks with millions of parameters, allowing for the extraction of complex features from large datasets. 

Machine learning algorithms are commonly categorized into supervised learning, which requires labeled data for training, unsupervised learning, which operates on unlabeled data to find patterns, and semi-supervised learning, which uses a combination of labeled and unlabeled data to train models~\cite{sammut2011encyclopedia}.
In certain real-world applications, supervised learning is preferred for training models with real-world annotated datasets, ensuring accurate identification and categorization of objects represented in the data~\cite{chai2021deep}.

In supervised learning, models are trained on datasets labeled by human experts~\cite{Cunningham2008}. During training, the supervised model adjusts its parameters by measuring the deviation from the actual labels~\cite{sammut2011encyclopedia}. Therefore, the annotation process involves pairing each training sample with its corresponding output labels, serving as a learning guide for the model. 
In computer vision tasks, such as image classification, labeled training images are used to predict classes on validation images~\cite{rawat2017deep}. In object detection tasks, the annotations and training extends for the classification and localization of objects in the image within bounding boxes. Segmentation tasks demand detailed annotations, labeling each pixel by class~\cite{zaidi2022survey}.

In summary, supervised learning has greatly advanced computer vision tasks, while also highlighting the continuous need for models that can learn effectively from annotated data.

\section{Deep-Learning-Based Object Recognition}
\label{sec:fundamentals_or}

\begin{figure}[h]
\includegraphics[width=10 cm]{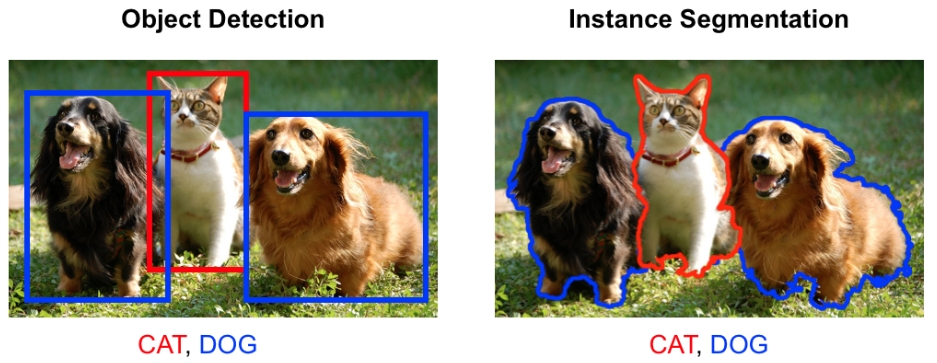}
\centering
\caption[Example of object detection and instance segmentation on an image.]{Example of object detection and instance segmentation on an image. Object detection involves bounding box localization and classification, whereas instance segmentation goes beyond that to provide a mask outlining the exact shape of each individual object instance. Adapted from~\cite{shanmugamani2018deep}.}\label{fig:od_is}
\end{figure}

Object recognition in computer vision involves identifying and classifying objects in images~\cite{szeliski2022computer}. Two main tasks in the field are object detection and instance segmentation, which are essential for machine interpretation of visual data and widely used in autonomous driving, monitoring, surveillance and medical imaging applications, among others~\cite{chai2021deep, zaidi2022survey}. Figure~\ref{fig:od_is} depicts the difference between object detection and instance segmentation.

\textbf{Object Detection} aims to locate and classify objects within an image, including the determination of their presence and exact location within bounding boxes. 

\textbf{Instance Segmentation} advances beyond detection with bounding box by identifying each object instance in an image at the pixel level, delineating its shape with a mask. Unlike semantic segmentation, which classifies each pixel within the image as belonging to a certain class, instance segmentation recognizes each object instance separately and the rest is considered background. 

\subsection{Standard Architecture Description}

Modern deep learning architectures for detection and segmentation tasks extensively use \glspl{cnn}, featuring a combination of a backbone, a neck and head structure~\cite{szeliski2022computer}.

\begin{figure}[h]
\includegraphics[width=9 cm]{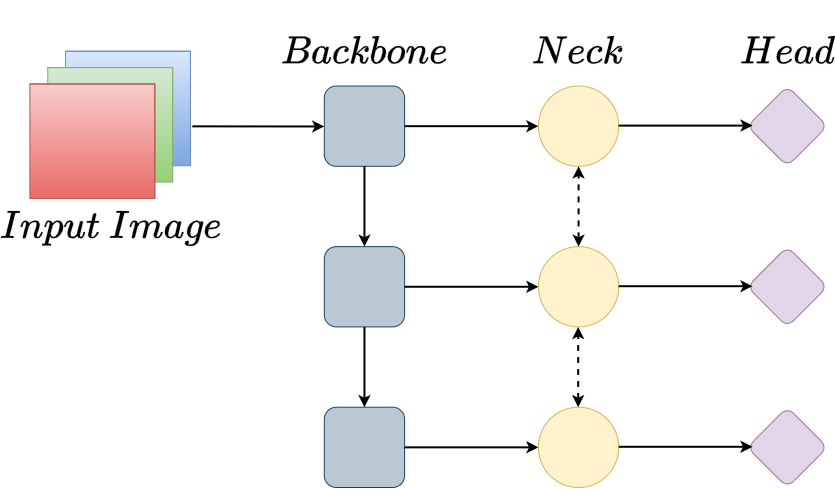}
\centering
\caption[Standard deep learning object recognition architecture.]{Standard deep learning object recognition architecture. See text for details.}\label{fig:std_arch}
\end{figure}

Figure~\ref{fig:std_arch} illustrates a standard deep learning object recognition architecture. In \gls{cnn}-like architectures, a feature map is the output of one filter (also known as kernel) applied across the previous layer to detect specific features~\cite{szeliski2022computer}. In the case of an object recognition architecture, the backbone focuses on extracting features by learning to recognize task-relevant patterns in visual data, performing changes in feature map resolution (width, height and channel number). 
The arrows in Figure~\ref{fig:std_arch} represent the flow of data through the network layers. 
As data progresses through the backbone, the resolution typically decreases to reduce the spatial dimensions while increasing the depth (number of channels) to create more abstract and complex feature representations~\cite{szeliski2022computer}.
Conversely, in the neck, the resolution can either decrease or increase. Typically, in object detection and instance segmentation tasks, the neck often includes upsampling to increase spatial information, allowing a more accurate location of objects of interest~\cite{szeliski2022computer}.
The neck fuses and aggregates features from different resolutions, acting as a bridge between the backbone and the head. The head performs specific tasks based on these features, such as detection, segmentation, and classification. Although they are different tasks, detection and segmentation share similarities in terms their architecture, with each head designed to perform the desired task. However, the design of backbone and neck can be tailored to perform better for the task of interest. 

As illustrated by Figure~\ref{fig:std_arch}, deep learning architectures for object recognition that use ~\glspl{cnn} typically comprise blocks that represent combinations of structured layers to process the visual data. These blocks include include a combination of convolutional, pooling, upsampling, activation and regularization layers~\cite{lecun2015deep}. 

\begin{figure}[h]
\includegraphics[width=13 cm]{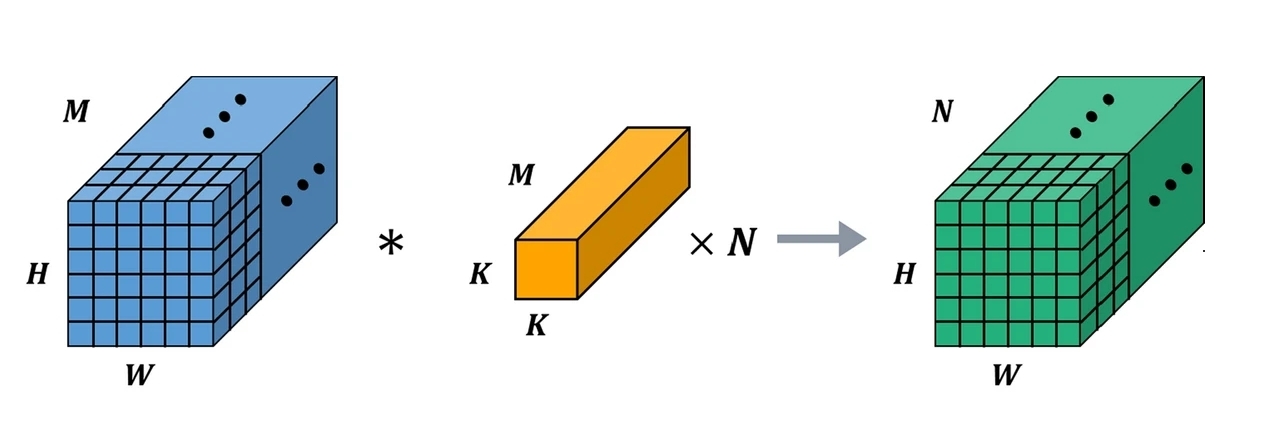}
\centering
\caption[Illustration of a standard convolution operation.]{Illustration of a standard convolution operation, taken from~\cite{zhang2020lightweight}. The input volume has dimensions $ H \times W \times M $ (height, width, and number of channels). A filter, also named kernel, of size $ K \times K \times M $ is convolved with the input, producing an output volume of dimensions $ H \times W \times N $, where $ N $ is the number of filters. This process involves sliding the filter over the input and computing the dot products between the filter weights and local regions of the input.}
\label{fig:conv_layer}
\end{figure}

Convolutional layers apply filters (kernels) to the input to create feature maps. These filters contain learned weights, which are adjusted during training to optimize feature extraction (see Figure~\ref{fig:conv_layer}). The convolution operation involves sliding the filter over the input to compute dot products between the filter weights and local regions of the input, generating feature maps that capture different aspects of the input data.
Though not depicted in Figure~\ref{fig:conv_layer}, to further adjust the output, a learnable bias term is normally also added to each output element.

\begin{figure}[h]
\includegraphics[width=8 cm]{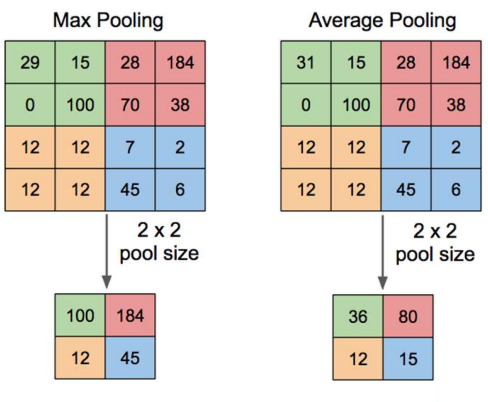}
\centering
\caption[Illustration of max pooling and average pooling operations.]{Illustration of max pooling and average pooling operations with a $2\times2$ pool size, taken from~\cite{zhang2020lightweight}. Max pooling selects the maximum value from each $2\times2$ block, while average pooling computes the average value from each $2\times2$ block, reducing the spatial dimensions of the input feature maps.}
\label{fig:pooling_layer}
\end{figure}

Pooling layers perform a fixed operation to reduce the spatial dimensions of the feature maps, down-sampling the input to reduce computational load and enhance invariance to small translations. As seen in Figure~\ref{fig:pooling_layer}, there are two common types of pooling: max pooling and average pooling. Both types of pooling effectively reduce the spatial dimensions while preserving important spatial features. Moreover, pooling layers reduce the spatial resolution of feature maps to combat overfitting, which happens when a model memorizes training data, failing to generalize to new, unseen data.

In contrast to pooling, upsampling layers perform the opposite operation by increasing the spatial dimensions of the feature maps. Upsampling can be achieved through various methods, such as nearest-neighbor interpolation, bilinear interpolation, or transposed convolutions~\cite{szeliski2022computer}. 
Upsampling layers are used to increase resolution when finer detail is necessary. 

Activation layers introduce non-linearity, enabling the network to capture complex patterns~\cite{lecun2015deep}. Typical activation functions include \gls{relu} and \gls{silu}. The \gls{relu}, given by $f(x) = \max(0,x)$, function provides a linear output that is zero for negative inputs and linear with a slope of 1 for positive inputs. \gls{silu}, given by $f(x) = x \cdot \sigma(x)$, incorporates a sigmoid function, described as

\begin{equation}
\sigma(x) = \frac{1}{1 + e^{-x}}
\end{equation}

allowing it to handle negative inputs more dynamically by scaling the output in a non-linear fashion.

Regularization layers, such as dropout and batch normalization, are also integrated into \gls{cnn} architectures. Dropout randomly omits neurons during training to enhance generalization, while batch normalization scales the output of a layer to have a mean of zero and a variance of one, which can expedite training and improve overall performance~\cite{lecun2015deep}.

State-of-the-art deep learning architectures, such as those that will be described in Chapter~\ref{chap:sota}, often combine the above-presented layers in innovative ways to enhance model performance and efficiency. For instance, methods like ResNet, normally used as backbone in object recognition architectures, introduce skip connections that allow gradients to flow more easily through very deep networks, facilitating training ~\cite{he2016deep}. Other advancements involve combining the layers in specific configurations to achieve desired properties. For example, Feature Pyramid Networks (FPNs) employ a combination of convolutional and upsampling layers to create feature maps at different scales, allowing the model to better handle objects of varying sizes within an image~\cite{lin2017feature}. Additionally, various forms of attention mechanisms (see Section~\ref{sec:attention}) can be integrated using these building blocks to selectively focus on relevant parts of the feature maps, leading to improved performance~\cite{vaswani2017attention}. 
In summary, the combination of layers are foundational and their interactions is crucial for the advancements in contemporary deep learning models. 

\subsection{Attention Mechanisms}
\label{sec:attention}

Neural networks use attention mechanisms to allow models to dynamically focus on the most relevant parts of the input data, enhancing their ability to process complex information~\cite{vaswani2017attention,bahdanau2014neural}. Initially introduced in \gls{nlp} for tasks like language translation~\cite{bahdanau2014neural}, attention mechanisms have since become integral to various deep learning applications, including computer vision. At the core of these mechanisms are attention weights, learned during training, which determine the importance of different parts of the input~\cite{Luong2015}. For example, in language translation, this results in an $n \times n$ attention matrix or map, where $n$ is the number of words~\cite{bahdanau2014neural}. 

The implementation of attention mechanisms involves the calculation of the attention weights using a score function~\cite{bahdanau2014neural}. This is achieved by applying the input to a learned weighted matrix that computes relevance scores using functions like dot product~\cite{vaswani2017attention}, or pooling operations (e.g., max pooling, average pooling)~\cite{woo2018cbam}. These scores create an attention map, highlighting the importance of different input parts. The network uses this map to prioritize crucial information, learning to distribute focus effectively across the input data.

In computer vision, the term attention translates to individual pixels attending to other pixels, or patches of pixels attending to other patches, leading to an attention map that captures the relationships across different regions of the image~\cite{guo2022attention}. However, this poses significant computational challenges. To address these challenges, various strategies have been proposed in the literature:

\begin{itemize}
    \item \textbf{Dimensionality reduction} with convolutional layers and pooling operations are often used to reduce the dimensions of the input before applying attention, decreasing the computational load by working with smaller feature maps~\cite{woo2018cbam, Hu2018}.
    \item \textbf{Hierarchical attention} mechanisms apply attention at different scales or hierarchies, allowing the model to first focus on broad, coarse details and progressively refine its attention to finer details, thus significantly reducing complexity~\cite{Yang2016}.
    \item \textbf{Local attention} restricts the attention mechanism to a local neighborhood around each pixel, limiting the number of interactions and thereby reducing the computational burden~\cite{Parmar2018}. 
    \item \textbf{Spatial attention} mechanisms identify important spatial locations within an image, thus allowing the model to concentrate on critical regions~\cite{Xu2015}.
    \item \textbf{Channel attention} mechanisms focus on identifying important feature channels within a \gls{cnn}, thereby improving the model's feature representation~\cite{Hu2018}.
\end{itemize}

Attention mechanisms are, therefore, used to enhance the ability of models to focus on key parts of an image. Specifically for computer vision tasks, they have been incorporated to \glspl{cnn} to improve performance in large-scale image classification tasks~\cite{Hu2018, Wang2017}. Additionally, they have been applied in object detection and instance segmentation tasks by incorporating spatial and channel-wise attention, which improves feature representation and accuracy~\cite{woo2018cbam, Hu2018, Chen2017, brauwers2021general}. 

\subsection{Object Classification and Postprocessing}

To accurately classify objects, after they are detected or segmented, the final layer of the head, usually a fully connected layer~\cite{goodfellow2016deep}, provides a confidence score for each potential prediction, which reflects the likelihood that a recognized object belongs to a specific class. Multiple prediction proposals (bounding boxes or masks) for the same object require further postprocessing beyond the head to enhance the accuracy and reliability of a prediction. Standard object recognition architectures include postprocessing after the head to refine the bounding boxes or segmentation masks recognized by the model and to eliminate redundant predictions~\cite{szeliski2022computer}. 
A key component of this postprocessing phase is \gls{nms}~\cite{gong2021review}, a technique designed to eliminate redundant bounding boxes or segmentation masks that pertain to the same object. Essentially, \gls{nms} ensures that each detected object is represented exclusively by the single, most accurate bounding box or mask, thereby preventing clutter and providing a clearer output.

\begin{figure}[h]
\includegraphics[width=13cm]{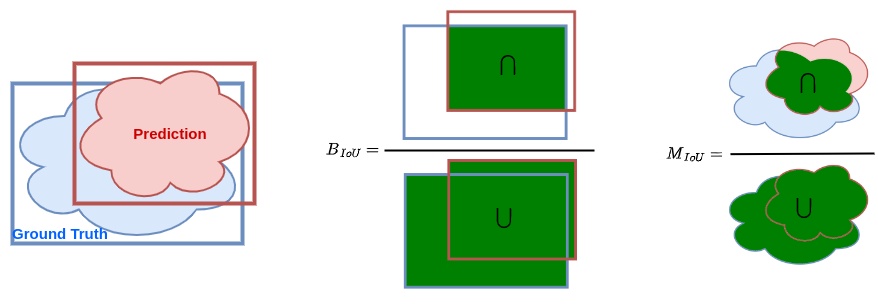}
\centering
\caption[Joint illustration of Intersection over Union calculation.]{Joint illustration of Intersection over Union calculation for boxes and masks. On the left, the ground truth, and predicted bounding boxes and masks are shown. In the middle, the IoU for the bounding box is visualized. On the right, in the case of instance segmentation, the IoU for the mask is calculated.}\label{fig:iou}
\end{figure}

To decide which bounding boxes or masks to keep and which to discard, \gls{nms} relies on the confidence scores of the predictions, together with a metric known as \gls{iou}. This metric measures the overlap between two areas—in this case, the area of overlap between a predicted bounding box or mask and the ground truth, as shown in Figure~\ref{fig:iou}. The \gls{iou} helps in determining the accuracy of the predictions by quantifying how closely the predicted bounding boxes or masks align with the actual objects in the image.

Mathematically, $B_{IoU}$ (bounding box) and $M_{IoU}$ (mask) can be denoted as:
\begin{multicols}{2}
  \begin{equation}
    B_{IoU} = \frac{\text{area}(B_{pred} \cap B_{gt})}{\text{area}(B_{pred} \cup B_{gt})}
  \end{equation}\break
  \begin{equation}
    M_{IoU} = \frac{\text{area}(M_{pred} \cap M_{gt})}{\text{area}(M_{pred} \cup M_{gt})}
  \end{equation}
\end{multicols}

This ratio ranges from 0 to 1, where 0 indicates no overlap and 1 indicates perfect overlap. In practice, an \gls{iou} threshold is set (e.g., 0.5 or 50\%) to classify predictions as true positives or false positives. The \gls{nms} filters the best final prediction from the possible proposals, represented as $P_{final} = NMS(P,S,\tau)$, for a set of predictions $P$ (either masks or bounding boxes) with associated confidence scores $S$ and an \gls{iou} threshold $\tau$.

Following the discussion of object recognition architectures and postprocessing, it becomes relevant to address the practical aspects of implementing these frameworks. PyTorch~\cite{NEURIPS2019_9015} is a popular deep learning library for computer vision, valued for its dynamic computation graph and efficient \gls{gpu} memory management. Its straightforward syntax simplifies the implementation of supervised \glspl{cnn}, making it ideal for research and development. This thesis leverages PyTorch to develop models for ship recognition.

\subsection{Training Process}

\begin{figure}[h]
\includegraphics[width=10 cm]{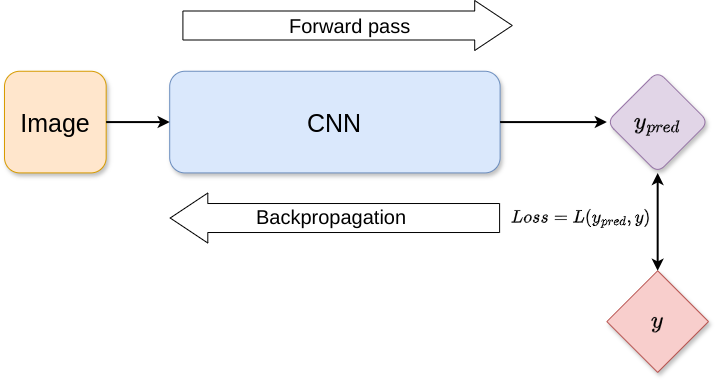}
\centering
\caption[Schematic of the Convolutional Neural Network (CNN) Training Process.]{Schematic of the Convolutional Neural Network (CNN) Training Process. An input image is passed through the CNN during the forward pass, resulting in a predicted output ($y_{pred}$). The prediction is compared to the true value ($y$) using a loss function ($L(y_{pred},y)$), and the error is propagated back through the network during backpropagation to adjust and improve the model weights.}\label{fig:train_process}
\end{figure}

The training process of \glspl{cnn} for object detection and segmentation includes forward and backward propagation, as illustrated by Figure~\ref{fig:train_process}. 

During forward propagation, the input data is fed through the network to output a prediction. The choice of loss function is crucial for model performance. Commonly used loss functions for different tasks include Binary Cross-Entropy Loss, Focal Loss, Bounding Box Loss, Objectness Loss, and Pixel-wise Cross-Entropy. Each of these loss functions addresses specific aspects of the prediction problem, such as class imbalance (Focal Loss) or spatial localization accuracy (Bounding Box Loss). 
These loss functions, sometimes used in combination or with other loss functions, help the model learn from its mistakes and achieve optimal performance~\cite{talaei2023deep}.
Further description and mathematical definitions of these loss functions can be found in reference~\cite{szeliski2022computer}.

Backward propagation, based on a loss function, then adjusts the network weights to minimize discrepancies between the predictions and the ground truth~\cite{lecun2015deep}. This adjustment process involves calculating gradients of the loss function with respect to the network parameters. These gradients indicate the direction and magnitude of the changes needed to reduce the loss.

Optimization algorithms use these gradients to update the network parameters iteratively. Common optimization algorithms include Stochastic Gradient Descent (SGD), Adam, RMSprop, and AdaGrad~\cite{goodfellow2016deep}. Each algorithm has its strengths and is chosen based on the specific requirements of the task. The goal is to minimize the loss function, thereby improving the performance of the model. The learning rate, a key factor in this process, determines the size of the updates. Optimization involves multiple passes through the dataset, known as epochs, where the network parameters are refined to achieve better accuracy and generalization~\cite{goodfellow2016deep}.

\subsection{Evaluation Metrics}

Evaluating the performance of object recognition models is critical to understanding their effectiveness and accuracy. 
The \gls{map} is a commonly-used metric to evaluate object detection and segmentation performance~\cite{lin2014microsoft}. 
Expressed in percentage, it is calculated as the mean of all the Average Precisions (AP) for all classes present in the dataset at a given \gls{iou} threshold. This is, mathematically:

\begin{equation}
    mAP_{\tau} = \frac{1}{C} \sum_{c=1}^{C} AP_{\tau,c}
\end{equation}

Here, $mAP_\tau$ represents the \gls{map} at an IoU threshold $\tau$, calculated by averaging the AP values across all $C$ classes. For the calculation of AP for each class, true positives are counted when the \gls{iou} of the prediction exceeds the given threshold $\tau$. In the case of object detection, a true positive is confirmed when the ~\gls{iou} of the predicted bounding box exceeds the~\gls{iou} threshold. For instance segmentation, true positives are based on the overlap between the predicted mask and the ground truth mask at the IoU threshold. This distinction in true positive calculation signifies the different evaluation approaches between object detection and instance segmentation.

It is common in the field of object recognition to refer to \gls{map} as a short form of $mAP_{0.5:0.95}$~\cite{szeliski2022computer}. The $mAP_{0.5:0.95}$ accounts for \gls{map} values at \gls{iou} thresholds that range from 0.5 to 0.95, in increments of 0.05. The formula would therefore be defined as:

\begin{equation}
    mAP = \frac{1}{N} \sum_{\tau=0.5}^{0.95} mAP_{\tau}
\end{equation}

Where $N$ represents the number of thresholds, which is 10 in the case of the range $0.5:0.95$. This comprehensive evaluation across several \gls{iou} thresholds provides insights into the performance of the model at different levels of strictness in object localization against the ground truth. 

Additionally, the \gls{map} can accomodate objects of varying sizes by further categorizing it based on the pixel area of the detected objects~\cite{lin2014microsoft}:

\begin{itemize}
    \item $mAP_s$ (small) if $area \leq 32^2$ pixels
    \item $mAP_m$ (medium) if $32^2 < area \leq 96^2$ pixels
    \item $mAP_l$ (large) if $area > 96^2$ pixels
\end{itemize}

This distinction per object size allows for a more detailed analysis of performance, especially in datasets with a wide range of object sizes, by highlighting its ability to detect small, medium, and large objects.

In order to compare results with existing standards, datasets such as COCO~\cite{lin2014microsoft}, with over 330,000 images and detailed annotations, are used as a resource for training and evaluating computer vision models in object detection and instance segmentation. Its diverse image collection makes it a valuable resource for researchers and developers. In the literature of experimental general purpose object recognition, it is a standard practice to evaluate general-purpose object recognition models performance using COCO as a benchmark~\cite{szeliski2022computer} with the \gls{map} as metric. 

\chapter{Relevant State of the Art}
\label{chap:sota}
\glsresetall

Expanding on the foundations of Chapter~\ref{chap:fundamentals}, this chapter identifies key limitations in approaches to maritime situational awareness prior to this thesis.  Specifically, this chapter reviews existing maritime datasets, real-time ship recognition algorithms, georeferencing techniques and gives an overview of the technologies available for deployment on embedded systems. By analyzing limitations, this chapter lays the groundwork for the studies and developments of a novel approach presented in later chapters.

\section{Real-world Maritime Datasets}

The accuracy of a supervised learning model is greatly dependent on the quality and volume of the annotated data it is trained on, especially for real-world applications~\cite{Cunningham2008}. As deep-learning-based ship detection and segmentation rely on supervised learning, it is necessary to use domain-specific training datasets~\cite{chai2021deep}. The training set and annotations must accurately represent the variety of ways objects can appear in different conditions~\cite{sarker2021deep}. 

Real-world maritime monitoring requires image data with precise annotations for a broad range of ships and ship classes~\cite{qiao2021marine}. General-purpose detection and segmentation datasets, such as COCO~\cite{lin2014microsoft} or PASCAL VOC~\cite{everingham2015pascal}, therefore, do not suit the task of ship recognition and georeferencing as benchmark datasets for maritime awareness.
Relevant datasets in the literature for ship detection on video monitoring cameras are the \gls{smd}~\cite{prasad2017video}, Seaships7000~\cite{shao2018seaships}, and a dataset introduced by Chen et al.~\cite{chen2020video}. Moreover, other private datasets exist~\cite{ghahremani2019multi,nita2020cnn}, however the restricted access makes the experimental validation using them not possible. 
The accessible datasets, lack a variety of ship classes in their annotations and do not provide ship masks, necessary for ship georeferencing. 
The MarSyn dataset~\cite{ribeiro2022real} is a synthetic ship dataset that contains images rendered from synthetic 3D scenes for instance segmentation in six ship classes, without georeference from the ships annotated.

A literature review on ship detection and localization~\cite{teixeira2022literature} highlights the fact that while annotations for ship datasets should include more complex data such as latitude and longitude of the ship, available datasets primarily focus on the object classes and bounding boxes, without masks or geographic positions. However, as motivated in Chapter~\ref{chap:intro}, ship segmentation provides a more suitable solution for georeferencing. Therefore, we find the need for a publicly available real-world dataset, for ship segmentation and georeferencing, that includes footage of a maritime infrastructure as well as mask and georeferencing annotations of several classes of ships. 
This dataset, should aim towards the advancement and evaluation of ship recognition methods for the improvement of maritime situational awareness. 

\section{Ship Recognition Using Maritime Monitoring Footage}
\label{sec:sota_ship_rec}

To enhance maritime situational awareness, it is crucial to use methods that perfom ship recognition on maritime footage~\cite{prasad2017video}. However, these methods should not only recognize ships but also allow the gathering of essential information about them, such as their class and geographic location (georeference). It is vital to present this information in a simplified format to maritime operators for quick and effective decision-making~\cite{flenker2021marlin}. Additionally, deploying an embedded system, with a monitoring camera on board, enables deep-learning object recognition directly on-site. This approach reduces network bandwidth, minimizes latency, improves security, and offers cost-efficiency~\cite{ning2020heterogeneous}, but comes with the trade-off of lower computational power compared to high-end systems~\cite{mittal2019survey}.
To address these complexities effectively, it is important that ship recognition methodologies not only ensure high accuracy across various ship sizes and types but are also optimized for the constraints of embedded hardware. Furthermore, the inference speed of video-based ship segmentation is paramount, as it significantly contributes to the improvement of data fusion with other sensor data, leading towards more cohesive maritime situational awareness systems~\cite{flenker2021marlin}. 

\begin{sidewaysfigure*}
\centering
\includegraphics[width=\textwidth]{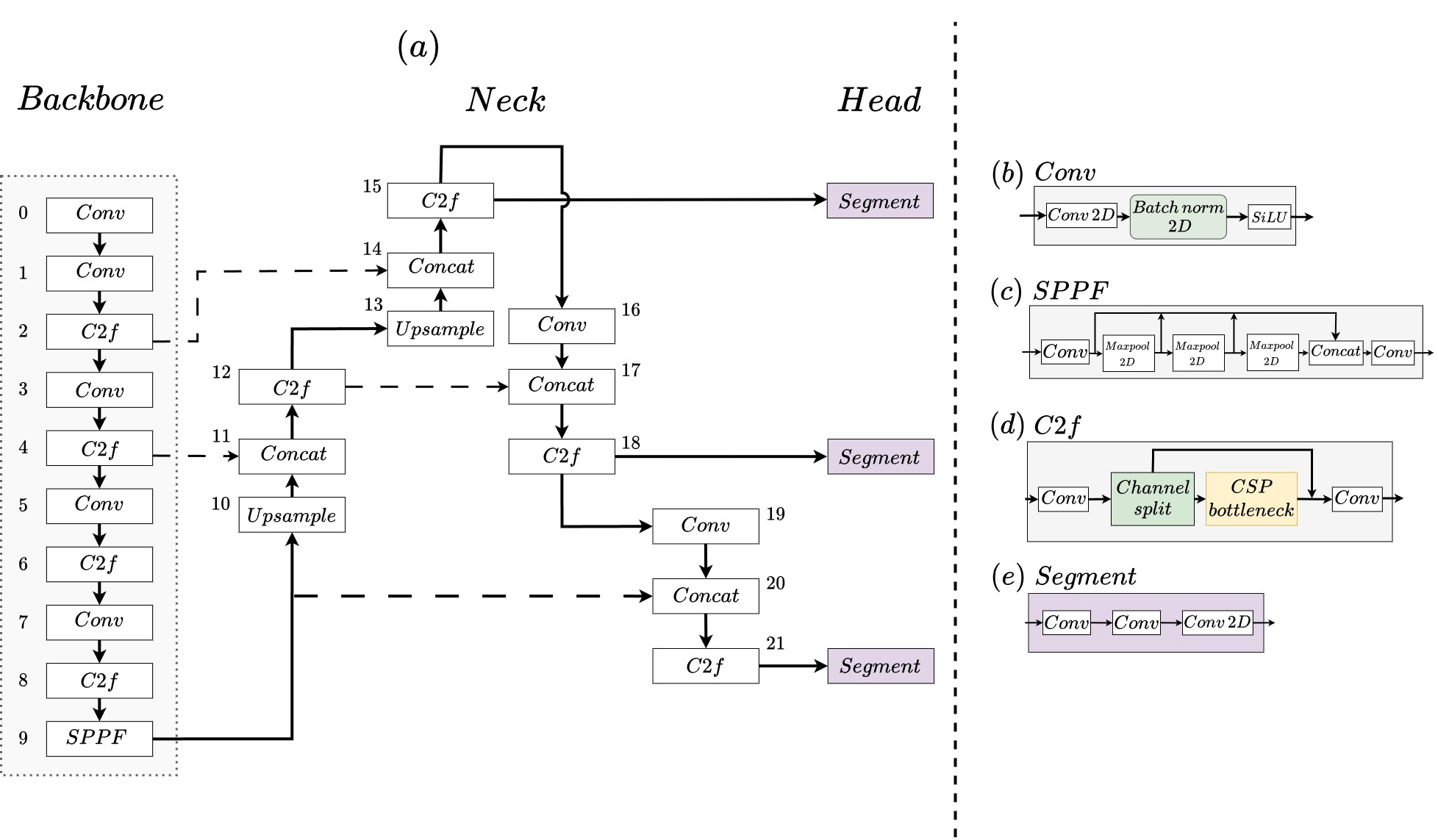}
\caption[The YOLOv8 architecture by Ultralytics.]{\footnotesize The YOLOv8 architecture by Ultralytics~\cite{jocheryolov8}. (a) Backbone, neck and head of the architecture. The numbers next to every block represent the sequential order followed by the implementation, from input to output. (b) Standard YOLOv8 convolutional block. (c) Spatial Pyramid Pooling Fast module of YOLOv8. (d) C2f with split channel operation and a \gls{csp} Bottleneck. (e) Segment block of YOLOv8 that performs segmentation. Modified from \ref{paper:5} \textcopyright 2023 IEEE.}\label{fig:yolov8}
\end{sidewaysfigure*}

These challenges underline the need to search for effective object recognition methods in the literature. The following list provides a brief overview of state-of-the-art object detection and instance segmentation methods that are particularly relevant:

\begin{itemize}
    \item \textbf{YOLOv4-CSP~\cite{wang2021scaled}}. The You-Only-Look-Once (YOLO) algorithm was introduced for real-time object detection~\cite{redmon2016you}. It divides the image into a grid and predicts bounding boxes and class probabilities for each grid cell, based on pre-defined anchors that act as reference points. These anchors are calculated by applying k-means clustering~\cite{sammut2011encyclopedia}, and serve a set of provisional bounding boxes for the object detection task.
    YOLOv4-CSP presented a substantial enhancement in speed and accuracy by leveraging two main innovations. The first innovation is the use of CSPDarknet53~\cite{bochkovskiy2020yolov4} as backbone, based on \gls{csp} network~\cite{wang2020cspnet}. The \gls{csp} network partitions the feature map into two parts: one part is processed through a series of layers while the other part bypasses these layers, ensuring better gradient flow and capturing patterns more effectively~\cite{wang2020cspnet}. The second innovation is the Bag of Freebies technique, which includes data augmentation, label smoothing, and additional regularization methods. These improvements reached a detection mAP of 47.5\% on COCO. 
    \item \textbf{Faster R-CNN~\cite{ren2015faster} and Mask R-CNN~\cite{he2017mask}}. Mask R-CNN is a two-stage instance segmentation method that was developed as an extension of the object detector Faster R-CNN. 
    They use the Region Proposal Network introduced in~\cite{ren2015faster} to identify object candidates and then refine these detections by classifying them and fitting precise bounding boxes. In Mask R-CNN, a fully convolutional network was added to regress the mask from the detected bounding boxes, reaching a mask mAP of 39.8\% on the COCO dataset with the ResNeXt-101 backbone~\cite{xie2017aggregated}.
    \item \textbf{DetectoRS~\cite{qiao2021detectors}} is a multi-stage instance segmentation method that enhances the use of recursive feature pyramid network~\cite{lin2017feature} and feedback connections~\cite{NIPS2014_19de10ad} for improved performance. It features an atrous convolution, a type of dilated convolution used to expand the receptive field, allowing it to capture larger areas of the input~\cite{yu2015multi}. DetectoRS uses ResNet-50~\cite{he2016deep} as its backbone, and achieves a 44.4\% mask mAP on the COCO dataset. 
    \item \textbf{YOLACT~\cite{bolya2019yolact}} emerged as one of the first real-time instance segmentation approach, operating in one-stage. It generates prototype masks through an independent fully convolutional network~\cite{long2015fully} and computes coefficients for adjusting these masks to the predicted bounding boxes. After suppressing overlapping detections with \gls{nms}, it filters the masks using anchor boxes. With a ResNet-101 backbone~\cite{he2016deep}, YOLACT attains a mask mAP of 34.1\% on the COCO dataset. 
    \item \textbf{Centermask-Lite~\cite{lee2020centermask}} is a one-stage instance segmentation method, optimized for real-time applications. It utilizes a spatial attention-guided mask branch within a fully convolutional object detector~\cite{guo2020fully} to refine proposed regions. It incorporates a novel backbone, VoVNet~\cite{lee2019energy}, which enhances feature map integration and, with VoVNet-39, achieves a 36.3\% mask mAP on the COCO dataset. 
    \item \textbf{YOLOv5~\cite{jocheryolov5}}, developed within Ultralytics framework\footnote{\url{https://github.com/ultralytics/}}, builds on YOLOv4 but utilizes PyTorch and introduces the AutoAnchor algorithm to automatically fine-tune anchor boxes over multiple iterations. It achieves a detection mAP of 50.7\% on COCO. 
    \item \textbf{YOLOv8~\cite{jocheryolov8}}, also developed by Ultralytics, builds upon previous YOLOv5. With a focus on real-time applications, this version supports a full range of vision tasks, including detection and instance segmentation. In this thesis, YOLOv8 plays a central role in the customized instance segmentation architecture proposed in~\ref{paper:5} and improved in~\ref{paper:6}, ScatYOLOv8+CBAM (Section~\ref{sec:scatyolo}). Therefore, YOLOv8 is described further than the previous state-of-the-art methods. The YOLOv8 architecture is divided into three main parts: Backbone, Neck, and Head, as illustrated in Figure~\ref{fig:yolov8}. The model uses the backbone CSPDarknet53~\cite{bochkovskiy2020yolov4} as previous YOLO versions, but includes the novel C2f module (Figure~\ref{fig:yolov8}(a)). The blocks found in the backbone are:

\begin{itemize}
    \item \textbf{Conv Block (Figure~\ref{fig:yolov8}(b))}: Each Conv block includes a 2D convolution, followed by batch normalization and a \gls{silu} activation function. This block reduces the spatial dimensions (width and height) and increases the number of channels.
    \item \textbf{SPPF Block (Figure~\ref{fig:yolov8}(c))}: The Spatial Pyramid Pooling Fast (SPPF) block performs multiple max-pooling operations at different scales, concatenates the results, and then applies a convolution. This block maintains the number of channels.
    \item \textbf{C2f Module (Figure~\ref{fig:yolov8}(d))}: This module contains a series of convolutional layers, a channel split operation, and a \gls{csp}bottleneck. The \gls{csp}bottleneck splits the input feature map into two parts: one part goes through a series of convolutional layers (bottleneck), while the other part bypasses these layers. The outputs are then concatenated, which helps in maintaining gradient flow and reducing computational load. This design enhances feature extraction and learning efficiency. The C2f module increases the number of channels.
\end{itemize}

The neck of YOLOv8 (Figure~\ref{fig:yolov8}(a)) is designed to generate feature pyramids that help the model handle objects at different scales. Further blocks of the neck are:
\begin{itemize}
    \item \textbf{Upsample}: This block increases the spatial dimensions (height and width) of the feature maps.
    \item \textbf{Concat}: The Concat blocks merge feature maps from different stages, increasing the number of channels.
\end{itemize}

The head (Figure~\ref{fig:yolov8}(a)) is responsible for producing the final output of the model. It includes the following:
\begin{itemize}
    \item \textbf{Segment Block (Figure~\ref{fig:yolov8}(e))}: This block consists of a series of convolutional layers and generates segmentation masks and classifies each detected object. The postprocessing of YOLOv8, not shown in Figure~\ref{fig:yolov8}, combines the output of both the pixel-level masks and class labels for each object.
\end{itemize}

YOLOv8 also offers five model sizes, these being, from the lightest and fastest to the deepest and most accurate: YOLOv8$n$, YOLOv8$s$, YOLOv8$m$, YOLOv8$l$, and YOLOv8$x$. It achieves a detection mAP of 53.9\% and a mask mAP of 43.4\%.

\end{itemize}

The above-listed methods were originally designed by their authors using COCO~\cite{lin2014microsoft} as benchmark dataset, serving this as a way of identifying robust and real-time models in the literature that could be suited for ship recognition tasks. 

An effective object recognition method for ship recognition should offer potential for integration with additional tasks, such as georeferencing, which, is vital for enhancing maritime situational awareness~\cite{flenker2021marlin}.
This necessitates a move beyond the utilization of existing state-of-the-art object detection and segmentation models, towards developing an advanced approach. Such an approach must be lightweight enough for embedded system deployment while maintaining or enhancing precision and speed. This highlights the need for innovative, efficient solutions capable of meeting the stringent demands of both performance and practicality in maritime situational awareness.

To tackle the challenges mentioned and boost performance in real-world scenarios with scarce data, real-time ship recognition can be improved by leveraging advanced methods like attention mechanisms and techniques from other fields, such as the 2D scattering transform. The 2D scattering transform, which uses wavelets, has been widely used in signal processing tasks such as speech recognition, time series analysis, astrophysics, and geosciences~\cite{singh2021deep, pan2020spatio, cheng2020new, rodriguez2021recurrent}.
Prior research has utilized wavelets in the computer vision field to perform multi-resolution geometric processing, multi-scale oriented filtering and image denoising~\cite{szeliski2022computer}.
In combination with deep learning and computer vision, the scattering transform has been used for image classification tasks~\cite{bruna2013invariant}, proving to provide a deep systematic understanding of how invariant features can be captured and utilized. It captures the essence of geometric and structural properties, which are crucial for recognizing complex patterns under various conditions~\cite{oyallon2018compressing}. 
The integration of the 2D scattering transform and attention to an object recognition method should aim to enhance ship recognition performance, offering an efficient and targeted solution to the previously outlined challenges. 
The technicalities of the 2D scattering transform are given in Chapter~\ref{chap:adv_ship_rec}.

Enhancing the performance of ship recognition systems through the integration of advanced techniques represents a significant step forward in maritime situational awareness. However, the practical application of these advancements in real-world monitoring scenarios brings to the forefront additional challenges, particularly in the recognition of small and distant ships. The effective recognition of small and distant ships ensures protection at the infrastructure by enabling early threat detection and accident prevention at maritime infrastructure~\cite{chen2020deep}. Using images at their original full-resolution, or even high-resolution cameras, is essential for this task~\cite{rekavandi2022guide}. However, deep learning techniques for object recognition on high-resolution images consume significant memory and necessitate larger neural networks, which complicates their real-time deployment~\cite{saponara2021impact}. Moreover, high-resolution processing on embedded systems with limited memory presents additional challenges, impacting performance and latency~\cite{mao2016towards}. 
Image super-resolution, which consists of the synthetic increase of the input image resolution, has been used in the literature~\cite{rekavandi2022guide}. In reference~\cite{wang2023uav}, their work introduced additional blocks and layers using transformers, which leverage self-attention extensively~\cite{zhu2023biformer} for better small object accuracy from aerial views~\cite{wang2023uav}. These solutions increase computational complexity beyond the capacity of embedded systems for real-time operation.
The \gls{sahi} method, introduced in reference~\cite{sahi23akyon}, splits high-resolution images into slices, enabling detection and segmentation of small objects. 
Slicing mechanisms, therefore, allow the processing of high-resolution images on embedded systems by dividing the images into manageable sections, thus reducing the computational load and memory requirements.
However, \gls{sahi} lacks native batch inference processing, and instead, processes slices sequentially. This highlights the need for a slicing method which can benefit real-time applications such as small ship segmentation, as proposed in this thesis.

\section{Georeferencing of Recognized Ships}

The field of object georeferencing from images extends across various domains, from aerial vehicle tracking using airborne cameras~\cite{han2011geolocation} to the vehicle geolocation in urban environments for autonomous driving~\cite{shami2024geo}. These methods, while effective within their respective ranges of operation (dozens of meters), require positioning systems on board for calibration.
The method proposed by~\cite{milosavljevic2017method} introduced an alternative that estimates georeferences from surveillance camera videos by aligning video frame points with geographic locations using a homography transformation to project the camera space onto orthophoto maps, which are geometrically corrected aerial photos, and Digital Elevation Models (DEMs). However, the application of high-resolution orthophotos and DEMs for maritime environments presents significant challenges, as these methods primarily model terrestrial elevations and are less effective for water surfaces, where the dynamic nature of water and its reflective properties complicate the creation of DEMs that could be used for ship georeferencing.

\begin{table}[h]
\caption[Ship georeferencing accuracy in existing literature.]{Ship georeferencing accuracy in existing literature. Note: Some entries lack reported uncertainty values for the positioning error.}
\label{tab:geo_sota}
\resizebox{\textwidth}{!}{%
\begin{tabular}{cccc}
\hline
\textbf{Source} & \textbf{System} & \textbf{Range to Object} & \textbf{Error (m)} \\ \hline
\cite{naus2021assessment} & Radar Antenna + GPS* & 1000 m & 6.5 \\ 
\cite{livingstone2014ship} & Synthetic Aperture Radar & 800 km & 13~$\pm$~23 \\
\cite{wei2020geolocation} & Opt. Remote Sensing & 36000 km & 165~$\pm$~109 \\
\cite{helgesen2020low} & Opt. Camera + GPS + IMU** & 400 m & 20 \\ \hline
\multicolumn{4}{l}{\footnotesize *Global Positioning System, **Inertial Measurement Unit}
\end{tabular}%
}
\end{table}

To transition from terrestrial to maritime applications, as discussed in Chapter~\ref{chap:intro}, ship georeferencing is a critical aspect. This process involves the assignment of geographic coordinates to ships detected in various data sources. Table~\ref{tab:geo_sota} shows a summary of ship georeferencing accuracies in the literature using different technologies.
Radar technologies have been a cornerstone in this field, providing real-time georeferencing at a speed of 1 Hz, as detailed in~\cite{naus2021assessment}. Despite their accuracy, radar systems often involve high costs and complex deployment requirements.
Parallel to radar, satellite technologies including optical remote sensing~\cite{wei2020geolocation} and synthetic aperture radar (SAR)~\cite{livingstone2014ship}, extend georeferencing capabilities over larger coverage areas. However, the effectiveness of these methods is constrained by their data cycle times ($\sim$minutes) and the satellite revisit schedules, limiting their temporal resolution.
Recent advancements have explored the use of video sequences for ship georeferencing. The work in~\cite{helgesen2020low} proposed a method which relies on the pinhole camera model calibration matrix~\cite{zhang2000flexible} to georeference ships detected in video frames. This approach, while innovative, requires prior knowledge of camera calibration, and its application has been limited to controlled conditions with a single video sequence of two small ships.

The methodologies and technologies reviewed reveal a landscape where accuracy, range, and cost are in constant negotiation. While radar and satellite methods offer comprehensive coverage, their practical deployment is often hindered by high costs and technical complexities. Conversely, camera-based approaches present a cost-effective alternative but are limited by the need for prior calibration or additional sources, such as orthophotos or DEMs. A solution that uses cameras without pose calibration would facilitate scalability in the deployment of the georeferencing method to existing monitoring cameras at the maritime infrastructure.

\section{Deployment on Embedded Systems}

Utilizing a \gls{gpu} on an embedded system, equipped with a monitoring camera, can allow for on-site deep-learning object recognition, streamlining the process significantly~\cite{mittal2019survey}. Processing images directly on the embedded system, rather than transferring them to a cloud or server, produces notable reduction in network bandwidth and latency, alongside cost savings and enhanced security~\cite{ning2020heterogeneous}. This integration facilitates real-time access to recognized and georeferenced ships through web services, enabling their display on maps for operators, boosting maritime monitoring~\cite{flenker2021marlin}.

\begin{table}[h]
\caption{Comparison of NVIDIA GPU modules, with focus on the Jetson family and high-end GPU-powered systems.}
\label{tab:gpu_summary}
\resizebox{\textwidth}{!}{%
\begin{tabular}{c|c|c|c|c}
\hline
\textbf{System Type} & \textbf{Module} & \textbf{CUDA Cores} & \textbf{Memory} & \textbf{\begin{tabular}[c]{@{}c@{}}Max Power \\ Consumption\end{tabular}} \\ \hline
\multirow{3}{*}{Edge Computing Device} & Jetson Nano & 128 & 4 GB & 10 W \\
 & Jetson TX2 & 256 & 8 GB & 15 W \\
 & Jetson AGX Xavier & 512 & 16GB & 30 W \\
 \hline
\multirow{2}{*}{High-End Device} & GV100 & 5120 & 32 GB & 250 W \\
 & A100 & 6912 & 80 GB & 400 W
\end{tabular}%
}
\end{table}

Within the spectrum of embedded systems widely utilized for deep learning and computer vision, the NVIDIA Jetson family\footnote{\url{https://developer.nvidia.com/embedded/jetson-modules}} stands out in the literature, offering both mobile and energy-efficient embedded \gls{gpu}-systems~\cite{chen2019deep}. Table~\ref{tab:gpu_summary} shows a comparison of three modules of the Jetson family compared against high-end server-based \gls{gpu} systems to contextualize their capabilities. We observe that the Jetson modules provide optimized balance between performance and energy efficiency, marking them as an optimal solution for vision-based systems where the on-site deployment is a requirement. Larger servers, are typically used for the training of the models that are later deployed on the Jetson. 

Jetson modules allow the use of \gls{gpu} computing for deep learning models developed with PyTorch. Additionally, to enhance deep learning efficiency, models can be converted into optimized engines using TensorRT~\cite{NVIDIATensorRT2024}, a practice recommended for deploying models on NVIDIA hardware, which leads to faster inference speeds~\cite{stepanenko2019using}. 
TensorRT is an NVIDIA library designed for high-performance deep learning inference, which includes optimizations for NVIDIA hardware.

The transition to export weights from PyTorch-trained models to TensorRT involves converting the trained deep learning models into a format that is optimized for inference on NVIDIA \gls{gpu}s. This process begins with the trained model in PyTorch. The model is then exported to an intermediate representation, often using \gls{onnx}\footnote{\url{https://github.com/onnx/onnx}}, which standardizes the model format for use across different deep learning frameworks~\cite{onnxruntime}. Once in \gls{onnx} format, the model is ready to be optimized by TensorRT, which analyzes the network to fuse layers, optimize kernel selection, and apply other enhancements that reduce memory footprint. The optimization process is automatically tailored to the unique architecture of the \gls{gpu}, making it more effective when it is performed directly on the intended target system.
By converting PyTorch models to TensorRT, deep learning models can achieve faster inference times, reduced memory usage, and the ability to choose precision formats (such as FP16 or INT8) that balance speed and accuracy.  

Several studies have leveraged NVIDIA Jetson modules for deploying deep learning models in various computer vision applications. For instance, in~\cite{zhao2019embedded}, the Jetson TX2 is employed for ship detection, showcasing the utility of Jetson modules in maritime object detection. The work in~\cite{heller2022marine} explored a comparison of marine object detection methods using the \gls{smd} dataset~\cite{prasad2017video} on the NVIDIA Jetson Xavier AGX. In the field of instance segmentation, the work in~\cite{panero2021real} utilized the NVIDIA Jetson AGX Xavier for real-time instance segmentation in driving traffic videos, showing the capability of the module to handle complex vision tasks in real-time scenarios. 

While the NVIDIA Jetson modules have been effectively utilized in various object detection and recognition tasks, there is a notable absence of research focusing on tailored architectures for real-time ship segmentation deployable on these embedded systems. This highlights a significant opportunity for innovation in developing efficient, real-time processing solutions specifically designed for maritime monitoring applications.
\chapter[ShipSG: Ship Segmentation and Georeferencing Dataset]{ShipSG: Ship Segmentation and Georeferencing Dataset}
\label{chap:shipsg}

In Chapter \ref{chap:sota}, we explored the current state of maritime situational awareness, highlighting the critical need for robust and efficient ship recognition methodologies and the challenges associated with georeferencing ships using maritime monitoring footage. This exploration underlined the limitations of existing datasets in supporting the development and evaluation of advanced ship recognition and georeferencing techniques. Motivated by these insights, the creation of a comprehensive dataset that includes precise annotations for ship segmentation and accurate georeferencing has become paramount. 
This chapter presents ShipSG, a novel dataset for ship segmentation and georeferencing using images from a fixed oblique perspective at maritime facilities. 
ShipSG serves as a foundational component of this thesis, enabling the evaluation of existing instance segmentation methods as detailed in Chapters~\ref{chap:ship_rec} compared against the custom architecture proposed in~\ref{chap:adv_ship_rec}. 
Additionally, the dataset has been instrumental in the quantitative assessment of our georeferencing approaches, as outlined in Chapter~\ref{chap:ship_geo}. 
A further description of the dataset is given in~\ref{paper:2}.
\textbf{The dataset was made public and is accessible upon request\footnote{\url{https://dlr.de/mi/shipsg}}}.

\section{Dataset Overview}

\begin{figure}[h]
\includegraphics[width=15.5 cm]{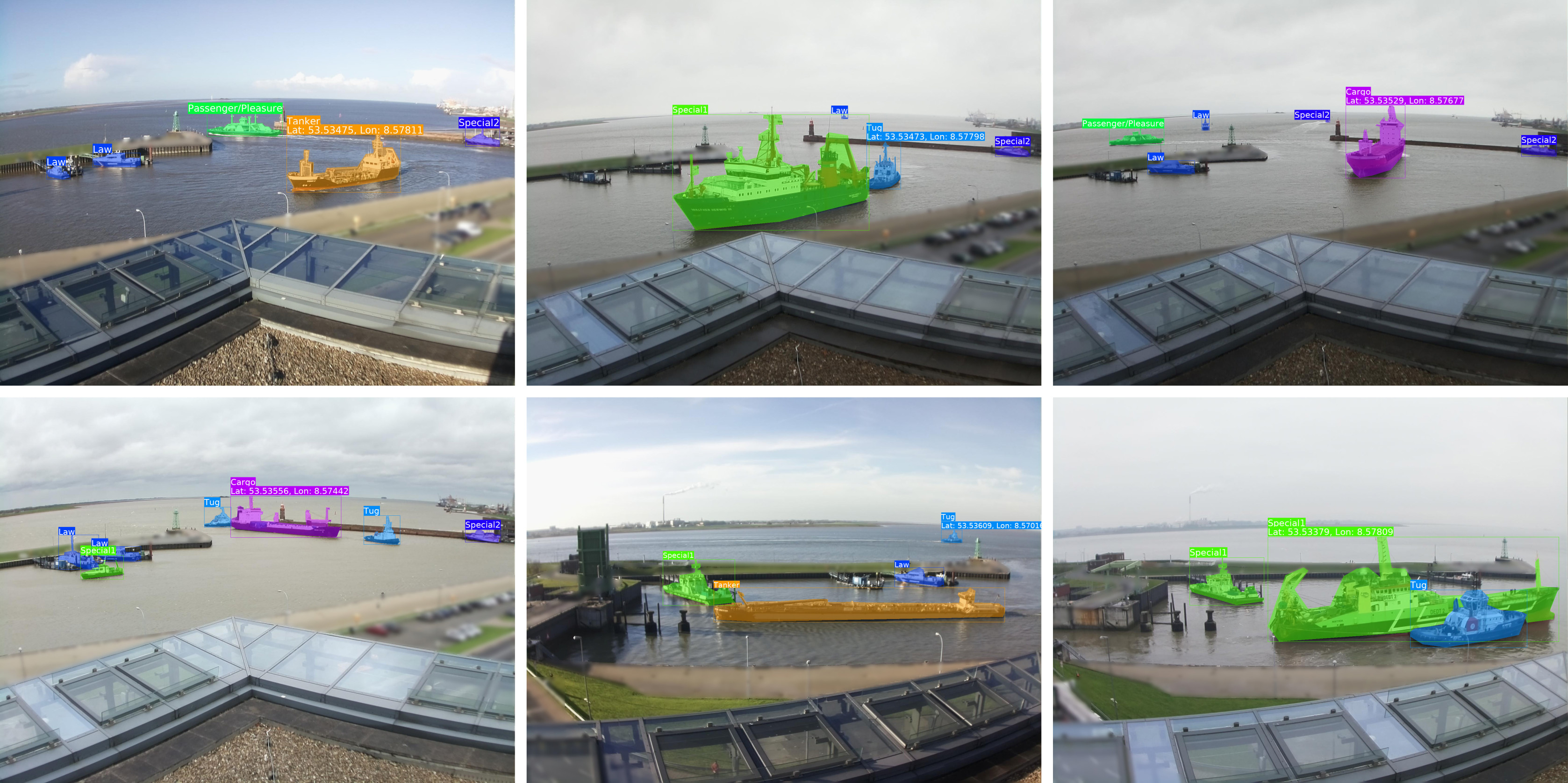}
\centering
\caption[Visualisation of ShipSG dataset samples]{Visualisation of ShipSG dataset samples with annotated ship masks, classes, and one ship position per image. Reprinted from the dataset site with permission from~\gls{dlr}.}\label{fig:dataset}
\end{figure}

The ShipSG dataset dataset was introduced in~\ref{paper:2} for the development and evaluation of instance segmentation and georeferencing methods using computer vision and deep learning, thus advancing the research field of ship recognition for maritime situational awareness. Some samples of ShipSG with annotated ship masks can be seen in Fig.~\ref{fig:dataset}. The dataset contains:

\begin{itemize}
    \item 3505 images ($2028\times1520$ pixels) from two cameras with static oblique view to the Doppelschleuse, Bremerhaven, Germany.
    \item 11625 annotated ship masks grouped in seven classes (see Fig.\ref{fig:class_samples}) with COCO format~\cite{lin2014microsoft}.
    \item 3505 geographic positions, consisting of the latitude and longitude of one of the masks within each image.
    \item 3505 \gls{ais} ship types\footnote{\href{https://coast.noaa.gov/data/marinecadastre/ais/VesselTypeCodes2018.pdf}{https://coast.noaa.gov/data/marinecadastre/ais/VesselTypeCodes2018.pdf}}, one per geographic position annotated.
    \item 3505 ship lengths, one per geographic position annotated.
\end{itemize}

The dataset was split into two sets: training and validation. 
The training set contains 80\% of the dataset, with 2804 images, and the remaining 20\% is used for validation, with 701 images. 

\section{Acquisition and Annotation}

\begin{figure}[h]
\includegraphics[width=10.5 cm]{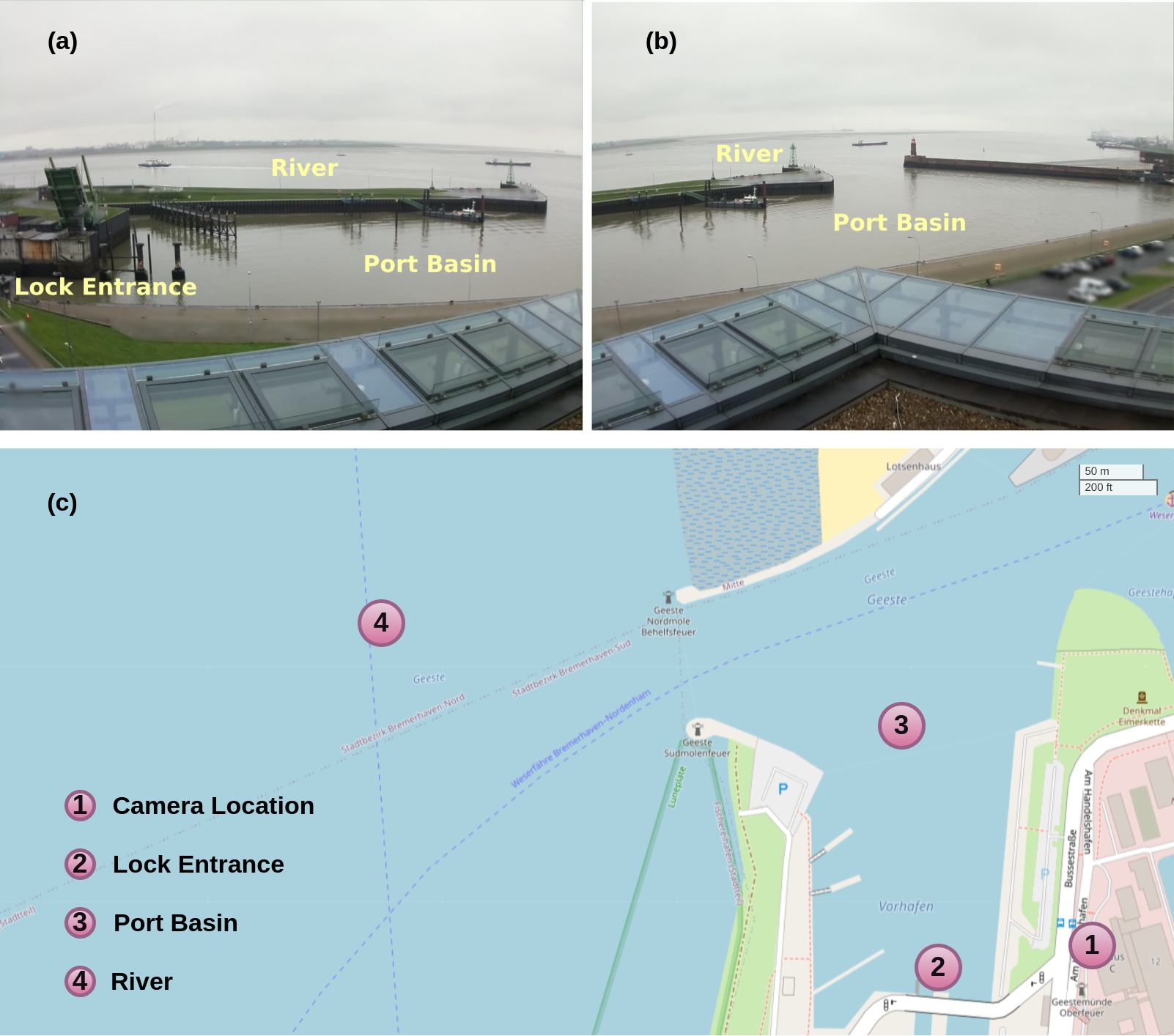}
\centering
\caption[View of each camera and identification of important elements in the scene.]{View of each camera and identification of important elements in the scene. (a) View of first camera. (b) View of second camera. (c) Notable elements in the scene (OpenStreetMap~\cite{OpenStreetMap}). Modified from [BCP-II] (CC BY 4.0).}\label{fig:cam_view}
\end{figure}

The ShipSG dataset was collected through two strategically positioned cameras at the Fischereihafen-Doppelschleuse in Bremerhaven, Germany, aiming to cover a broad view of the lock's entrance and adjacent Weser river area (see Fig.~\ref{fig:cam_view}). 
With a height above water level of 23 meters, these cameras captured the dynamic maritime activities within the port basin, ranging distance up to 400 meters from the cameras, and on a distance of up to 1200 meters on the Weser river.
The images were captured under various weather conditions including sunny, cloudy, windy, and rainy days during Autumn 2020.
The dataset also ensures privacy by anonymizing non-relevant entities like vehicles and people. 
This comprehensive collection approach ensures a diverse and realistic dataset for maritime research.

The dataset utilized \gls{ais} data to identify ships in each image, accessing real-time positional and static information from ships to annotate images accurately. 
This included both the exact locations and lengths of the ships. 
By matching the timestamps of \gls{ais} messages with those of captured images, the dataset ensures high precision in ship positioning, limiting the time offset to 100 ms to achieve a close correspondence.
The recommended speed within the port of Bremerhaven is 10 knots (18.5 km/h) to ensure safe and efficient navigation, in accordance with the guidelines provided by the port authority~\cite{Bremerhaven2024}, which means a displacement of approximately 0.5 m in 100 ms. 
Therefore, the impact of the time offset between \gls{ais} and image capture can be considered negligible for the precision of the dataset ground truth.
This approach allowed for the annotation of 3505 images using accurate \gls{ais} data. 

\begin{figure}[h]
\includegraphics[width=13.5 cm]{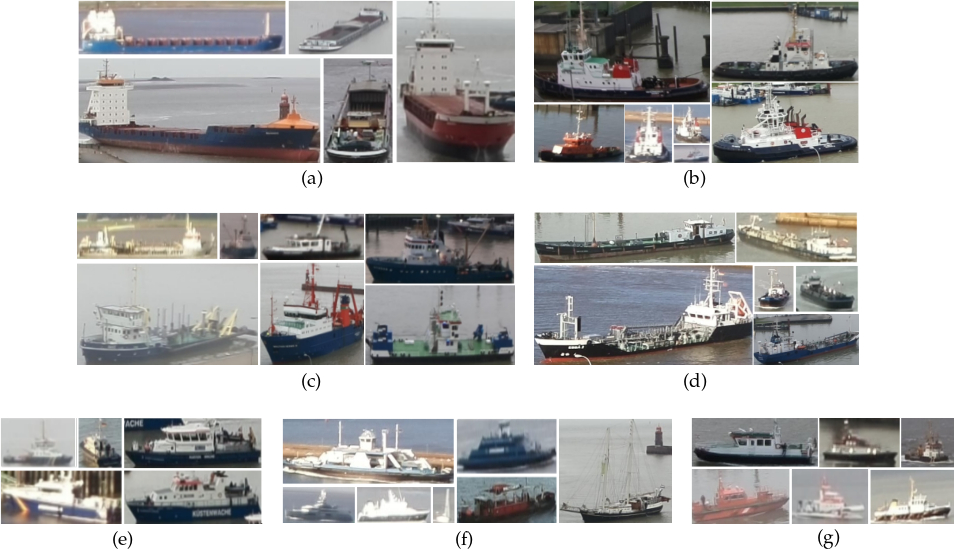}
\centering
\caption[Examples extracted from the dataset that show the seven ship classes]{Examples extracted from the dataset that show the seven ship classes. Each class contains a variety of sizes and orientations of the ships. (a) Cargo, (b) Tug, (c) Special 1, (d) Tanker, (e) Law Enforcement, (f) Passenger/Pleasure, (g) Special 2. Reprinted from~\ref{paper:2} (CC BY 4.0). \label{fig:class_samples}}
\end{figure}  

By providing \gls{ais} ship types with the dataset, we enable users to create their own tailored classes. In our case, we categorized seven ship classes (see Figure~\ref{fig:class_samples}) for the dataset based on an observation of their purpose and visual similarities:

\begin{itemize}
\itemsep0em
\item Cargo: All types of cargo ships.
\item Law Enforcement: Police watercrafts and coast guard ships.
\item Passenger/Pleasure: Ferries, yachts, pleasure and sailing crafts.
\item Special 1: Crane vessels, dredgers and fishing boats.
\item Special 2: Research and survey ships, search and rescue ships and pilot vessels.
\item Tanker: All types of tankers.
\item Tug: All types of tugboats.
\end{itemize}

To train and validate instance segmentation algorithms, ship masks were manually annotated in each image, identifying the ships and their classes using the LabelMe software~\cite{labelme2016}. Moreover, ShipSG can also be used not only for the development of ship segmentation but also ship detection algorithms, by considering the surrounding bounding box of annotated masks.

\section{Summary and Discussion}

Featuring 3505 images, 11625 ship masks and the corresponding georeferences, a novel dataset, ShipSG, for ship segmentation and georeferencing using a static oblique view of a port has been presented. This dataset contains images with mask annotations of ships present, and their corresponding class, position and length.

ShipSG stands as a pivotal contribution to the field of maritime research, setting a new benchmark for ship segmentation and georeferencing. 
The validation of innovative methodologies using ShipSG lays the groundwork for future advancements in maritime situational awareness. 

The creation and use of ShipSG is an essential pillar for this thesis, as it allowed the validation of the recognition methods proposed in Chapters~\ref{chap:ship_rec} and~\ref{chap:adv_ship_rec}.
Our proposed georeferencing methods are also quantitatively validated using ShipSG, as presented in Chapter~\ref{chap:ship_geo}.
In total, ShipSG has been used in publications~\ref{paper:2},~\ref{paper:3},~\ref{paper:5} and~\ref{paper:6}.

While methods trained on ShipSG were cross-validated with other similar datasets in~\ref{paper:2} to study generalizability to other maritime scenes (see Sec.~\ref{sec:std_ship_seg}), a key limitation is its reliance on only two views of the same area, coupled with the high costs and logistical challenges of new image capture and manual annotation. 
Therefore, improvements of the dataset will focus on introducing a broader spectrum of data, crucial for mitigating issues caused by the limited variability in real-world annotated data. 

Moreover, future iterations of ShipSG could enhance ship recognition algorithms by leveraging \gls{ais} data for annotating ship heading, in addition to the ship lengths. 
The incorporation of this, combined with further annotations such as ship cuboids or keypoints, would offer valuable insights into the development of algorithms that automatically recognize ship heading and dimensions.

\chapter[Ship Recognition for Improved Maritime Awareness]{Ship Recognition for Improved Maritime Awareness}
\label{chap:ship_rec}
\glsresetall

We now delve into the initial exploration of deep learning techniques for ship detection and instance segmentation, that allowed further development of tailored solutions for ship recognition in the maritime domain as discussed in subsequent chapters.

Firstly, Section~\ref{sec:ship_det} shows that ship detection serves as a proof of concept for the feasibility of using deep-learning-based object detection and georeferencing. 
This proof of concept reveals its potential to be applied to existing problems, as proposed in \ref{paper:1}, \ref{paper:3} and \ref{paper:4}: abnormal vessel behaviour detection, camera integrity assessment and 3D reconstruction. 
Secondly, the chapter continues with the journey through standard instance segmentation methods shown in~\ref{sec:std_ship_seg}, performed in~\ref{paper:2}, setting the stage for the custom developments for real-time ship segmentation and georeferencing provided in Chapter~\ref{chap:adv_ship_rec} and~\ref{chap:ship_geo}. Therefore, this chapter outlines the impact of ship detection and segmentation in the development of advanced methodologies for the improvement of maritime situational awareness.

\section{Ship Detection for Maritime Applications}
\label{sec:ship_det}

In this section we navigate through the implementations for this thesis in the field of ship detection from monitoring video and images as proposed in \ref{paper:1}, \ref{paper:3} and \ref{paper:4}. 
We explore how the automatic recognition of the bounding box of ships provides information that can be used by further processes for three different applications.
The first application is the detection of abnormal vessel behavior, which is crucial for maritime safety and security, as it enables early identification and mitigation of potential threats or navigational hazards~\cite{riveiro2018maritime}.
The second is the assessment of optical camera obstruction using ship detection, vital to maintain the reliability of surveillance systems, ensuring consistent monitoring quality under various environmental conditions~\cite{de2022partial}.
Finally, the third application discussed in this section is 3D reconstruction of detected ships, which plays a pivotal role in enhancing situational awareness by offering three-dimensional visualizations which improve available semantic information of the situation~\cite{FraunhoferCML2021}.

\subsection[Detection of Abnormal Vessel Behaviour from Video]{Abnormal Vessel Behaviour from Video~\ref{paper:1}}
\label{sec:ship_det_abnormal}

\begin{figure}[h]
\includegraphics[width=13.5 cm]{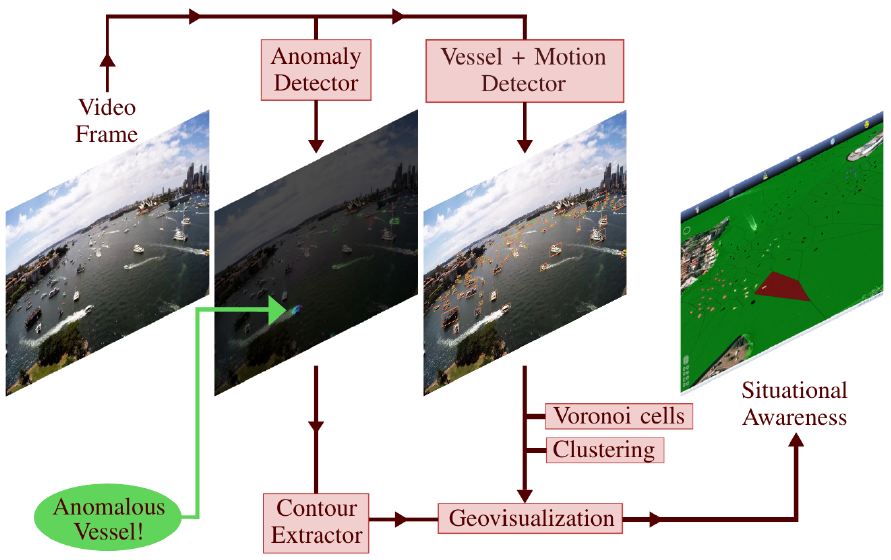}
\centering
\caption[Framework proposed in~\ref{paper:1} for maritime anomaly detection from video.]{Framework proposed in~\ref{paper:1} for maritime anomaly detection from video, including my contributions (Vessel + Motion Detector and Geovisualization). The framework interprets the anomalies using the detections and georeferences for the geovisualization. Reprinted from~\ref{paper:1}. \textcopyright 2021 IEEE. \label{fig:paper1_diagram}}
\end{figure}  

The proof of concept for ship detection and georeferencing as tool to support maritime situational awareness, paving the way for this thesis, was conceived in~\ref{paper:1}.
The publication presents a framework (see Figure~\ref{fig:paper1_diagram}) for detecting and geovisualizing abnormal vessel behavior using video. 
This framework aims to enhance maritime situational awareness by offering a tool that leverages \gls{ai} for monitoring and interpreting anomalous vessel activities, thereby improving safety and security in maritime environments.

\begin{figure}[h]
\includegraphics[width=13.5 cm]{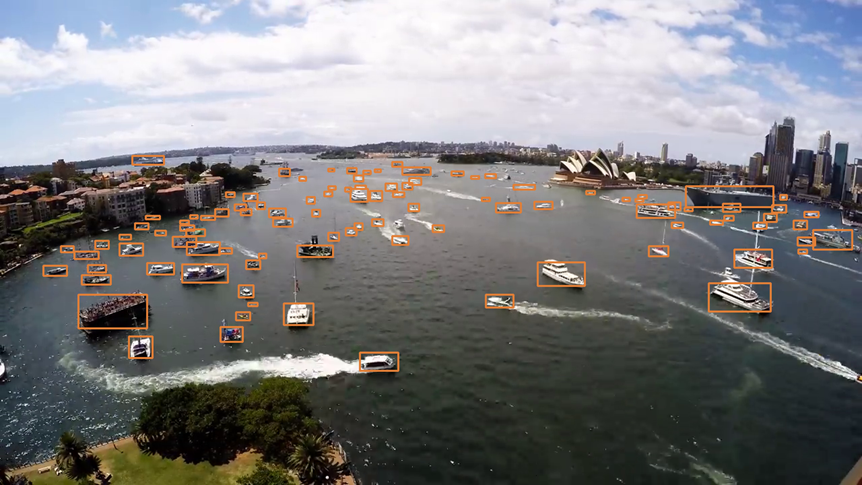}
\centering
\caption[Inference of the YOLOv4-CSP based vessel detector.]{Inference of the YOLOv4-CSP based vessel detector. The orange bounding boxes correspond to the detected vessels. Reprinted from~\ref{paper:1}. \textcopyright 2021 IEEE. \label{fig:formal_dets}}
\end{figure}

In my contribution to \ref{paper:1}, the focus was on ship detection, motion analysis, and georeferencing, to facilitate global geospatial localization and visualization of abnormal behavior (in latitudes and longitudes). 
The georeferenced ships are then represented on a real-world coordinate map, and the motion is used to represent the direction of movement in the form of ship heading. Details for the motion and georeferencing are given in Chapter~\ref{chap:ship_geo}.
The anomaly detection module, which was not part of my contribution, uses a \gls{gan} for the identification of abnormal behavior. The behaviour is interpreted by combining the output of the anomaly detector with the ship detections, motion and georeferencing. The abnormal vessel behavior was defined by categorizing anomalies based on significant vessel fluxes and depletions, thereby establishing a nuanced criteria that captures a wide range of anomalous patterns without relying on supervised training. 
Both anomaly detection and ship detection model were trained and validated on an optical sequence video with a resolution of $1280 \times 720$ pixels and high density of vessels at the port of Sydney. The scene can be seen in Figure~\ref{fig:formal_dets}.

Focusing this section on the object detector used for the framework (motion and georeferencing are explained in~\ref{sec:qual_ship_geo}), I defined a custom dataset from the video using 75 random frames for training and 20 for validation. Then, the ships bounding boxes were annotated manually. 
Given the high density of vessels in each frame, this lead to a total of 4922 and 1387 bounding boxes on the training and validation set, respectively.

The object detector selected was YOLOv4-CSP~\cite{wang2021scaled}, with CSP-Darknet53 backbone~\cite{wang2021scaled}, to train and validate with the generated custom dataset. 
After 234 epochs, the model achieved a peak \gls{map} of 75.86\%.
Figure~\ref{fig:formal_dets} shows the inference of vessels with the resulting model on a frame that was not part of either the training or validation set for the ship detector.
Among the configuration parameters for training, there was the need to increase the image resolution to $1536\times1536$ pixels to obtain meaningful results with small ships or those located far away from the camera (see Figure~\ref{fig:formal_dets}), to the detriment of real-time performance.
The goal of the publication was the proof of concept for a framework for ship detection and georeferencing for the identification of abnormal behaviour.
Therefore, real-time processing was not a concern and all modules were run on high-end servers in an off-line manner.

The vessel detector presented in~\ref{paper:1} plays a key role within the framework. 
It identifies vessels and ships in video data, enabling further analysis such as motion detection through optical flow and accurate mapping of vessel locations using georeferencing. 
This process allows for the identification of vessel movements and anomalies on maps using web services, crucial for improving maritime safety and security.
The methodology of bounding box georeferencing and optical-flow based course calculation, which form a significant part of the contribution of~\ref{paper:1}, are discussed in Chapter~\ref{chap:ship_geo}. 

Moreover, while the presented framework demonstrates promising results in a controlled setting, transitioning to a real-time, real-world application remained unexplored in~\ref{paper:1}. 
This thesis further explores solutions to bridge this gap with regards to the recognition and georeferencing of ships.
As motivated in Chapter~\ref{chap:intro}, georeferencing results are evidently superior when using the mask of ships rather than the bounding box, due to the unnecessary background included within bounding boxes and the inaccuracies of using the bounding box center as the georeferencing point.
Selecting an incorrect pixel from the bounding box for georeferencing introduces more error, leading to the consideration of instance segmentation over object detection in the following chapters of this thesis, aiming for more precise georeferencing.

\subsection[Ship Detection for Integrity Assessment of Camera Obstruction]{Ship Detection for Integrity Assessment of Camera Obstruction~\ref{paper:3}}

The study presented in~\ref{paper:3}, focuses on evaluating the resilience and reliability of maritime object detection algorithms under conditions of partial camera obstruction. 
The work is based around the ShipSG dataset, to explore the effects of various simulated obstructions on detection performance. 
An obstruction is defined as a physical anomaly, mostly static in nature, that obstructs the camera in close proximity to the lens, potentially requiring intervention to remove or clean.
The goal was to quantify the detrimental impact on the system's ability to detect maritime objects correctly, treating the obstruction as a type of fault to investigate its effects on object detection performance.

\begin{figure}[h]
\includegraphics[width=9.5 cm]{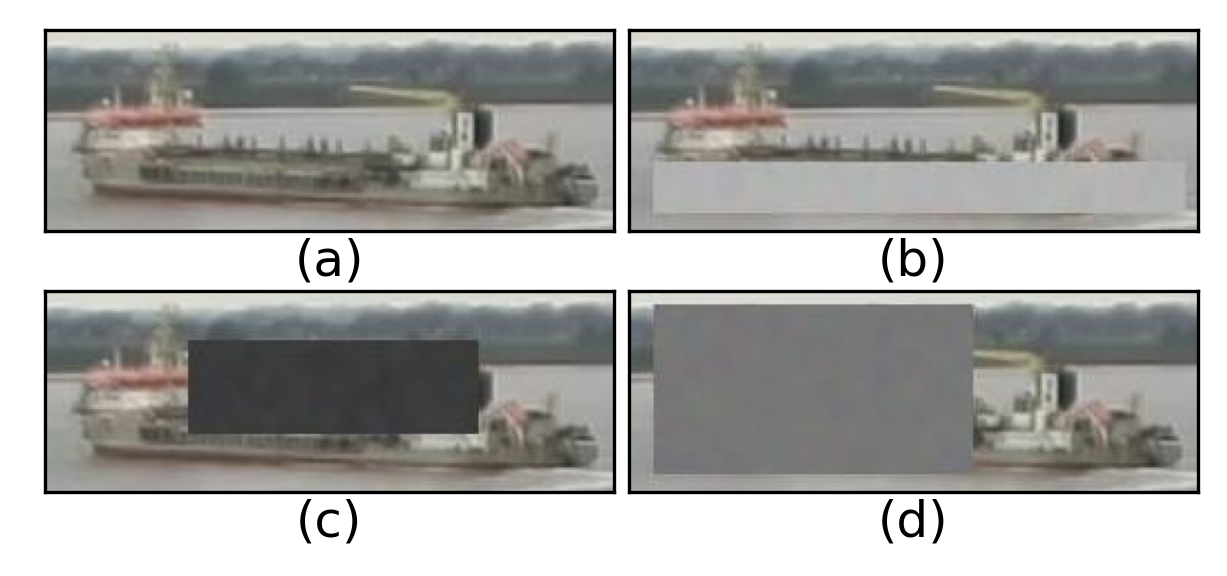}
\centering
\caption[Examples of a ship with different synthetic partial obstruction profiles.]{Examples of a ship with different synthetic partial obstruction profiles. (a) No obstruction; (b) 30\% bright obstruction at the bottom; (c) 30\% dark
obstruction at the center; (d) 60\% gray obstruction at the right. Reprinted from~\ref{paper:3}. \textcopyright 2023 IEEE. \label{fig:obstructions}}
\end{figure}  

My contributions to \ref{paper:3} lie in the use of ShipSG dataset and the detection of ships, that allow an investigation on how the obstructions affect false positive, misclassification, and false negative ratios, along with the detection score distributions.
The work employs Faster R-CNN~\cite{ren2015faster} as the object detector for ship detection. While not specifically designed for real-time applications, its robust architecture offers a strong foundation for object detection tasks.
Faster R-CNN was trained on the seven different ship classes of the dataset, using 80\% of images for training and 20\% for validation, with $1333\times800$ pixel resolution with the ResNeXt-101 backbone~\cite{xie2017aggregated}.
The training was initiated using pre-trained COCO weights~\cite{lin2014microsoft} and after additional training on ShipSG of 11 epochs, the model achieved a \gls{map} of 82.6\%. 

The work of~\ref{paper:3} aligns with the expectations, that true positives (correct detections) decrease and false negatives rise linearly with obstruction. False positives peak from 50 to 60\% obstruction, and beyond 60\%, as the occlusion covers most of the ship, the detector often fails to recognize any object, leading to an increase in false negatives.
Incorporating an obstruction detection step can support maritime stakeholders in identifying camera faults, saving operational time and the subsequent costs. 

Further advancements will employ instance segmentation instead of object detection to offer significant integrity assessment improvements, as it allows for precise delineation of ship contours, minimizing background noise in detections. 
Therefore, segmentation could provide deeper insights into how lens obstructions specifically affect the visibility and classification of ships, by isolating the object from obstructive elements more effectively than bounding boxes.
Moreover, more realistic obstructions should be used to increase the understanding of their impact in the recognition. 

\subsection[Ship Detection for 3D Reconstruction]{Ship Detection for 3D Reconstruction~\ref{paper:4}}
\label{sec:ship_det_3d}

\begin{figure}[h]
\includegraphics[width=13.5 cm]{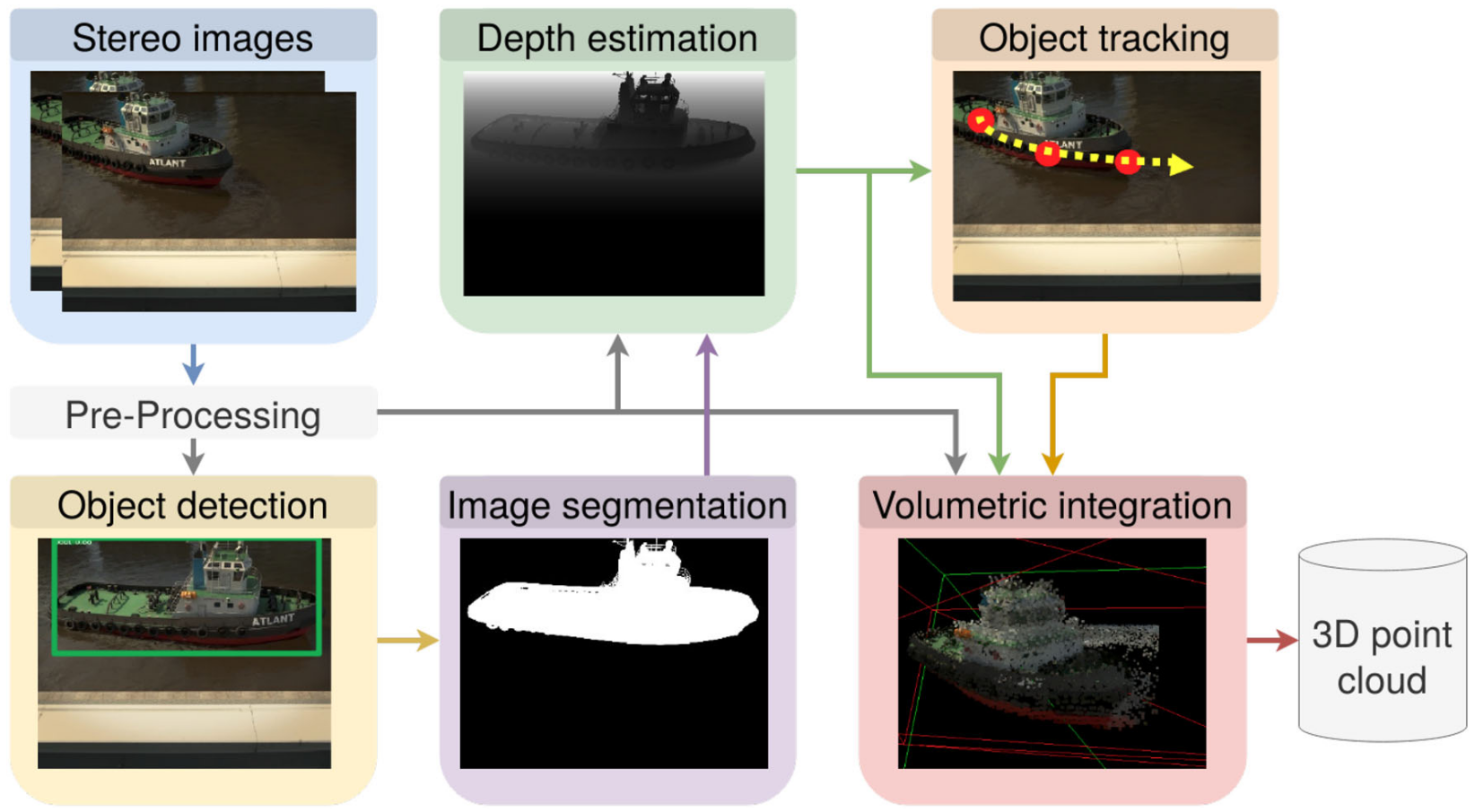}
\centering
\caption[Framework proposed in~\ref{paper:4} for 3D reconstruction of ships using synthetic stereo images.]{Framework proposed in~\ref{paper:4} for 3D reconstruction of ships using synthetic stereo images. Reprinted from~\ref{paper:4}. \textcopyright 2023 Springer. \label{fig:paper4_diagram}}
\end{figure}  

The work in~\ref{paper:4} proposes a novel experimental framework (see Figure~\ref{fig:paper4_diagram}) for real-time 3D reconstruction of a detected ship, using synthetic stereo images, and is deployed on an NVIDIA Jetson AGX Xavier.
The goal is to enhance maritime situational awareness by processing 2D video data for display into a single consistent 3D display using an embedded system. This transformation from 2D videos into 3D displays provides an intuitive, comprehensive maritime environment understanding with enhanced visualization.
The framework is validated using a synthetic and controlled environment created with Blender3D~\cite{blender3D}, that represents a simulated sequence of a tugboat.
It introduces a pipeline prototype for dynamic 3D reconstruction using virtual stereoscopic cameras on a \gls{gpu}-accelerated embedded device, where object detection plays the role of locating the tugboat on the frames. 

\begin{figure}[h]
\includegraphics[width=13.5 cm]{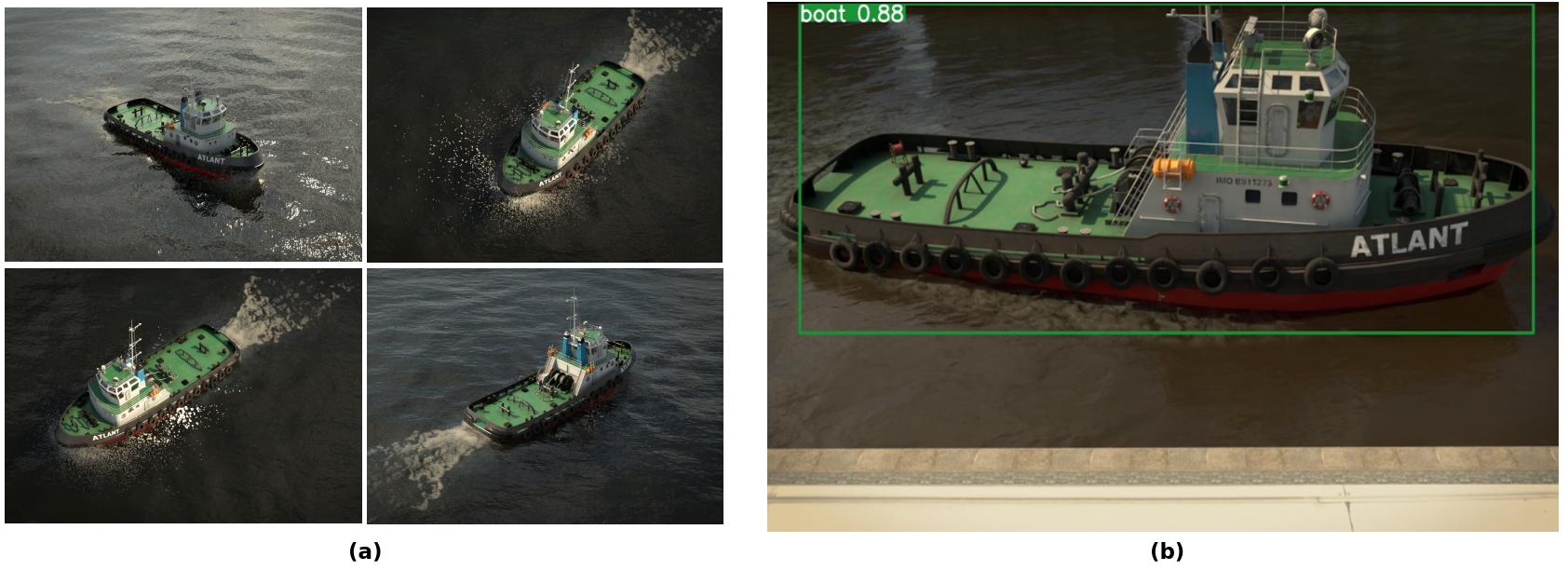}
\centering
\caption[Object detection example for 3D reconstruction.]{Object detection example for 3D reconstruction. (a) Four samples of the rendered dataset for object detection training. (b) An example of tugboat detection using YOLOv5 on one frame of the synthetic sequence for reconstruction. Reprinted from~\ref{paper:4}. \textcopyright 2023 Springer. \label{fig:synthetic_yolov5}}
\end{figure}  

As part of my contributions to~\ref{paper:4}, YOLOv5~\cite{jocheryolov5} was selected as object detector, in its lightest configuration (nano or $n$).
It was trained on a custom synthetic dataset of a tugboat in various sizes and perspectives (see Figure\ref{fig:synthetic_yolov5} (a)), for 50 epochs with an image resolution of $640\times640$ pixels.
For inference, the object detector was deployed, using Pytorch weights, on an NVIDIA Jetson Xavier AGX for real-time processing, using Pytorch~\cite{NEURIPS2019_9015}, achieving a speed of 74 ms per frame and a \gls{map} of 90.7\%. 
The tugboat detector, the fast and accurate detection, enables the framework to focus the rest of the pipeline on the content of the bounding box, therefore supporting the subsequent 3D reconstruction process. 

Testing this framework on a synthetic dataset presents challenges in extrapolating results to real-world scenarios. 
Real-world deployment faces issues such as varied lighting conditions, diverse ship designs, and environmental factors like sea state and weather, which can significantly impact detection and reconstruction accuracy.
We observe that the high \gls{map} (90.7\%) is a result of the dataset just being constituted by the same boat throughout the whole sequence, with no other boats or classes being present. 
Addressing these challenges requires robust algorithmic improvements and real-world datasets to ensure the framework's effectiveness in practical maritime monitoring and safety applications.

Moreover, exploring instance segmentation instead of object detection for future work could yield benefits, particularly for accurate 3D reconstruction, where ship segmentation is essential. 
After the tugboat detector, the framework relies on traditional segmentation techniques, due to the absence of a real-time capable segmentation solution using deep learning deployable on the NVIDIA Jetson AGX Xavier at the time of the implementation of~\ref{paper:4}.
This underscores the potential of real-time instance segmentation developments deployed on embedded systems shown in Chapter~\ref{chap:adv_ship_rec}. 
For example, the framework presented in Chapter~\ref{chap:adv_ship_rec}, a deep-learning-based instance segmentation method could provide a more unified approach, enhancing 3D reconstruction efficiency and potentially accuracy.

\section[Standard Ship Segmentation Using ShipSG]{Standard Ship Segmentation Using ShipSG~\ref{paper:2}}
\label{sec:std_ship_seg}

Building on the foundational work of this thesis in ship detection and its implications for maritime applications, as outlined in the preceding sections, we look now into the experimental evaluation of standard instance segmentation methods on the ShipSG dataset, as presented in~\ref{paper:2}.
The use of instance segmentation is motivated by the significant enhancement that it would provide in applications such as abnormal vessel behavior detection with more accurate georeferencing, integrity assessment of camera obstruction with more accurate analysis, and 3D reconstruction of detected ships with the unification of parts in the pipeline.

The creation of the ShipSG dataset (see Chapter~\ref{chap:shipsg}), provided a comprehensive basis to perform an evaluation of robust instance segmentation methods like Mask R-CNN~\cite{he2017mask} and DetectoRS~\cite{qiao2021detectors}, as well as real-time methods including YOLACT~\cite{bolya2019yolact} and Centermask-Lite~\cite{lee2020centermask}. 
The latter evaluations seeking real-time performance involved configurations that sought to balance inference speed with \gls{map}. 
Therefore, two configurations for each were selected, one deeper and another one lighter, as can be seen in Tables~\ref{tab:std_train} and~\ref{tab:std_result}.
All methods initiated training on ShipSG with COCO pre-trained weights.
It is notable that in~\ref{paper:2}, inference speed was measured using the NVIDIA GV100, a high-end \gls{gpu}, boasts Tensor Cores for \gls{ai} acceleration, 32 GB of memory, and over 5000 \gls{cuda} cores for unparalleled computational performance. 
Embedded system based deployment will be discussed in Chapter~\ref{chap:adv_ship_rec}.

\begin{table}[h] 
\caption[Configurations during training for each instance segmentation method evaluated in~\ref{paper:2}]{Configurations during training for each instance segmentation method evaluated in~\ref{paper:2}.  (CC BY 4.0)\label{tab:std_train}}
\centering 
\footnotesize
\begin{tabular}{c|ccccc}
\toprule
\textbf{Method}	& \textbf{Input Size (Pixel)}& \textbf{Backbone} & \textbf{Number of Epochs} \\
\midrule
Mask R-CNN			& 1333~$\times$~800					& ResNeXt-101				& 11\\[0.1cm]
DetectoRS			&1333~$\times$~800 & ResNet-50 & 11\\[0.1cm]
YOLACT\textsubscript{550}			& 550~$\times$~550 & ResNet-50 & 18\\[0.1cm]
YOLACT\textsubscript{700}			& 700~$\times$~700 & ResNet-101 & 16\\[0.1cm]
Centermask-Lite\textsubscript{V19}		& 800~$\times$~600 & Vovnet-19& 17\\[0.1cm]
Centermask-Lite\textsubscript{V39}	& 800~$\times$~600 & Vovnet-39& 17\\
\bottomrule
\end{tabular}
\end{table}

\begin{table}[h]
\caption[Resulting instance segmentation APs and inference speed per initial method evaluated. ]{Resulting instance segmentation APs and inference speed per method evaluated. Inference times are measured on a high-end NVIDIA GV100 \gls{gpu}. Adapted from [BCP-II] (CC BY 4.0).\label{tab:std_result}}
\footnotesize
\begin{tabular}{c|cccccccc}
\toprule
\textbf{Method}	& \textbf{mAP (\%)} & \textbf{mAP\textsubscript{s} (\%)} & \textbf{mAP\textsubscript{m} (\%)} & \textbf{mAP\textsubscript{l} (\%)} & \textbf{Inference (ms)}\\
\midrule
Mask R-CNN&73.3&50.3&75.2&77.2&117\\[0.1cm]
DetectoRS&74.7&55.6&75.7&79.2&151\\[0.1cm]
\midrule
YOLACT\textsubscript{550}&52.7&8.6&51.5&70.9&28	\\[0.1cm]
YOLACT\textsubscript{700}&58.2&14.0&58.2&75.1&36 \\[0.1cm]
Centermask-Lite\textsubscript{V19}&63.5&45.5&64.0&65.7&24	 \\[0.1cm]
Centermask-Lite\textsubscript{V39}&64.4&46.1&64.8&66.1&28	 \\
\bottomrule 
\end{tabular}
\end{table}

As shown in Table~\ref{tab:std_result}, the robust methods, Mask R-CNN and DetectoRS, demonstrated superior mask \gls{map} across all categories when compared to their real-time counterparts. DetectoRS, in particular, achieved the highest overall mAP, underscoring its effectiveness in accurate ship segmentation, however at the highest computational cost, even when using a high-end server.

For real-time applications, Centermask-Lite showcased better \gls{map} performance, especially in handling small and medium-sized objects, while YOLACT was more adept at segmenting larger objects. 
However, we observe that small-sized objects are segmented with a significant lower performance than the rest of object sizes for all methods. 
As explained in Chapter~\ref{chap:intro}, small ship segmentation a critical problem in maritime monitoring and it will be tackled by this thesis in Chapter~\ref{chap:adv_ship_rec}.

While Centermask-Lite in its deeper form exhibits the best trade-off in \gls{map} and inference speed, deployment of instance segmentation on an embedded system, remained open. 
This fact highlights the ongoing challenge of optimizing for both speed and accuracy in maritime object detection and segmentation.
The challenge stemmed from framework incompatibilities with \gls{gpu}-powered embedded systems, such as the NVIDIA Jetson AGX Xavier. Notably, this system utilizes an architecture based on ARM (Advanced Reduced Instruction Set Computing Machine)~\cite{sloss2004arm}.\glsunset{arm}
Memory constraints and the \gls{arm}-specific architecture compounded the difficulty of deploying Centermask-Lite.
The effective deployment of our custom real-time ship segmentation on the edge using an embedded system to fill this gap is shown in Chapter~\ref{chap:adv_ship_rec}.

\begin{figure}[h]
\centering
\includegraphics[width=13.5 cm]{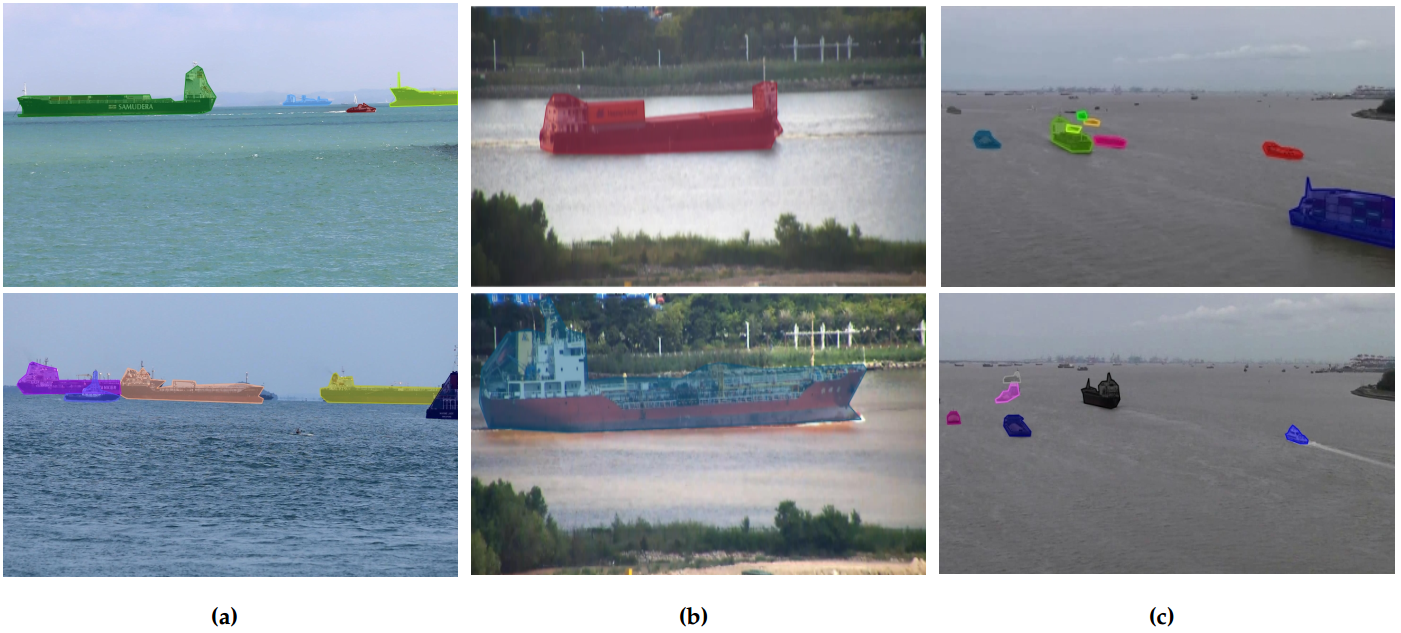}
\caption[Annotated masks on existing datasets to study the generalization of our models.]{Annotated masks on existing datasets to study the generalization of our models.
(a) Annotated examples of the \gls{smd}. (b) Annotated examples of Seaships7000. (c) Annotated examples of the dataset by Chen et al. Reprinted from [BCP-II] (CC BY 4.0)}
\label{fig:generalization_samples}
\end{figure}

The generalization capability of models trained on ShipSG is crucial for deploying these models in diverse real-world maritime environments, where conditions and scenarios can vary significantly.
To assess the generalizability of the instance segmentation models presented in this section, I tested their performance on a mini-dataset of 100 images derived from other maritime datasets, namely the \gls{smd}~\cite{prasad2017video}, Seaships7000~\cite{shao2018seaships}, and the dataset by Chen et al~\cite{chen2020video}. 
These datasets, which only used for testing in this work, provide a diverse range of maritime scenarios and vessel types to challenge the ability of the models to accurately segment ships in different conditions.
Since these datasets did not contain mask annotations, ship masks were annotated on the 100 images manually (see Figure~\ref{fig:generalization_samples}).
With DetectoRS leading the \gls{map} with 48.6\%, the test revealed that models trained on the ShipSG dataset could predict ships from other datasets with reasonable accuracy, given the complex diversity of the mini-dataset. 

The work presented in this section underpins the importance of instance segmentation in enhancing maritime safety and security applications. 
By providing a detailed evaluation of different segmentation methods, it paves the way for the next research focus on optimizing instance segmentation models for real-time applications while deployed on an embedded system, which is tackled in Chapter~\ref{chap:adv_ship_rec}. 
Moreover, georeferencing from the resulting masks of the evaluated methods of this section is quantitatively analyzed in Chapter~\ref{chap:ship_geo}.

\section{Summary and Discussion}

This chapter showcased the initial exploration in applying deep learning techniques for ship detection and instance segmentation within maritime applications for the improvement of situational awareness, presented in \ref{paper:1}, \ref{paper:2}, \ref{paper:3} and \ref{paper:4}. 
Along the chapter, the pivotal role of ship detection in facilitating a range of applications has been highlighted, from abnormal vessel behavior detection or camera integrity assessment to 3D ship reconstruction. 
Moreover, it underscored the enhanced capabilities that instance segmentation offers over bounding box detection, particularly in extracting detailed ship features crucial for applications like georeferencing, which is specially interesting for this thesis to improve maritime situational awareness.

Despite the demonstrated potential of bounding box ship detection and success in controlled~\ref{paper:1} or synthetic~\ref{paper:4} settings, several challenges and open tasks remain. 
For example, the improvement of georeferencing and integrity assessment using segmented masks instead of bounding boxes or to unify detection and segmentation in the case of our 3D reconstruction framework.
Therefore, this opens the way to use instance segmentation instead of bounding box detection.

In the initial exploration of standard instance segmentation techniques on the ShipSG dataset, we studied the precision of ship feature extraction, crucial for the various maritime applications. 
This highlights the challenges associated with real-time processing. 
The evaluation of real-time methods on the NVIDIA GV100 \gls{gpu} revealed that the best trade-off between computational efficiency and segmentation accuracy, was given by Centermask-Lite.

Though deployment of YOLOv5 (bounding box) using Pytorch weights is reported in~\ref{paper:4} (see Sec.~\ref{sec:ship_det_3d}), the deployment of instance segmentation on \gls{gpu}-powered embedded systems was not reported in \ref{paper:2} (see Sec.\ref{sec:std_ship_seg}). 
Deployment challenges arose from the incompatibility between deep learning and the \gls{arm} architectures of \gls{gpu}-powered embedded systems like the NVIDIA Jetson AGX Xavier.  
This highlights the need for adaptable methods to enable advanced on-board instance segmentation processing. 
The move towards embedded systems is essential for practical deployment in dynamic maritime environments, where processing speed and accuracy are paramount.
This transition to real-time instance segmentation on embedded systems is addressed in Chapter~\ref{chap:adv_ship_rec}.

Another critical aspect discussed is the importance of instance segmentation for accurate georeferencing of ships. We motivated in Chapter~\ref{chap:intro} that while ship detection provides valuable insights for several maritime applications, instance segmentation offers a more detailed analysis crucial for precise georeferencing. 
The ability to extract exact ship contours rather than relying on bounding boxes allows for more accurate positioning of vessels. 
This capability is explored further in Chapter~\ref{chap:ship_geo}, which delves into the application of the methods proposed for improving maritime situational awareness through enhanced georeferencing.

Furthermore, the initial instance segmentation study shown in this chapter, while promising, highlighted a precision decrease in segmenting small or distant ships. 
This issue is particularly pertinent for maritime situational awareness, where the ability to accurately identify all vessels, independent their size and within the proximity of the port area is crucial. 
Chapter~\ref{chap:adv_ship_rec}, Section~\ref{sec:small_ship_rec}, addresses this by introducing a solution that enhances the segmentation of small ships, thereby filling this gap in the initial methodology.

In essence, while the initial exploration into ship detection and instance segmentation revealed significant potential for enhancing maritime situational awareness, it also uncovered several challenges and areas for further development. 
Specifically, the need for real-time processing on embedded systems, improved detection of small or distant ships, and the utilization of instance segmentation for accurate georeferencing. 
The subsequent chapters of the thesis aim to address these gaps, presenting custom-tailored solutions that bring these advanced computer vision techniques closer to practical deployment in the maritime domain.
\chapter{Advanced Ship Recognition for Real-time Operation}
\label{chap:adv_ship_rec}
\glsresetall
\glsunset{cuda}
\glsunset{gpu}
In Chapter~\ref{chap:ship_rec}, we explored the impact of ship detection and segmentation in the improvement of maritime situational awareness and how the development of advanced methodologies can further improve results.
Moreover, the deployment of such algorithms on embedded systems has been proven important for practical deployment in dynamic maritime environments, where processing speed and accuracy are paramount.
We discussed the applicability of ship detection when deployed on an embedded system, as reported in~\ref{paper:4}.
However, in the case of instance segmentation, task required for more accurate ship georeferencing, deployment on \gls{gpu}-powered embedded systems remained open. The deployment of instance segmentation methods highlighted the need for adaptable methods to enable advanced on-board processing. 

We investigate in this chapter the proposed improvements for real-time ship segmentation proposed in this thesis, as introduced in~\ref{paper:5}, \ref{paper:6}. 
First, I present the ScatBlock, a 2D Scattering-transform-based block to be used in the proposed tailored deep-learning architecture.
Second, I delve into the design of the custom architecture, ScatYOLOv8+CBAM, that integrates the ScatBlock and attention mechanisms, and demonstrate its superior performance using ShipSG.
Thirdly, I propose an optimization to the architecture, followed by the deployment with TensoRT on the embedded system to measure inference times for real-time applicability.
Lastly, I address and propose a solution to the precision decrease in segmenting small and distant ships discussed in Chapter~\ref{chap:ship_rec} and essential for improved maritime situational awareness.

\section[The ScatBlock]{The ScatBlock~\ref{paper:5}}
\label{sec:scatblock}

The ScatBlock, introduced in~\ref{paper:5}, is a custom designed 2D scattering-transform-based block that is key for the novel and tailored architecture for ship segmentation developed in this thesis.
The 2D scattering transform is a specialized operator that extracts invariant feature representations by decomposing the input image data into a set of scattering coefficients. 
Each coefficient is a translation-invariant feature map representation that captures spatial and angular variations in an image. 

Mathematically, the scattering transform is computed using a set of dilated and rotated versions of a mother wavelet $\psi$ and a low-pass filter $\phi_{J}$, with $J$ being the spatial scale of the transform.
The process involves convolving the input image with a predefined filter bank, followed by an element-wise complex modulus operation:

\begin{equation}
U_{\lambda} = \left| (x * \psi_{\lambda}) \right|
\end{equation}

where $x$ represents the original input image, and $\psi_{\lambda}$ denotes the mother wavelet filter at a specific scale and orientation determined by \(\lambda\). The output is obtained with a smoothing operation using the low-pass filter \(\phi\):

\begin{equation}
    S_{\lambda} = U_{\lambda} * \phi_J
\end{equation}

where \(S_{\lambda}\) are the scattering coefficients after the smoothing operation, which capture the invariant and descriptive features of the original image. 
The total number of scattering coefficients (feature maps), $J\times L+1$, is determined by $L$, the number of orientations or rotations of the mother wavelet $\psi$, and $J$ the scale, or also known as order.

\begin{figure}[h]
\includegraphics[width=13 cm]{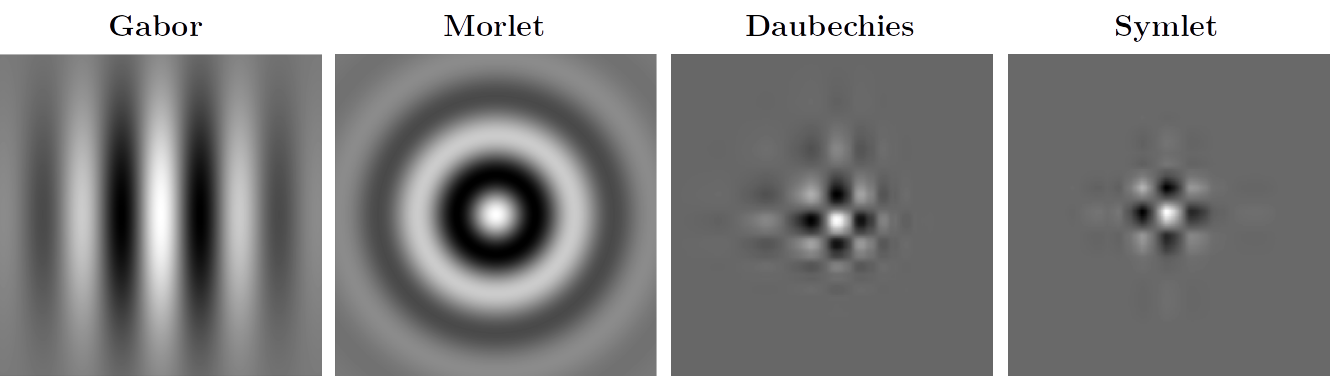}
\centering
\caption[Examples of commonly used 2D wavelets.]{Examples of commonly used 2D wavelets. From left to right: Gabor, Morlet, Daubechies and Symlet wavelets.}\label{fig:wavelets_examples}
\end{figure}  

Commonly used 2D wavelets~\cite{guo2022review} are represented in Figure~\ref{fig:wavelets_examples}. When used in the 2D scattering transform, Gabor wavelets, though computationally expensive, are very sensitive to spatial frequencies and variations in textures. Morlet wavelets are characterized by their sinusoidal shape and can perform better with periodical patterns. Daubechies wavelets, can be useful for images with specific geometric patterns. Lastly, Symlet wavelets offer symmetry to preserve features and minimize distortion, which prevents artifacts in the scattering coefficients. 

\begin{figure}[h]
\includegraphics[width=15.5 cm]{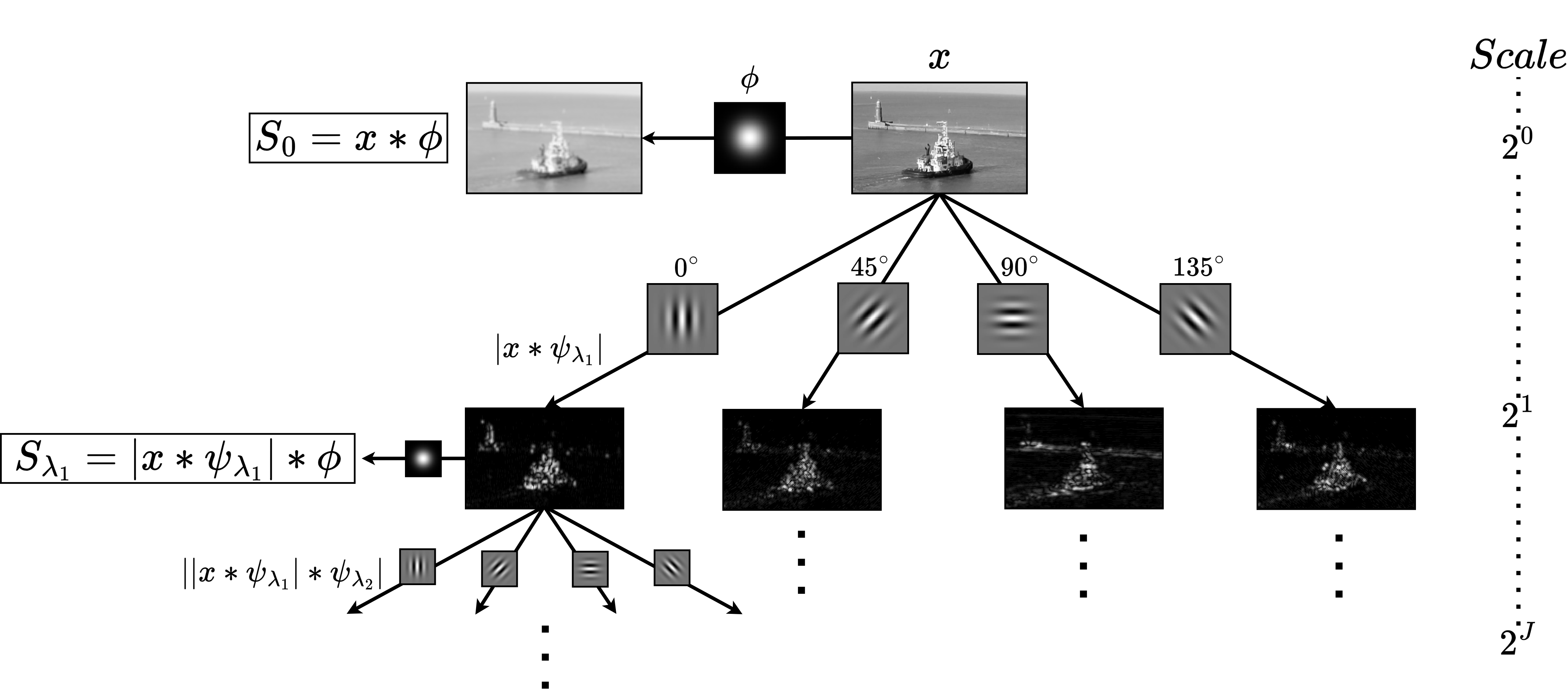}
\centering
\caption[Scattering coefficient decomposition of an image.]{Scattering coefficient decomposition of an image $x$, showing low-pass filtering to obtain $S_0$ and wavelet modulus operations at orientations 0°, 45°, 90°, and 135° for scale $2^1$ to produce first-order coefficients $S_{\lambda_1}$. Higher-order coefficients $S_{\lambda_1,\lambda_2,...}$ are obtained at scale $2^J$.}\label{fig:scat_decomposition}
\end{figure}  

An illustrative example of the multi-scale decomposition in a scattering network is shown in Fig.~\ref{fig:scat_decomposition}. The input image $x$ is subjected to a low-pass filtering to produce $S_0$, and to $Gabor$ wavelet convolutions with four orientations at the first scale $2^1$. The modulus of each convolved image is taken, followed by another low-pass filtering to yield the first-order scattering coefficients $S_{\lambda_1}$. This process is repeated iteratively to produce higher-order coefficients $S_{\lambda_1,\lambda_2}$, with wavelet convolutions at increasing scales $2^J$, capturing progressively coarser image features. The network cascades through multiple scales to extract robust, invariant features for image analysis.
We observe that each scattering coefficient is a translation-invariant feature map representation that captures spatial and angular variations in the input image. 

\begin{figure}[h]
\includegraphics[width=6.5 cm]{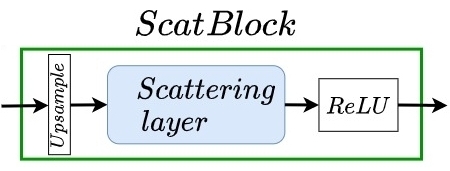}
\centering
\caption[ScatBlock as conceived in~\ref{paper:5}]{ScatBlock, as conceived in~\ref{paper:5}, contains an upsample operation, followed by the scattering layer and a \gls{relu} activation. Adapted from~\ref{paper:5}. \textcopyright 2023 IEEE.}\label{fig:scatblock}
\end{figure}  

To incorporate the scattering transform, the ScatBlock architecture (see Fig.~\ref{fig:scatblock}) begins by upsampling the input image to counteract the resolution decrease inherent to the 2D scattering transform, which typically downsamples the input image by a factor of $2^J$ to reduce computational complexity across scales. 
Specifically, the input image $x$ is upscaled to $(2 \times H) \times (2 \times W)$, ensuring the output dimensionality aligns with the input image size, a requirement for subsequent processing layers in the backbone.
These sparse feature maps resulting from the scattering transform are then forwarded to a \gls{relu} activation function.
The ScatBlock conceived in~\ref{paper:5} uses the first-order coefficients ($J = 1$).
Although the computation of subsequent orders is achievable by incorporating more layers, the first-order coefficients hold significant information for mainstream tasks~\cite{bruna2013invariant}.
In this case, the first order was found to be sufficient to enrich the capability of the model to discern and segment ships while maintaining fast computation time.

Regarding implementation, the ScatBlock developed in~\ref{paper:5} uses the approach developed in reference~\cite{cotter_2020} to achieve the scattering transform, using the open-source Python module $pytorch\_wavelets$\footnote{\url{https://github.com/fbcotter/pytorch\_wavelets/}}.
The scattering transform implementation in $pytorch\_wavelets$ is based on the \gls{dtcwt}. The \gls{dtcwt}, initially introduced by~\cite{selesnick2005dual}, computes the scattering transform with enhanced efficiency while ensuring theoretical consistency with the traditional approach. The \gls{dtcwt}, employs wavelet trees for signal decomposition in frequencies, facilitating information capture and directional selectivity. 
This decomposition allows for a detailed analysis of the isolated frequency content across both horizontal and vertical dimensions of the image, revealing textures and patterns. The approach proposed by~\cite{cotter_2020} in $pytorch\_wavelets$ to achieve the scattering transform not only enhances efficiency on \gls{gpu}s due to the \gls{dtcwt} fast convolution capabilities and suitability for parallel processing, but also ensures the robustness and invariance of the extracted scattering coefficients. A detailed technical description of how the \gls{dtcwt} is used for a faster scattering transform can be found in~\cite{cotter_2020}.

Another option to implement the 2D scattering transform is the package Kymatio~\cite{andreux2020kymatio}, which closely mirrors the traditional scattering transform in its approach but falls short in computational speed compared to $pytorch\_wavelets$. The use of \gls{dtcwt} by the latter for \gls{gpu}-optimized computations significantly accelerates performance, making $pytorch\_wavelets$ the preferred choice for the scattering transform implementations of the ScatBlock due to its efficiency.

As introduced in Chapter~\ref{chap:sota}, the work proposed in \cite{bruna2013invariant} demonstrated that the features extracted by the scattering transform are quite meaningful for \glspl{cnn}. 
The motivation drawn in~\cite{bruna2013invariant} lies in the ability of the transform to provide a deep systematic understanding of how invariant features can be captured and utilized for improved deep-learning-based image classification. 
This approach is particularly relevant for maritime awareness, where the recognition of ships across diverse sizes and types, under varying lighting and weather conditions, demands a robust feature extraction mechanism that can handle the complexities of real-world scenarios. 
The scattering transform using wavelets are particularly suited for ship recognition because the output coefficients excel at capturing multi-scale geometric and structural features, and the relatively uniform water background provides a good contrast for highlighting ships against the static background.
Having this in mind, the ScatBlock has been designed to capture the shape of ships by extracting their inherent geometric and structural properties, and will be added to the custom architecture for ship recognition explained in the following section.

\section[ScatYOLOv8+CBAM]{ScatYOLOv8+CBAM~\ref{paper:5}}
\label{sec:scatyolo}

To conform the customized ship segmentation architecture presented in~\ref{paper:5}, named ScatYOLOv8+CBAM, two additions were implemented.

Firstly, the ScatBlock was blended (Section~\ref{sec:scatblock}) at the beginning of the backbone of YOLOv8 to enhance the input image for instance segmentation, replacing the first convolutional block of YOLOv8. 
This was motivated by~\cite{oyallon2018compressing}. In their work, the authors explore the efficiency of using the scattering transform to preprocess images before feeding them into \gls{cnn}s, showcasing how this technique can significantly enhance the quality of feature representations. The relevance of the findings of~\cite{oyallon2018compressing} to the ScatBlock lie in the practical application of the scattering transform to streamline the processing pipeline for real-time instance segmentation and object recognition tasks.
The advantage of YOLOv8-like architectures is their deployability on \gls{gpu}-powered embedded systems, which enables deployment of the custom architecture as well.

\begin{figure}[h]
\includegraphics[width=6.5 cm]{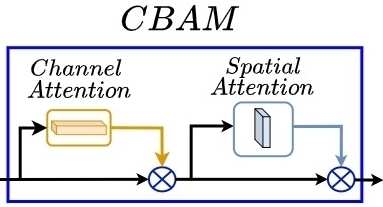}
\centering
\caption[CBAM as conceived in~\ref{paper:5}]{\gls{cbam} module introduced by~\cite{woo2018cbam}, depicting the channel and
spatial attention mechanisms. Adapted from~\ref{paper:5}. \textcopyright 2023 IEEE.}\label{fig:cbam}
\end{figure} 

\begin{figure}[h]
\includegraphics[width=13 cm]{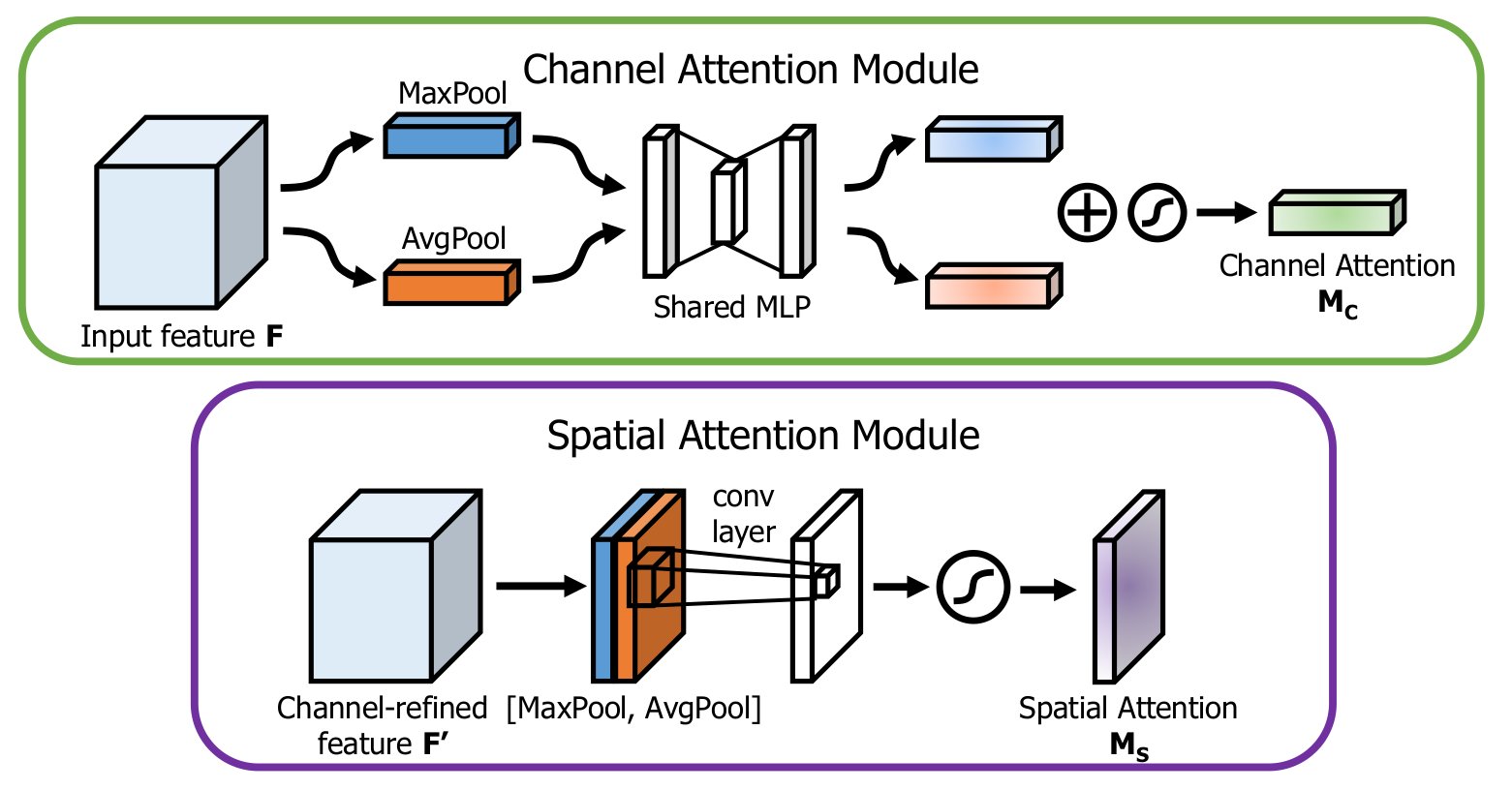}
\centering
\caption[Diagram of each attention sub-module in \gls{cbam}.]{Diagram of each attention sub-module in \gls{cbam}, taken from~\cite{woo2018cbam}. The channel attention module (top) uses both max-pooling and average-pooling operations, followed by a shared \gls{mlp}, to generate channel attention. The spatial attention module (bottom) applies max-pooling and average-pooling across the channels, then uses a convolutional layer to generate spatial attention.}\label{fig:cam_sam_cbam}
\end{figure} 

The second addition to the network is the \gls{cbam}. Introduced by~\cite{woo2018cbam}, this module advances \gls{cnn}s by embedding attention mechanisms with minimal computational overhead. It is structured around two key components (see Fig. \ref{fig:cbam}). The channel attention module, being the initial component, discerns the importance of each feature channel, effectively determining ``what" is significant in the given feature map and enhancing these channels while suppressing less relevant ones. This is achieved using global average pooling and max pooling, which aggregate information efficiently without heavy computation. The channel attention map $M_c$ is computed as follows:
\begin{equation}
M_c(F) = \sigma(W_1 \cdot (\text{ReLU}(W_0 \cdot \text{AvgPool}(F))) + W_1 \cdot (\text{ReLU}(W_0 \cdot \text{MaxPool}(F))))
\end{equation}
where $F$ is the input feature map, $\text{AvgPool}(F)$ and $\text{MaxPool}(F)$ are the global average pooled and global max pooled feature maps, $W_0 \in \mathbb{R}^{C/r \times C}$ and $W_1 \in \mathbb{R}^{C \times C/r}$ are the shared weight matrices of the fully connected layers, \(\text{ReLU}\) is the Rectified Linear Unit activation function, and \(\sigma\) is the sigmoid activation function.

The subsequent component, the spatial attention module, focuses on crucial spatial regions within the feature map, identifying ``where" the important information is located. It applies average pooling and max pooling across the channels, followed by a convolutional layer, which is computationally light. The spatial attention map $M_s$ is computed as follows:
\begin{equation}
M_s(F) = \sigma(f^{7 \times 7}([\text{AvgPool}(F); \text{MaxPool}(F)]))
\end{equation}
where $F$ is the input feature map, $\text{AvgPool}(F)$ and $\text{MaxPool}(F)$ are the average pooled and max pooled feature maps across the channel dimension, $[;]$ denotes the concatenation operation along the channel axis, $f^{7 \times 7}$ represents a convolution operation with a filter size of $7 \times 7$, and $\sigma$ is the sigmoid activation function.

By sequentially applying these attention mechanisms, \gls{cbam} refines the feature representation. The efficiency of \gls{cbam} comes from its simple yet effective design that avoids complex operations, ensuring fast processing and minimal computational overhead. Since the attention maps of \gls{cbam} are broadcast and applied element-wise, they do not alter the height, width, or number of channels of the input feature map.

Through the fusion of channel and spatial attention mechanisms with the YOLOv8 backbone and the 2D scattering transform, the \gls{cbam} enables the network to concentrate on pertinent spatial areas while highlighting critical channels, enhancing feature depiction and localization accuracy. According to~\cite{woo2018cbam}, integrating \gls{cbam} into various deep learning frameworks has yielded notable enhancements across a wide array of classification and detection tasks. Specifically, in tasks related to instance segmentation, the \gls{cbam} has been instrumental in refining object perimeters and improving the precision of segmented individual objects in images~\cite{woo2018cbam}.
The integration of \gls{cbam} into ScatYOLOv8+CBAM at the head enhances feature extraction through both channel and spatial attention mechanisms. These mechanisms leverage scattering transforms to concentrate on significant regions and features, thereby improving the capability to precisely segment ships within intricate maritime settings.

\begin{sidewaysfigure*}
\centering
\includegraphics[width=\textwidth]{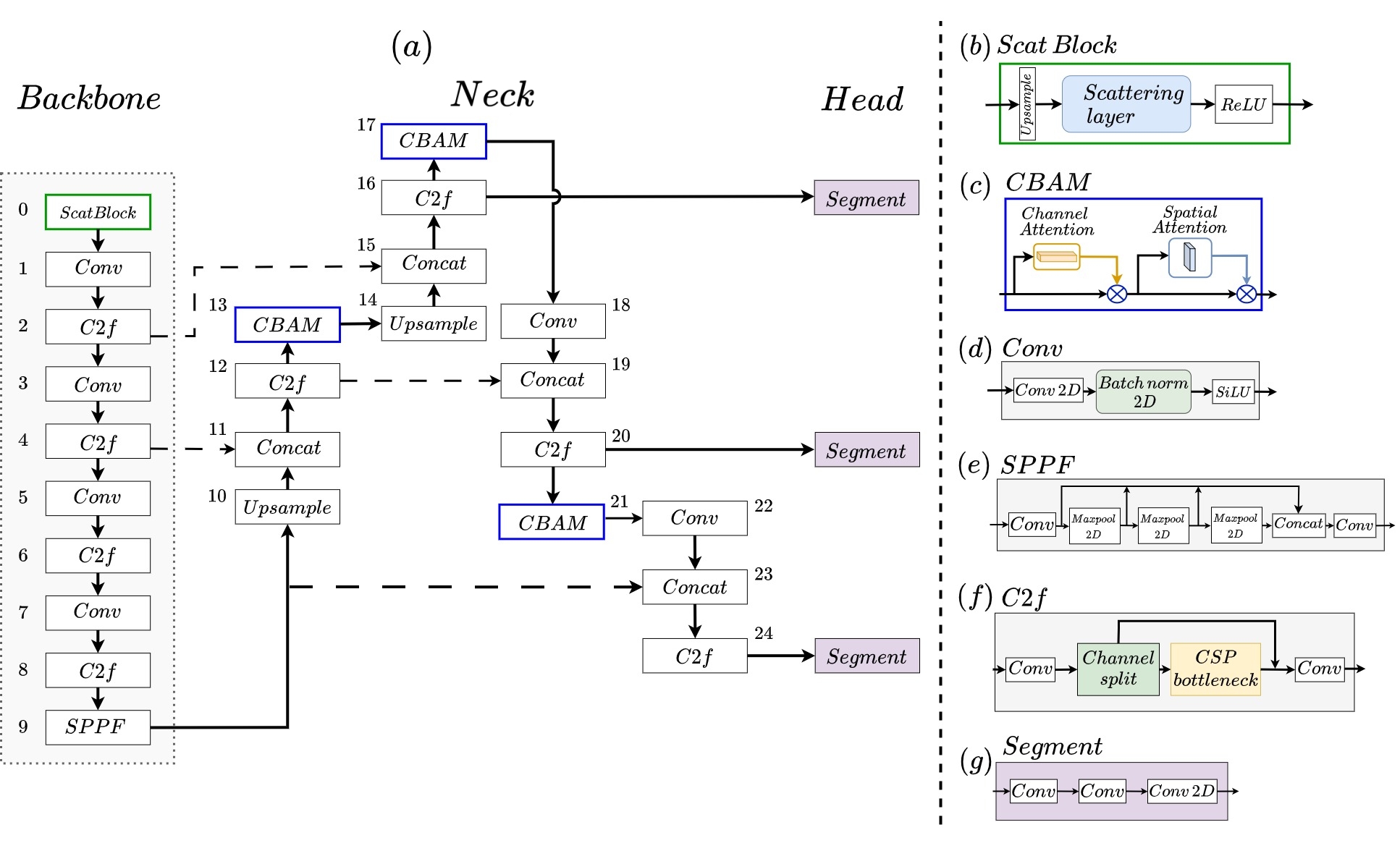}
\caption[The proposed architecture in~\ref{paper:5}.]{\footnotesize The proposed architecture in~\ref{paper:5}, with the ScatBlock in the backbone of YOLOv8 and CBAM modules in the head. (a) We place ScatBlock at the beginning of the YOLOv8 backbone to render a set of sparse feature maps, replacing the first Conv block of the original YOLOv8 backbone. The CBAM module is placed at the output of the head blocks of YOLOv8 to distill valuable object shape information for the segmentation task. The numbers next to every block represent the sequential order followed by the implementation, from input to output (b) The scattering block contains an upsample operation, followed by the scattering layer and a \gls{relu} activation (c) CBAM module depicting the channel and spatial attention mechanisms\cite{woo2018cbam} (d) Standard YOLOv8 convolutional block (e) Spatial Pyramid Pooling Fast module of YOLOv8 (f) C2f with split channel operation and a CSP Bottleneck. (g) Segment block of YOLOv8 that performs segmentation. Reprinted from \ref{paper:5}. \textcopyright 2023 IEEE.}\label{fig:scatyolov8+cbam}
\end{sidewaysfigure*}

The proposed ScatYOLOv8+CBAM, as illustrated in Figure~\ref{fig:scatyolov8+cbam}~(a), substitutes the initial convolutional block of the YOLOv8 backbone with a ScatBlock. Unlike the original first $Conv$ block of CSPDarknet53~\cite{bochkovskiy2020yolov4} (YOLOv8 backbone), the ScatBlock employs the first-order 2D scattering transform. 
The ScatBlock is operated only in forward mode, since the rotated wavelets and low-pass filter are fixed, and it does not allow for the backpropagation and filter parameter updates during the training phase. 
Following the insights from~\cite{zhu2021tph}, which applied CBAM to enhance the head of YOLOv5 for aerial object detection, the CBAM block is integrated after the C2f blocks (see C2f block explanation in Section \ref{sec:sota_ship_rec}) within the YOLOv8 neck. This integration serves a dual purpose: assisting the network in identifying areas of interest, specifically ships, and utilizing these identified regions as inputs for subsequent blocks in the head.

\begin{figure}[h]
\includegraphics[width=\columnwidth]{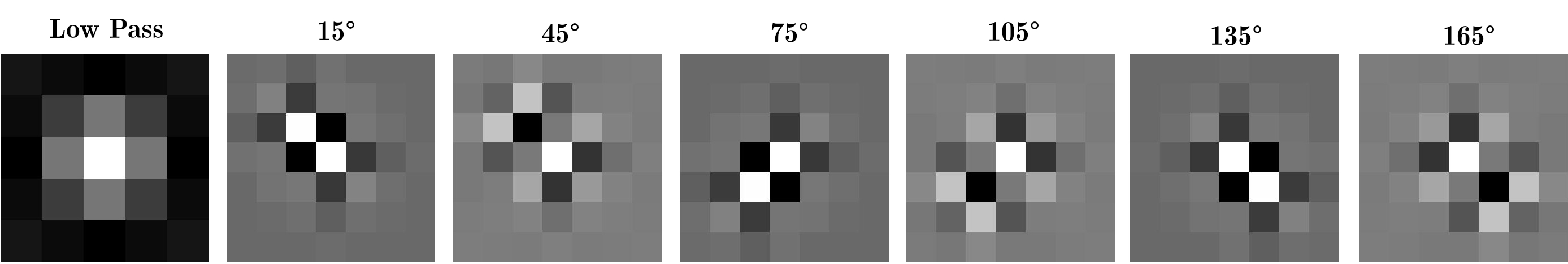}
\centering
\caption[Visualization of 2D Wavelet Filters used in the ScatBlock]{Visualization of 2D Wavelet Filters used in the ScatBlock. The first on the left shows the low-pass filter. The rest display the oriented filters, generated by combining and rotating to represent different directional sensitivities of the Symlet wavelet.}\label{fig:filters_plot}
\end{figure}  

As explained in Section~\ref{sec:scatblock}, the implementation of the ScatBlock (see block number $0$ of Figure~\ref{fig:scatyolov8+cbam}~(a)) uses the 2D Wavelet transformations by~\cite{cotter_2020}, with the open-source Python module $pytorch\_wavelets$, which has \gls{cuda} support for \gls{gpu} operations.

With regards to feature map resolution and channel number changes in the proposed architecture, changes in resolution occur primarily due to the Conv blocks, which reduce the height and width, and the Upsample blocks, which increase them. The other blocks (ScatBlock, C2f, SPPF, CBAM, and Segment) do not alter the height and width resolution. Regarding the number of channels, Conv blocks increase the number of channels, while CBAM blocks decrease them. The ScatBlock increases the number of channels (output scattering coefficients). The SPPF and CBAM block maintain the number of channels. The Concat blocks increase the number of channels by merging feature maps, and the Segment block adjusts the number of channels to match the number of output classes. 

To evaluate ScatYOLOv8+CBAM, in \ref{paper:5}, the ShipSG dataset was used.
Following the common practice in the field and for comparison with the results obtained in Section~\ref{sec:std_ship_seg}, \gls{map} was reported as performance metric. 
For a fair comparison with the state-of-the-art, YOLOv8 and ScatYOLOv8+CBAM models were trained using an NVIDIA A100 \gls{gpu} with random weight initialization for all models. The number of training epochs was 300, with the default settings provided by YOLOv8~\cite{jocheryolov8}. The input size used for all models is $640\times640$ pixels. 

\begin{figure}[h]
\centering
\includegraphics[width=1\columnwidth]{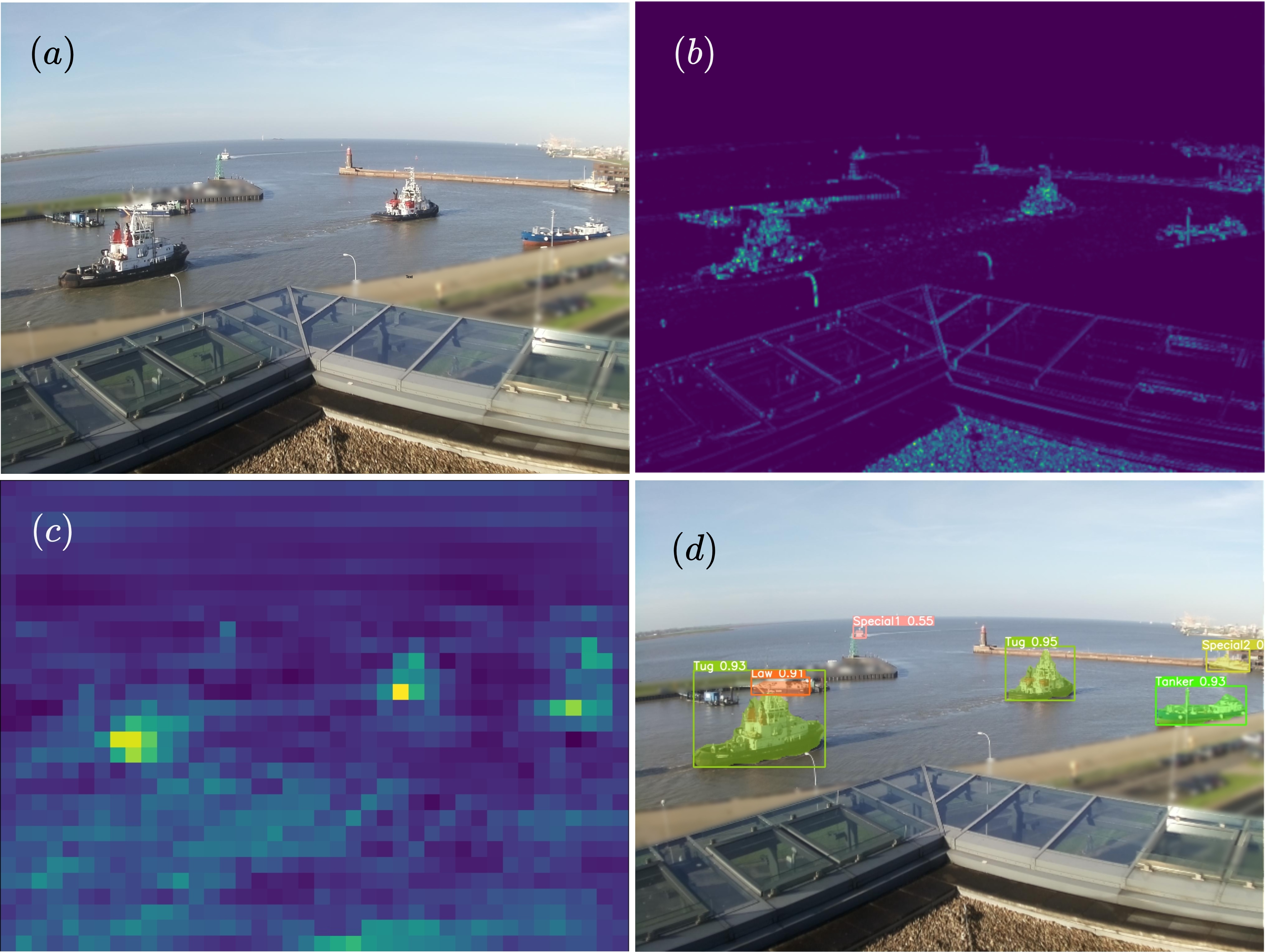}
\caption[Instance segmentation process of ScatYOLOv8+CBAM on ShipSG.]{Instance segmentation process of ScatYOLOv8+CBAM on ShipSG. (a) ShipSG input sample image. (b) Output of the ScatBlock (here, for visualization, the mean of output channels without $S_0$). (c) Output of CBAM (module number 21 in Fig. \ref{fig:scatyolov8+cbam}). (d) ShipSG image with segmented and classified ships using the proposed architecture ScatYOLOv8n+CBAM. Reprinted from~\ref{paper:5}. \textcopyright 2023 IEEE.}
\label{fig:scat_in_action}
\end{figure}

In the example of ship segmentation inference on ShipSG, as illustrated in Fig.~\ref{fig:scat_in_action}, it is observable that the inclusion of the ScatBlock enhances the ships within the image. This enhancement significantly improves the clarity and visibility of the ship edges, leading to a more defined and distinct outline of the ships present. Regarding the \gls{cbam} output, it is noted that the implemented attention mechanisms (both channel and spatial) effectively complement the scattering transform. The attention maps enhance the location of ships within the image while minimizing background influence. 

\begin{table}[h]
\centering
\caption[Comparison of state-of-the-art segmentation performances on ShipSG with YOLOv8n and ScatYOLOv8n+CBAM.]{Comparison of state-of-the-art segmentation performances on ShipSG with YOLOv8n and ScatYOLOv8n+CBAM. Adapted from~\ref{paper:5}. \textcopyright 2023 IEEE.}
\label{tab:scat_sota}
\begin{tabular}{lcc}
\hline
\textbf{Segmentation model} & \textbf{mAP (\%)}\\ \hline
Mask R-CNN&73.3\\
DetectoRS&74.7\\
YOLACT$_{700}$ & 58.20 \\
Centermask-Lite$_{V39}$ & 64.40 \\\hline
YOLOv8n  & 70.15\\
\textbf{ScatYOLOv8n + CBAM}& \textbf{75.46}\\ \hline
\end{tabular}
\end{table}

In \ref{paper:5}, I focused on the lightest version of YOLOv8 for the implementation ScatYOLOv8+CBAM, version $n$, due to its potential for real-time operation. 
As seen in Table~\ref{tab:scat_sota}, the baseline YOLOv8n shows improvement compared to the \gls{map} of previous real-time approaches for ship segmentation on ShipSG studied in Chapter~\ref{chap:ship_rec}, YOLACT and Centermask-Lite. Yet, it does not show advantage against more robust methods like Mask R-CNN and DetectoRS.
ScatYOLOv8n+CBAM achieves 5.31\% improvement with respect to standard YOLOv8n, 11.06\% improvement compared to Centermask-Lite and 0.76\% with respect to the most robust of the previous study, DetectoRS. 
While the \gls{map} increase is modest when compared to DetectoRS, DetectoRS provided an inference time of 151 ms using a high-end \gls{gpu}. 
As discussed in Chapter~\ref{chap:ship_rec}, the deployment of instance segmentation on \gls{gpu}-powered embedded systems was not reported for the methods presented in the initial study due to the incompatibility between deep learning and the \gls{arm} architectures of \gls{gpu}-powered embedded systems.
The advantage of YOLOv8-like architectures, such as ScatYOLOv8+CBAM, is the deployability on embedded systems. In \ref{paper:5}, the NVIDIA Jetson AGX Xavier was selected as the target embedded system for deployment. 
The ScatYOLOv8+CBAM model was deployed using native Pytorch weights.
This allow us to measure inference times of the new architecture on the system, which will show a great advantage against the previously studied instance segmentation methods.

\begin{table}[htbp]
\centering
\caption[Ablation study of YOLOv8 segmentation models and ScatYOLOv8+CBAM additions.]{Ablation study of YOLOv8 segmentation models and ScatYOLOv8+CBAM additions after training on ShipSG and inference times on the NVIDIA Jetson AGX Xavier. Reprinted from~\ref{paper:5}. \textcopyright 2023 IEEE.}
\label{tab:scat_times}
\begin{tabular}{lllll}
\hline
\multicolumn{1}{l|}{\textbf{Segmentation model}} & \multicolumn{2}{c|}{\textbf{\gls{map} (\%)}} & \multicolumn{2}{c}{\textbf{Inference (ms)}} \\ \hline
\multicolumn{1}{l|}{YOLOv8n} & 70.35 & \multicolumn{1}{l|}{-} & 28.7 & - \\
\multicolumn{1}{l|}{YOLOv8s} & 71.99 & \multicolumn{1}{l|}{($\uparrow$1.64)} & 32.2 & ($\uparrow$3.5) \\
\multicolumn{1}{l|}{YOLOv8m} & 74.84 & \multicolumn{1}{l|}{($\uparrow$4.49)} & 72.4 & ($\uparrow$43.7) \\
\multicolumn{1}{l|}{YOLOv8l} & 75.89 & \multicolumn{1}{c|}{($\uparrow$5.54)} & 127.1 & ($\uparrow$98.4) \\
\multicolumn{1}{l|}{YOLOv8x} & 76.45 & \multicolumn{1}{r|}{($\uparrow$6.10)} & 196.6 & ($\uparrow$167.9) \\ \hline
\multicolumn{1}{l|}{ScatYOLOv8n} & 74.42 & \multicolumn{1}{r|}{($\uparrow$4.07)} & 58.2 & ($\uparrow$29.5) \\
\multicolumn{1}{l|}{YOLOv8n + CBAM} & 70.75 & \multicolumn{1}{r|}{($\uparrow$0.40)} & 29.9 & ($\uparrow$1.2) \\
\multicolumn{1}{l|}{\textbf{ScatYOLOv8n + CBAM}} & \textbf{75.46} & \multicolumn{1}{r|}{\textbf{($\uparrow$5.11)}} & \textbf{59.3} & \textbf{($\uparrow$30.6)} \\ \hline
\end{tabular}
\end{table}

In assessing the inference times on the embedded system for the proposed enhancements within ScatYOLOv8+CBAM, specifically the ScatBlock and \gls{cbam}, their individual impacts were also examined in the ablation study presented in Table~\ref{tab:scat_times}. The first part of the table outlines the performance metrics of each YOLOv8 model variant. The second part provides the individual and combined contributions of the enhancements introduced in this work. The increments are noted in comparison to the baseline performance of the YOLOv8n model.
It can be seen in Table~\ref{tab:scat_times} that the addition of \gls{cbam} produces an increased \gls{map} at a very minimal computational cost.
The proposed architecture ScatYOLOv8+CBAM, in the lightest version $n$, provides a \gls{map} comparable to the deeper and heavier YOLOv8l (75.46\% vs 75.89\%). 
However, the proposed model demonstrates a substantially faster inference speed (59.3 ms versus 127.1 ms) on the NVIDIA Jetson AGX Xavier. This marks a significant improvement over the preliminary findings discussed in Chapter~\ref{chap:ship_rec}, Section~\ref{sec:std_ship_seg}.

It has been shown that ScatYOLOv8+CBAM enables the efficient handling of images in maritime environments deployed on embedded systems, facilitating faster and more accurate real-time ship segmentation by leveraging the capabilities of \gls{cnn}s with the added robustness provided by the scattering transform and attention mechanisms. 
This indicates its potential viability to enhance real-world maritime situational awareness applications.

It is important to note that one of the goals of this thesis is the improvement of maritime situational awareness leveraging ship segmentation for accurate georeferencing.
Therefore, to further validate the ScatYOLOv8+CBAM architecture, the output masks of this architecture are evaluated for ship georeferencing Chapter~\ref{chap:ship_geo}, Section~\ref{sec:quant_ship_geo}, as presented in~\ref{paper:5}. This evaluation shows, with higher segmentation \gls{map}, consistent results for georeferencing compared to the georeferencing evaluation of the standard segmentation methods studied in Chapter~\ref{chap:ship_rec}, Section~\ref{sec:std_ship_seg}.

\section[Optimized ScatYOLOv8+CBAM]{Optimized ScatYOLOv8+CBAM~\ref{paper:6}}
\label{sec:opt_scatyolo}

As it was motivated in Chapter~\ref{chap:intro}, to enhance maritime situational awareness, optimizing for real-time processing capabilities is key. Therefore, ship recognition should operate with the highest possible accuracy and the shortest inference times on embedded systems. 
This section studies optimizations to ScatYOLOv8+CBAM for a more time-efficient ship segmentation, by circumventing redundancies in the original ScatBlock.

As shown in Section~\ref{sec:scatblock}, the 2D scattering transform typically downsamples the input image by a factor of $2^J$ to reduce computational complexity across scales. 
For the ScatBlock, which uses only first-order coefficients, this translates to an output resolution that is half the input resolution. To address this, the ScatBlock upsamples the input image to $(2 \times H) \times (2 \times W)$ (see fig.~\ref{fig:scatblock}). This ensures that the output dimensions match the size of the input image, which is crucial for the compatibility with subsequent YOLOv8 backbone blocks.
However, the sequential upsampling and downsampling in image resolution increases computational burden. 
Furthermore, while the deployment on the NVIDIA Jetson AGX Xavier was documented in~\ref{paper:5} (Section~\ref{sec:scatyolo}) using Pytorch weights, the potential for model optimization with TensorRT to achieve more efficient real-time inference was not investigated.
The contributions of~\ref{paper:6} address these areas of improvement, by optimizing the custom architecture ScatYOLOv8+CBAM and performing a comprehensive evaluation for ship segmentation with all model sizes, focusing on real-time processing on the embedded system. 

The main optimization focuses on the ScatBlock. As explained in Section~\ref{sec:scatblock}, the ScatBlock was implemented using the open-source Python module $pytorch\_wavelets$ \cite{cotter_2020} to achieve the 2D scattering transform. 
In the optimization, the initial step involves removing the downsampling associated with the scattering transform. This refinement is achieved by bypassing, within the $pytorch\_wavelets$ package, the division of the image into distinct frequencies and considering all frequency components simultaneously.
Following this, the downsampling step of the scattering transform is omitted. The absence of quadrant division means that downsampling would inappropriately cause a mismatch in resolution by presupposing a quadrant-based reduction. As a result, the enhanced ScatBlock does not need the upsample to retain the original resolution in its output feature map. This ensures the preservation of vital image details crucial for accurate segmentation, while simultaneously boosting inference speed.

\begin{figure}[H]
\centering
\includegraphics[width=11.5 cm]{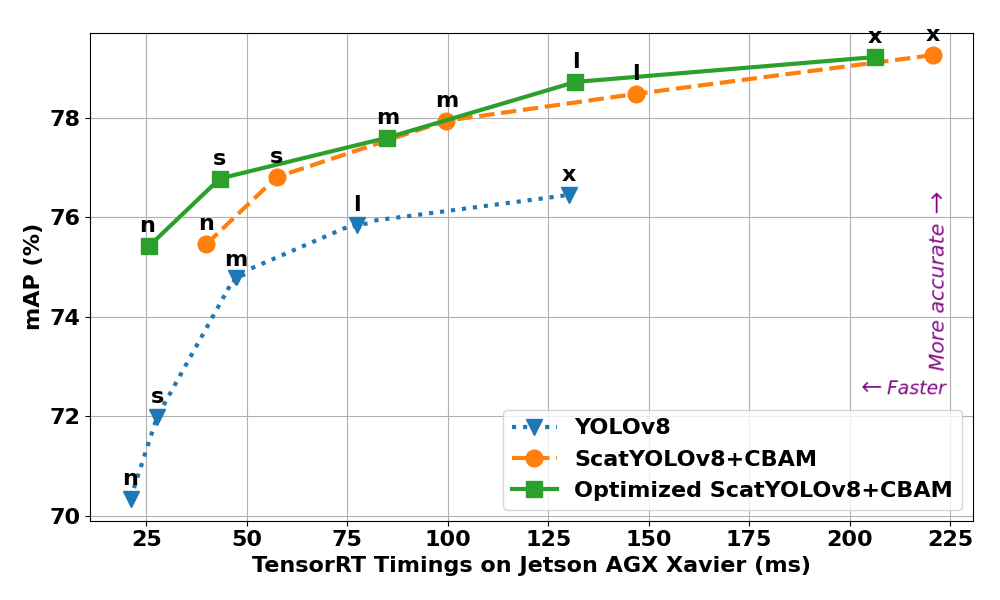}
\caption[mAP vs TensorRT inference times on NVIDIA Jetson AGX Xavier, showing the improved performance speed.]{\gls{map} vs TensorRT inference times on NVIDIA Jetson AGX Xavier, showing the improved performance speed of the optimized ScatYOLOv8+CBAM relative to the predecessor. Reprinted from~\ref{paper:6}. \textcopyright 2024 IEEE.}
\label{fig:scat_comparison}
\end{figure}

The optimization of the ScatYOLOv8+CBAM architecture introduced~\ref{paper:5} was presented in~\ref{paper:6}, which removes the upsample from the Scatblock (see upsample in Fig.~\ref{fig:scatblock}) and the downsampling operation eliminated from the $pytorch\_wavelets$ package~\cite{cotter_2020}.
To examine the optimization, the original ScatYOLOv8+CBAM serves as a benchmark for comparison, as well as standard YOLOv8 (see Fig.~\ref{fig:scat_comparison}).
The newly optimized ScatYOLOv8+CBAM was trained across all model sizes ($n$, $s$, $m$, $l$, $x$), employing the original parameters specified in~\ref{paper:5}. This includes using an input size of $640\times640$ pixels, initializing weights randomly, and setting the training period to 300 epochs. To measure inference times, the duration from when an image is inputted to when predictions are obtained is recorded, now using TensorRT-exported weights on the Jetson AGX Xavier. The approach in \ref{paper:6} of using TensorRT contrasts with the prior utilization of Pytorch weights in~\ref{paper:5} (Section~\ref{sec:scatyolo}). 

A comparison of mAP for ship segmentation and TensorRT inference times on the Jetson AGX Xavier can be seen in Figure~\ref{fig:scat_comparison}.
The comparison covers the optimized ScatYOLOv8+CBAM, its predecessor, and the standard YOLOv8.
The optimized ScatYOLOv8+CBAM $n$ model provides comparable precision compared with the earlier version and the larger $l$ model of YOLOv8, with \gls{map}s of 75.39\%, 75.46\%, and 75.89\%, respectively.
Yet, the optimized model is much faster, with an inference time of 25.3 ms versus 39.9 ms and 77.5 ms for the other models.
This shows that the $n$ model of the optimized ScatYOLOv8+CBAM is 36.5\% faster than its predecessor with a minor drop in mAP compared to the original ScatBlock-equipped architecture (0.06\%).
Additionally, the $s$ and $m$ models exceed in mAP over the largest YOLOv8$x$ but offering quicker inference.

The optimized ScatYOLOv8+CBAM speeds up inference against its predecessor, making it more suitable for real-time use.
Although the $l$ and $x$ sizes of the optimized model lead in \gls{map}, their slower inference speeds limit their real-time deployment.

In summary, the optimized ScatYOLOv8+CBAM has achieved faster inference speeds on the NVIDIA Jetson AGX Xavier with TensorRT, outpacing the previous version. Moreover, it surpasses standard YOLOv8 models in \gls{map}.
However, while shallower models demonstrate significant improvements in accuracy and speed compared to state-of-the-art ship segmentation on ShipSG, deeper and larger models exhibit a noticeable slowdown. This slowdown suggests an increase in computational complexity within the \gls{cnn} when it processes scattering coefficients in larger models, opening room for further improvements that will be discussed in Section~\ref{sec:adv_discussion}. 

\section[Enhanced Small Ship Segmentation Using Higher Resolution Images]{Enhanced Small Ship Segmentation Using Higher Resolution Images~\ref{paper:6}}
\label{sec:small_ship_rec}

In Chapter~\ref{chap:ship_rec}, we observed a \gls{map} decrease in segmenting small and distant ships, especially for the initially considered as real-time methods (see Table~\ref{tab:std_result}). 
As motivated in Chapter~\ref{chap:intro}, the recognition of small and distant ships in maritime footage holds significant implications for navigation, safety, and security. 
Typically, object detectors and instance segmentation methods reduce image size for quicker inference, losing critical details in the image.
On the other hand, high-resolution deep learning approaches strain memory and computational resources, particularly on embedded systems with limited capacity. 
Innovating segmentation architectures for such systems is crucial to overcome these challenges.
In~\ref{paper:6}, the proposed batch-processed \gls{sahi}, combined with the optimized version of ScatYOLOv8+CBAM, advance the state-of-the-art in small ship segmentation using embedded platforms.
 
Existing deep-learning approaches use image super-resolution or incorporate additional network blocks~\cite{rekavandi2022guide, wang2023uav}, which is not ideal for embedded systems due to memory constrains.
The \gls{sahi} framework~\cite{sahi23akyon} improves the recognition of small objects in high-resolution images by dividing images into overlapping patches that retain their original size. Then, object detection or segmentation is performed in sequentially on the patches using a compatible method, such as YOLOv8. The resulting detections are then merged with \gls{nms}. In the case of the segmented masks, they are merged using \gls{nms} and then combined appropriately with a logical OR operator. 

The sequential inference of \gls{sahi} limits speed, suggesting batch inference integration could boost efficiency on the embedded system.
The work in~\ref{paper:6} modifies the \gls{sahi} framework\footnote{\url{https://github.com/obss/sahi/}} by adding batch processing for the inference stage, addressing the original sequential slice processing constraint as documented in~\cite{sahi23akyon}. By enabling ScatYOLOv8+CBAM (and YOLOv8) to infer masks on multiple slices simultaneously ($batch = N_{slices}$), the inference phase is made more resource-efficient. Preprocessing (slicing) and postprocessing (merging) within \gls{sahi} remain unchanged. 

Splitting an image into slices, to form a batch, optimizes both memory usage and computational efficiency, especially on embedded systems with limited \gls{gpu} memory. Performing object recognition on the high-resolution images may exceed \gls{gpu} memory capacity if processed as a whole, leading to out-of-memory errors. By splitting the image, each slice can be processed independently within memory constraints. Batch processing these slices allows the \gls{gpu} to handle multiple slices concurrently, leveraging its parallel processing capabilities. This approach balances the need to manage memory effectively while maximizing computational throughput, making it ideal for resource-constrained environments like the NVIDIA Jetson AGX Xavier.

Additionally, the slicing mechanism was used for model fine-tuning, with slices of the full-resolution ShipSG images as new training set. This allows the model to focus on improving the ability to segment small ships from full-resolution image slices of the dataset. By incorporating batch inference for inference and targeted fine-tuning during training, the optimized ScatYOLOv8+CBAM with \gls{sahi} not only maintains quality for small ship segmentation but also boosts real-time performance on embedded systems.

For the fine-tuning process, a new training dataset comprising 33648 slices from ShipSG images was created, each of $640\times640$ pixels and incorporating a $20\%$ overlap between slices. This overlap guarantees that sufficient contextual information is retained for objects at the edges, yielding the most favorable experimental outcomes.
Per image of full-resolution in ShipSG ($2028\times1520$ pixels), the total number of slices of $640\times640$ pixels is $12$. 
This augmented dataset facilitated the training of the optimized ScatYOLOv8+CBAM model, initiating as pre-trained weights the models of Section~\ref{sec:opt_scatyolo}, and extending with an extra 30 epochs of training. Analogously, this fine-tuning approach was applied as well to standard YOLOv8 models for comparative analysis.

At the inference stage, the \gls{sahi} preprocessing includes the real-time slicing of the image from which ships are being segmented.
This means that inference times when using \gls{sahi} encompass the total time from inputting an image to obtaining predictions, with all processing steps, using TensorRT-exported weights on the Jetson AGX Xavier.
For inference, the framework uses the same slicing parameters as during the fine-tune training stage, that is, $12$ slices of $640\times640$ pixels per full-resolution image, with $20\%$ overlap. 

\begin{figure}[h]
\centering
\includegraphics[width=7.5cm]{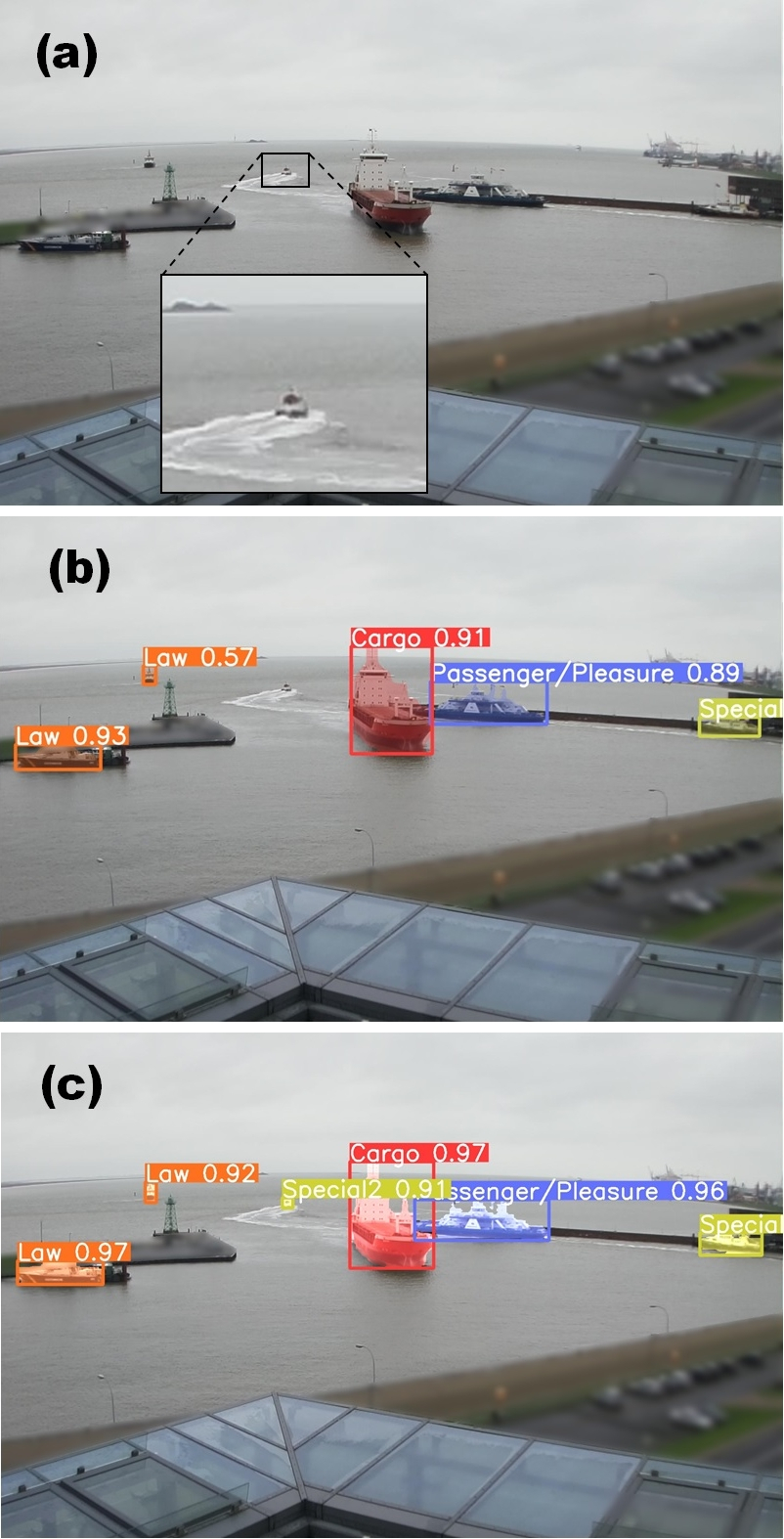}
\caption[Small ship segmentation on ShipSG.]{Small ship segmentation on ShipSG. (a) Original ShipSG image, with a small ship zoomed in for visualization. (b) Inference using the optimized ScatYOLOv8$x$+CBAM, where the small ship has been undetected. (c) Inference using the proposed optimized ScatYOLOv8$x$+CBAM with \gls{sahi}, where the small ship appears segmented, at a high confidence score. Reprinted from~\ref{paper:6}. \textcopyright 2024 IEEE.}
\label{fig:small_example}
\end{figure}

We can see in Figure \ref{fig:small_example} the effectiveness of the \gls{sahi} approach, illustrating how a small ship, undetected without \gls{sahi}, is accurately segmented, highlighting its significance in boosting maritime situational awareness. Furthermore, the enhanced confidence scores for larger ships in the same figure underscore the robustness of \gls{sahi}.

\begin{table}[h]
\caption[Comparison of \gls{map} scores for small objects with all model sizes using standard YOLOv8, our proposed optimized ScatYOLOv8+CBAM, and the addition of \gls{sahi}]{Comparison of \gls{map} scores for small objects with all model sizes using standard YOLOv8, our proposed optimized ScatYOLOv8+CBAM, and the addition of \gls{sahi}. Reprinted from~\ref{paper:6}. \textcopyright 2024 IEEE.}
\centering
\begin{tabular}{l|ccccc}
\multirow{2}{*}{\textbf{Model}} & \multicolumn{5}{c}{\textbf{mAP small objects (\%)}} \\ \cline{2-6}
 & \textbf{n} & \textbf{s} & \textbf{m} & \textbf{l} & \textbf{x} \\ \hline
\rule{0pt}{2.5ex}YOLOv8 & 39.9 & 40.8 & 41.8 & 42.9 & 43.4 \\
Opt. ScatYOLOv8+CBAM & 45.8 & 47.1 & 47.2 & 47.9 & 48.0 \\ \hline
\rule{0pt}{2.5ex}YOLOv8 \& \gls{sahi} & 53.1 & 54.2 & 55.0 & 55.4 & 55.7 \\
Opt. ScatYOLOv8+CBAM \& \gls{sahi} & \textbf{54.7} & \textbf{55.6} & \textbf{56.7} & \textbf{57.9} & \textbf{58.9}
\end{tabular}
\label{tab:small_comparison}
\end{table}

\begin{figure}[h]
\centering
\includegraphics[width=11.5cm]{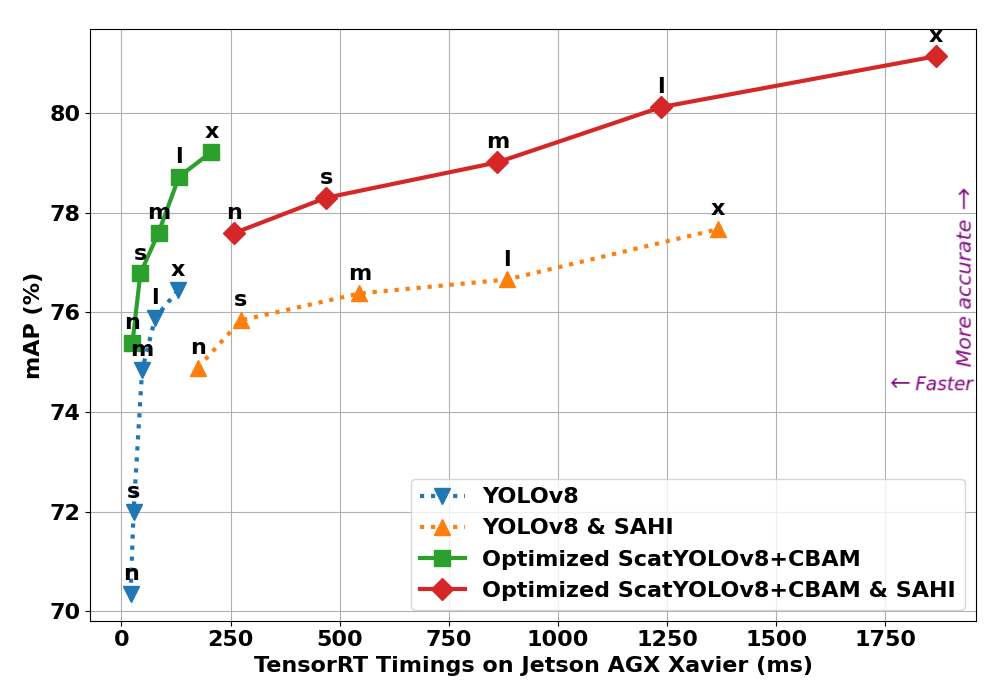}
\caption[mAP vs TensorRT inference times on NVIDIA Jetson AGX Xavier of ScatYOLOv8+CBAM with SAHI.]{mAP vs TensorRT inference times on NVIDIA Jetson AGX Xavier. 
ScatYOLOv8+CBAM with \gls{sahi} provides the best accuracy for ship segmentation. Reprinted from~\ref{paper:6}. \textcopyright 2024 IEEE.}
\label{fig:sahi_comparison}
\end{figure}

Table~\ref{tab:small_comparison} presents the results of the analysis on small ship segmentation across different model sizes. 
The optimized ScatYOLOv8+CBAM outperforms the standard YOLOv8 for small objects across all sizes. 
\gls{sahi} significantly boosts performance for both standard YOLOv8 and the optimized ScatYOLOv8+CBAM, showing gains between $8\%$ and $11\%$ over configurations without \gls{sahi}. Specifically, the optimized ScatYOLOv8+CBAM with \gls{sahi} achieves the highest mAP improvements for small objects, ranging from $1.4\%$ to $3.2\%$ over standard YOLOv8 equipped with \gls{sahi}. Moreover, the advantage of integrating our optimized model with \gls{sahi} becomes more pronounced with model depth, highlighting a scalable improvement in small ship segmentation with increased network complexity.
When comparing to the small ship segmentation performance provided in Chapter~\ref{chap:ship_rec}, Table~\ref{tab:std_result}, the optimized ScatYOLOv8+CBAM with \gls{sahi} in model size $s$ onwards performs as good or better than the best one from the initial study, DetectoRS. It is important to recall that DetectoRS was not compatible with embedded system deployment, which underlines another superiority of the customized architecture.

Figure~\ref{fig:sahi_comparison} compares the overall mAP (all mask sizes) of models using batch-processed \gls{sahi} against those without it on the Jetson AGX Xavier.
The ScatYOLOv8+CBAM model with \gls{sahi} outperforms all standard YOLOv8 models in ship segmentation accuracy.
The ScatYOLOv8+CBAM $n$ with \gls{sahi} not only achieves higher mAP than the largest YOLOv8$x$ with \gls{sahi} but also performs substantially faster, sparing resources of the embedded system even on full-resolution images.

Compared to its own non-\gls{sahi} version, the optimized ScatYOLOv8+CBAM with \gls{sahi} in $n$ size shows equivalent mAP to a the $m$ model without \gls{sahi}, albeit with increased computation time of $\sim$170 ms, significantly enhancing small ship segmentation by 7.5\%, as shown in Table~\ref{tab:small_comparison}.
This shows that the benefits of \gls{sahi} in densely populated maritime areas where monitoring small, distant ships in real-time is crucial for safety and security, carry considerable associated computational costs. 

During the tests, it was measured that 25\% of the processing time during \gls{sahi} batch-inference goes to slicing and merging the masks form the slices, suggesting room for optimization.

\section{Summary and Discussion}
\label{sec:adv_discussion}

This chapter presented an advancement in the field of real-time ship segmentation with the design of ScatYOLOv8+CBAM. In lightest version $n$, the architecture demonstrated its capability on ShipSG with a \gls{map} of 75.46\%, 5\% higher than baseline, comparable to the performance of larger baseline models but at half inference speed per frame (59.3 ms).
We analyzed an optimization of the architecture, that removes the upsampling and downsampling from the ScatBlock to save computing time, and deployed it with TensoRT on the Jetson AGX Xavier to measure inference times for real-time applicability. The optimized ScatYOLOv8+CBAM in model size $n$ performs 36.5\% faster than its predecessor, achieving 25.3 ms per frame.
Finally, I proposed a batch-processed \gls{sahi} to increase the segmentation of small and distant ships that is able to run within embedded system resources. Though bringing along increased computation, the \gls{map} for small ships increased in ranges from 8\% to 11\% in comparison with the baseline without \gls{sahi}.
The work presented bridges the transition from standard methods to real-time instance segmentation on embedded systems, and addresses the ability to accurately identify all ships, independent from their size, and within the vicinity of the port area. 

Additionally, the chapter has uncovered various areas for enhancement. Further research could focus on improving ScatYOLOv8+CBAM and advancing the performance of \gls{sahi}.
As seen in Section~\ref{sec:opt_scatyolo}, shallower models show notable improvements in both accuracy and speed over the standard ship segmentation models explored with ShipSG. 
However, larger models become slower. This indicates that larger models face increased computational demands when handling scattering coefficients, hinting at potential strategies for refinement to be explored.

One possible strategy is making the scattering transform learnable. This would involve introducing adjustable parameters within the wavelet filters (parametrization) by merging deep learning adaptability with theoretical wavelet analysis. However, this is a challenging task due to the complicated balance between maintaining the translation and deformation invariant properties of the transform while allowing for sufficient flexibility to learn from data.

The second possible strategy is enhancing the transform with learnable downsampling and attention mechanisms for a more practical approach for real-time use. For example, learnable downsampling, such as strided convolutions with batch normalization, can compress scattering coefficients while learning to preserve key information.
On the other hand, attention mechanisms focus computational resources on significant features, improving processing efficiency. An example is the integration of transformers, that are based on attention and could enable models to combine the invariant feature extraction of the scattering transform with the transformer contextual learning, offering a focus on the most relevant features for the segmentation.

Moreover, it was measured during the experiments that 25\% of the processing time during \gls{sahi} batch-inference is dedicated to the slicing (preprocessing) and merging of the masks after segmentation (postprocessing). In the pursuit of enhancing \gls{sahi}, by using parallel processing, the slicing and merging tasks could be improved. An example is the division of the workload with multi-threading or a pool of processes to enable concurrent execution of slicing and merging, optimizing computational resources and improving efficiency.

Given the slowdown in larger models, is pertinent to address how potential practitioners and users of ScatYOLOv8+CBAM should select the size of the architecture (from $n$ to $x$), and the use of \gls{sahi}, in their specific application.
The decision should consider the critical balance between real-time processing demands and computational limitations. The optimized ScatYOLOv8+CBAM, with its \gls{sahi} adaptation for small and distant ships, serves as a guide for such selections. Users should weigh the operational complexity, the necessity for rapid data processing, and accuracy requirements against the computational resources available. By aligning architecture choices with these considerations, practitioners and users can ensure the an effective deployment tailored to their unique application needs.
For instance, the optimized ScatYOLOv8+CBAM in smaller configurations ($n$ and $s$) is ideal for scenarios demanding rapid response with considerable accuracy, such as real-time port surveillance or navigation aid systems.
Additionally, though optional, the use of \gls{sahi}, could enhance access to small and distant ships approach the maritime infrastructure, when the application requires higher levels of responsiveness.
Conversely, in scenarios where computational resources are less constrained, larger configurations ($m$, $l$ and $x$) could be leveraged for enhanced precision. This choice underscores the necessity for a strategic approach in deploying these technologies, where understanding the specific maritime application context and corresponding computational trade-offs guides the optimal use of the architecture.

Integrating the methodologies presented in this chapter with other processing chains and sensors can also further enhance maritime situational awareness. 
This could include displaying real-time georeferenced ships on maps through web services, merging various data sources such as ship tracking, 3D reconstruction and anomaly detection or information from other sensors as thermal imaging or radar.

In conclusion, the results presented in this Chapter establish a new standard for maritime monitoring on embedded systems and create a foundation for future work aimed at enhancing real-time, high-resolution processing within the limits of resource-restricted settings.
\chapter{Ship Georeferencing for Maritime Situational Awareness}
\label{chap:ship_geo}

In Chapters~\ref{chap:shipsg}, \ref{chap:ship_rec} and~\ref{chap:adv_ship_rec}, we explored the creation of a custom maritime dataset, the impact of ship recognition on maritime applications, and the advancements and optimizations for real-time processing on embedded systems with a customized architecture, even with higher resolutions to enhance detail in small ship recognition. 
Building on these foundations, this chapter studies the georeferencing of recognized ships using monocular images. 
The aim is the implementation of a method that provides meaningful information from the recognized ships to the situational awareness picture for a better semantic understanding of the maritime situation. This process involves the development of methods for the representation of the recognized ship on a global scale using single images and without prior knowledge of the camera.
We call this ship georeferencing.

First, we delve into understanding the concept of homographies for georeferencing, which serves as fundamental for the ship georeferencing techniques proposed.

Then, this chapter presents my proposed ship bounding box georeferencing method and calculation of ship heading direction from optical flow, which form a significant part of my contributions to~\ref{paper:1}, in addition to what has been presented in Sec.~\ref{sec:ship_det_abnormal}.  

Moreover, I expand upon the georeferencing methodology and show an in depth quantitative studies of the use of homographies for ship georeferencing using ShipSG and the results from Sec.~\ref{sec:std_ship_seg} and Sec.~\ref{sec:scatyolo}, based on~\ref{paper:2} and~\ref{paper:5}, respectively.

\section[Homographies for Image Georeferencing]{Homographies for Image Georeferencing~\ref{paper:2}}
\label{sec:homo}
\begin{figure}[h]
\includegraphics[width=1\columnwidth]{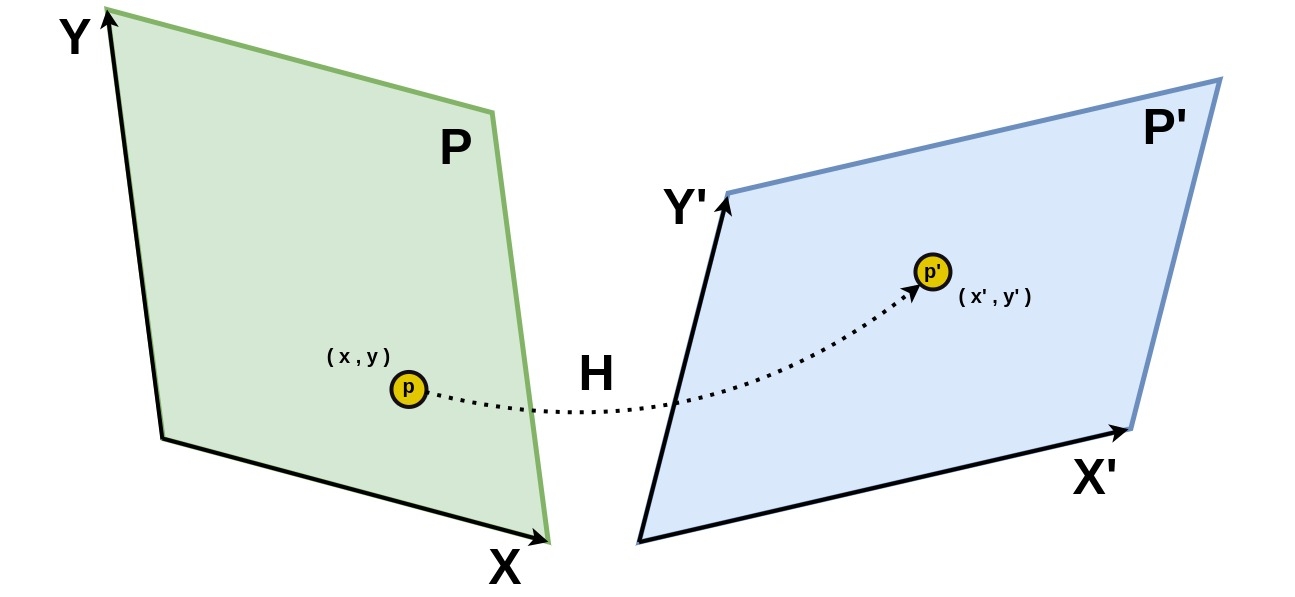}
\centering
\caption[Homography between two planes.]{Homography between two planes. Plane $P$ and $P'$ represent two different surfaces or image planes. The point $p$ on $P$ and $p'$ on $P'$ illustrate a pair of points that are related by a homography ($\gls{h}$), showcasing how a point in one plane can be mapped to another plane through a projective transformation (dashed arrow).}\label{fig:homography}
\end{figure}

Homography~\cite{hartley2003multiple} is a mathematical concept widely used in the fields of computer vision and image processing to describe the transformation of points between two planes (see Fig.~\ref{fig:homography}). 
This transformation, encapsulated by the homography matrix, allows for the conversion of coordinates between these two planes, highlighting the essence of homography in linking different spatial perspectives.

Essentially, a homography is represented by a $3\times3$ matrix: 

\begin{equation}
H = \begin{bmatrix}
h_{11} & h_{12} & h_{13} \\
h_{21} & h_{22} & h_{23} \\
h_{31} & h_{32} & 1 \\
\end{bmatrix}
\end{equation}

When the \gls{h} matrix is multiplied by a point in homogeneous coordinates, it maps it from one plane to another. 
It offers a powerful tool for projective transformations, allowing for the mapping of the geometric correspondence between points $p$ and $p'$ in different planes, such that

\begin{equation}
\label{eq:p=hp}
    p' = H \cdot p
\end{equation}

where $p = \begin{bmatrix} x \\ y \\ 1 \end{bmatrix}$ and $p' = \begin{bmatrix} x' \\ y' \\ 1 \end{bmatrix}$, as seen in Fig.~\ref{fig:homography}.

The creation of a homography matrix requires the identification of correspondences between points in the two planes, typically two images.
Once a sufficient number of point pairs (usually at least four non-collinear points) is established, the homography matrix can be solved using linear algebra techniques. 
The process involves setting up a system of equations based on the point correspondences and solving for the eight unknowns of the homography matrix (the ninth element is conventionally set to 1 for normalization). 

\begin{equation}
\label{eq:h_solve}
\begin{bmatrix}
    x_1 & y_1 & 1 & 0 & 0 & 0 & -x'_1x_1 & -x'_1y_1 \\
    0 & 0 & 0 & x_1 & y_1 & 1 & -y'_1x_1 & -y'_1y_1 \\
    x_2 & y_2 & 1 & 0 & 0 & 0 & -x'_2x_2 & -x'_2y_2 \\
    0 & 0 & 0 & x_2 & y_2 & 1 & -y'_2x_2 & -y'_2y_2 \\
    x_3 & y_3 & 1 & 0 & 0 & 0 & -x'_3x_3 & -x'_3y_3 \\
    0 & 0 & 0 & x_3 & y_3 & 1 & -y'_3x_3 & -y'_3y_3 \\
    x_4 & y_4 & 1 & 0 & 0 & 0 & -x'_4x_4 & -x'_4y_4 \\
    0 & 0 & 0 & x_4 & y_4 & 1 & -y'_4x_4 & -y'_4y_4 \\
    \vdots & \vdots & \vdots & \vdots & \vdots & \vdots & \vdots & \vdots \\
    x_n & y_n & 1 & 0 & 0 & 0 & -x'_nx_n & -x'_ny_n \\
    0 & 0 & 0 & x_n & y_n & 1 & -y'_nx_n & -y'_ny_n \\
\end{bmatrix}
\begin{bmatrix}
    h_{11} \\
    h_{12} \\
    h_{13} \\
    h_{21} \\
    h_{22} \\
    h_{23} \\
    h_{31} \\
    h_{32} \\
\end{bmatrix}
=
\begin{bmatrix}
    x'_1 \\
    y'_1 \\
    x'_2 \\
    y'_2 \\
    x'_3 \\
    y'_3 \\
    x'_4 \\
    y'_4 \\
    \vdots \\
    x'_n \\
    y'_n \\
\end{bmatrix}
\end{equation}

To solve the homography, these equations are stacked for all correspondences to form a system $Ah = b$.
Typically, \gls{dlt}, \gls{ransac}, or \gls{ls} are employed to find the optimal homography matrix that minimizes the re-projection error between the observed and predicted points.
\gls{ls} provided the most accurate results for this application and so was exclusively considered for the remainder of this work.
The LS solution to $Ah = b$ is given by minimizing $\|Ah - b\|^2$, which leads to the solution:
\begin{equation}
h = (A^TA)^{-1}A^Tb
\end{equation} 
where $h$ is the vector containing the eight unknown elements of $\gls{h}$.
Finally, $h$ is reshaped back into the $3 \times 3$ homography matrix $\gls{h}$.

Once the homography matrix $\gls{h}$ is created, it can be used as shown in Eq.~\ref{eq:p=hp}. 
In computer vision, homography has been extensively utilized across a range of applications. 
These include image stitching in panoramic photography, feature matching, camera calibration, 3D reconstruction, augmented reality for overlaying virtual objects onto real-world scenes, motion estimation in video sequences, and perspective correction for rectifying images of planar surfaces.
In the field of \gls{gis}, it has been used to map ground surfaces in surveillance video frames to topographies that are small enough (few hundred meters range) to be approximated as a plane~\cite{xie2017integration,shao2020accurate}.
However they do not provide quantitative assessments of their accuracy.

The ship georeferencing methodology proposed in this thesis also treats the Earth's surface observed by the camera as a planar area for simplification. For example, in the case of ShipSG, given the 23 meter altitude of the cameras and considering the georeferencing range of up to 1200 meters on the Weser river, the curvature of the Earth introduces a minimal height difference. This difference can be approximated by $h \approx \frac{d^2}{2r}$, where $d$ is the horizontal distance between two points on the Earth's surface, and $r$ is the radius of the Earth. This formula derives from the geometric properties of a circle, where the Earth's curvature over a small distance can be represented as the segment height ($h$) of a circular segment with radius $r$. For the observation range of 1200 meters, this formula gives the height difference due to curvature to be approximately 0.113 meters. This value is significantly small, especially when compared to the elevation of the camera and and length of the ships. Consequently, we simplify the water surface visible from the cameras as the tangent plane to the Earth's curvature.

By establishing correspondences between known geographic points on the water surface (e.g., buoys, landmarks~\ref{paper:1}, or \gls{ais} signals from ships~\ref{paper:2}) and their representations in the image frame, a homography matrix can be computed. 
This matrix then allows for any point captured on the water surface to be mapped to geographic coordinates. 
This method not only facilitates the accurate mapping of static water surfaces but also paves the way for dynamic georeferencing applications, such as real-time ship recognition. 
By leveraging the homography matrix, ships on the water can be precisely georeferenced, offering valuable insights for maritime situational awareness and bridging the digital and physical worlds.

The static view of a monitoring camera allows the water surface captured in the image, with pixel coordinates, to be transformed to the water surface with geographic latitude and longitude coordinates.
Homography offers therefore a potent solution that enables ship georeferencing without the need for detailed camera calibration (i.e., intrinsic or extrinsic parameters).
The work in~\ref{paper:1} explores the homography-based method, and an experimental analysis of the method is presented in~\ref{paper:2} and~\ref{paper:5}. 
A significant advantage of the proposed georeferencing method, lies in its applicability to any existing camera setup, provided there are identifiable reference points on the surface to create the homography. 

In the following sections of this chapter, we explore my proposed method to use homography for the mapping of ships on water surfaces captured by monitoring camera images. 
We will delve into the detailed methodologies for recognizing and georeferencing ships on the water, further illustrating the practical implications and benefits of homography in real-world scenarios.

\section[Ship Detection and Georeferencing Using Homographies]{Ship Detection and Georeferencing Using Homographies~\ref{paper:1}}
\label{sec:qual_ship_geo}

As discussed in~\ref{sec:ship_det_abnormal}, the vessel detector plays a key role within the anomaly detection framework presented in~\ref{paper:1} as it identifies vessels and ships in video data, enabling accurate mapping of vessel locations using georeferencing. The motion detector leverages the ship's bounding box and uses optical flow for motion detection. Optical flow analyzes pixel intensity changes between sequential frames to quantify displacement vectors, indicating motion. I use this displacement to indicate the course of the ship (heading). 

Once the vessels are detected per video frame using YOLOv4-CSP~\cite{wang2021scaled} (see Sec. \ref{sec:ship_det_abnormal}), the pixel-based locations and course estimations (heading) are translated to a geographic coordinate system using a homography.
This georeferencing process allows for further analytics pursuing vessel abnormal behavior interpretation and for visualization on the situational awareness tool in~\ref{paper:1}.

\begin{figure}[h]
\includegraphics[width=1\columnwidth]{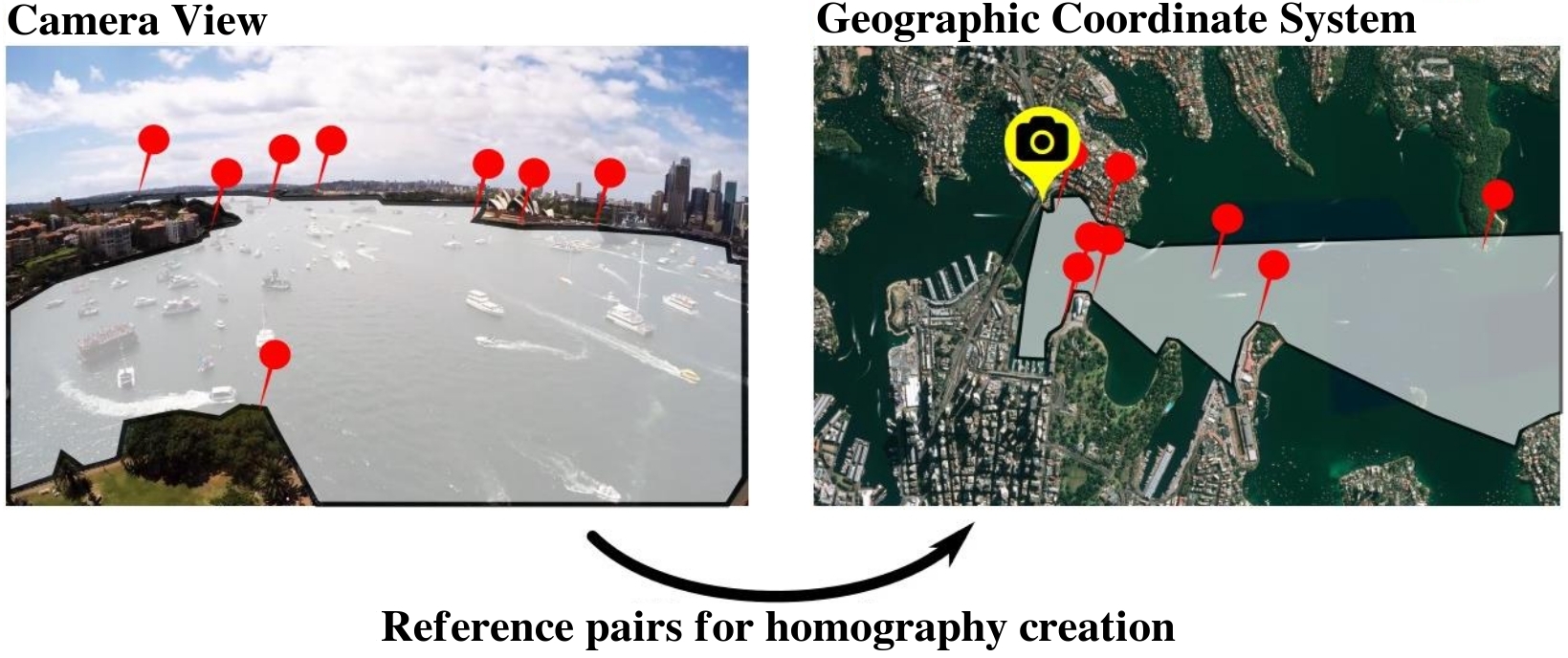}
\centering
\caption[Representation of the two planes used to create the homography of publication~\ref{paper:1}.]{Representation of the two planes used to create the homography of publication~\ref{paper:1}. Left: camera view. Right: geographic coordinate system. The white surfaces show the same water plane on both views. The red pins correspond with the selected points on both planes to calculate our homography matrix H. The yellow pin shows the location of the camera. Modified from~\ref{paper:1}. \textcopyright 2021 IEEE.}\label{fig:homography_paper1}
\end{figure}  

Given that the video used in~\ref{paper:1} has a static perspective, the pixels representing the water surface within its field of view can be treated as a planar area (as discussed in Section~\ref{sec:homo}), where all detections of vessels occur.
The generation of \gls{h} from these two planes is done with selected visual landmarks as reference points (see red pins in Fig.\ref{fig:homography_paper1}), that are used to solve the linear system specified by Eq.~\ref{eq:h_solve}. 
These references are pixel coordinates from the camera view and latitudes and longitudes, in decimal degrees, of the geographic coordinate system.
Once the homography is solved, by taking the center of the bounding boxes provided by the YOLOv4-CSP detector, the georeferencing of vessel positions is possible. 
The georeferencing, together with the YOLOv4-CSP object detector, allows the geo-location of observed anomalies in the scene and the calculation of vessel heading for their display on a map.

As discussed in Chapter~\ref{chap:ship_rec}, an additional contribution to~\ref{paper:1}, along with vessel detection and georeferencing, is the calculation of the heading direction of the vessels. The course of a vessel is defined by its steering direction with respect to the geographic north pole, also called the heading angle. 
\begin{figure}[h]
\includegraphics[width=1\columnwidth]{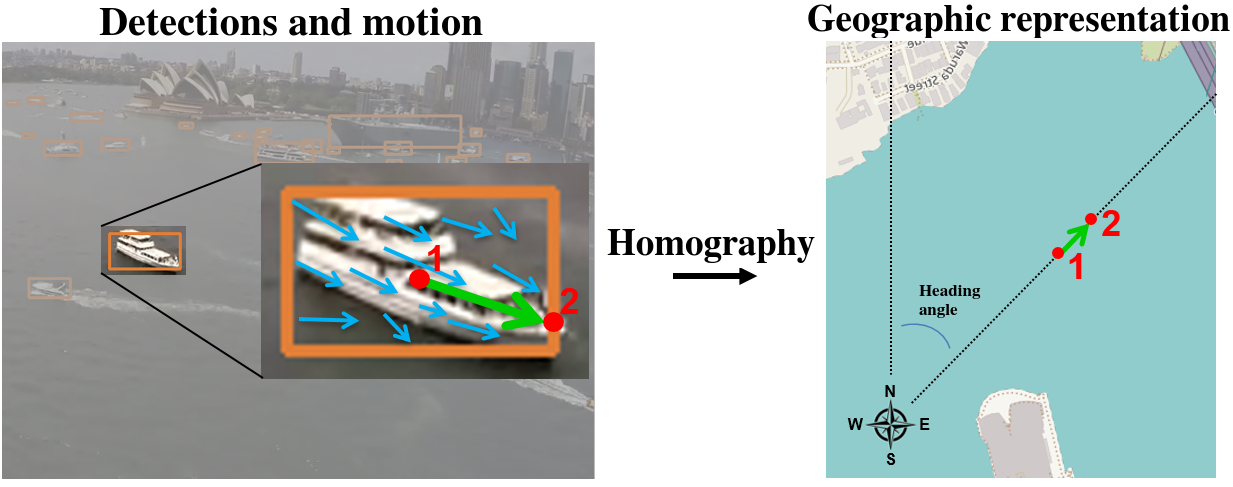}
\centering
\caption[Illustrative representation of detected and georeferenced vessel with heading.]{Illustrative representation of detected and georeferenced vessel with heading performed in~\ref{paper:1}. Left: example of detection using YOLOv4-CSP. The blue arrows represent displacement vectors (optical flow) with respect to the previous frame, and the green arrow the median direction within the bounding box. The red dots represent the georeferenced points and are defined by the median direction, from the center to the cutting point with the bounding box edge. Right: geographic representation (OpenStreetMap~\cite{OpenStreetMap}) of the vessel heading angle with respect to north using the created homography.}\label{fig:motion}
\end{figure}  

In the framework presented in~\ref{paper:1}, Brox's optical flow~\cite{brox2004high} is calculated to train the \gls{gan} that performs unsupervised anomaly detection. 
The optical flow was used to determine the direction of the displacement vectors of the detected bounding boxes, represented by blue arrows in Fig.~\ref{fig:motion} (left). This allows the estimation of the angle of the course of the vessels (heading) using georeferencing. 
From the displacement vectors, the main direction of motion is determined from the median of all displacement directions within the bounding box, represented in Fig.~\ref{fig:motion} by the green arrow. 
The homography created is then multiplied by the bounding box center and the tip of the displacement arrow to obtain their corresponding geographic coordinates.
The tip of the arrow is defined by the cutting point of the median displacement direction with the bounding box edge.
Once the two points (center and tip) are georeferenced, the heading angle ($\theta$) is calculated to obtain the course of the vessel using:

\begin{equation}
    \theta = \text{atan2}\left( \sin \Delta\lambda \cdot \cos \phi_2, \cos \phi_1 \cdot \sin \phi_2 - \sin \phi_1 \cdot \cos \phi_2 \cdot \cos \Delta\lambda \right)
\end{equation}

where ($\phi_1$, $\lambda_1$) and ($\phi_2$, $\lambda_2$) represent the coordinates of the georeferenced points (latitude and longitude, respectively), and $\Delta$ represents difference. 

\begin{figure}[h]
\includegraphics[width=9.5 cm]{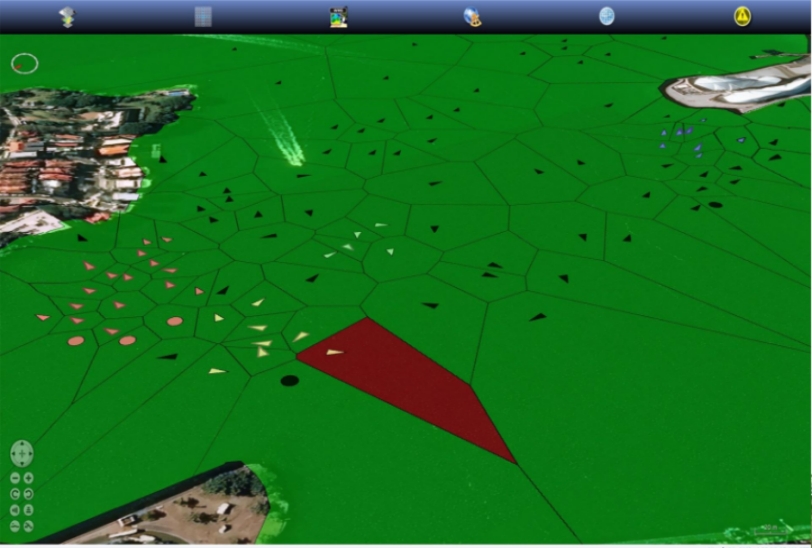}
\centering
\caption[Visualization of detected and georeferenced vessels with heading.]{Visualization of detected and georeferenced vessels with heading. The vessels detected are provided to a web service map (WorldWind~\cite{bell2007nasa}) in the form of triangles pointing towards their motion direction. Vessels for which the motion displacement was near zero are represented by circles. The rest of the figure represents the anomaly detection and visualization pipeline presented in~\ref{paper:1}. The red cell represents the interpreted area in which the anomalous ship is navigating.  Reprinted from~\ref{paper:1}. \textcopyright 2021 IEEE.}\label{fig:worldwind}
\end{figure}  

For display purposes, each vessel detected and georeferenced, is converted to an isosceles triangle, centered around the detection coordinate and whose vertex is pointing in the direction determined by the heading angle. 
Lastly, the three vertices are provided to a web service for map visualization (see Fig.~\ref{fig:worldwind}).

This section completes my contributions to~\ref{paper:1}.
These include vessel detection using YOLOv4-CSP (see Sec.\ref{sec:ship_det_abnormal}), vessel heading calculation using optical flow and georeference for map visualization using a homography.
These contributions are paramount to contextualize the identification of the abnormal vessel behaviour, providing useful geographic and spatial information regarding the anomaly.

Since ground truth latitudes and longitudes of the vessels present in the video were not available, the quantitative georeferencing error of the methodology was not reported in~\ref{paper:1}. 
The results, therefore, serve as a qualitative proof of concept of how ship recognition and georeferencing can improve maritime situational awareness. 
The lack of availability of ground truth for vessel geographic positions motivates the creation of ShipSG (see Chapter~\ref{chap:shipsg}), where vessel geographic positions (using \gls{ais}) were collected together with the images. 
The quantitative analysis of homographies for ship georeferencing is shown in the following section.

\section[Analysis of Ship Segmentation and Georeferencing Using Homographies]{Analysis of Ship Segmentation and Georeferencing Using Homographies~\ref{paper:2}~\ref{paper:5}}
\label{sec:quant_ship_geo}

As motivated in Chapter~\ref{chap:intro}, georeferencing results are superior from the mask of ships, due to the unnecessary background of bounding boxes as well as the inaccurate georeferencing result when using the bounding box center. 

I expand upon the georeferencing methodology of the Section~\ref{sec:qual_ship_geo} and show in depth quantitative studies of the use of homographies for ship georeferencing using ShipSG.
These studies use the resulting masks from Sec.~\ref{sec:std_ship_seg} and Sec.~\ref{sec:scatyolo}, based on~\ref{paper:2} and~\ref{paper:5}, respectively.

In Section~\ref{sec:std_ship_seg}, we have seen the results of an initial evaluation of different segmentation methods for the recognition of ships on the ShipSG dataset.
The annotation, using \gls{ais}, of latitudes and longitudes of ships present in the images allow to discuss now the evaluation of georeferencing from the resulting masks provided in~\ref{paper:2}.

Following the same principle presented in Section~\ref{sec:qual_ship_geo}, the ships, after being segmented, are georeferenced to provide their location to the situational awareness system in the form of latitude and longitude.
Since the views of the cameras on ShipSG are static, we can perform a transformation between the camera pixel coordinates ($C_x, C_y$) and Earth’s geographic latitude and longitude ($\phi, \lambda$) in decimal degrees using a homography. 

I took $200$ samples of the training set of ShipSG to create the homographies for the two camera views, and solved Equation \ref{eq:h_solve} to obtain H. 
The validation set is later used to quantitatively analyse how well the georeferencing performs.
The separation of homographies by high or low tides was not found to provide a significant improvement in results. Likewise, the correction of lens distortion prior to the homography calculation did not show an experimental improvement of the method. 
Therefore, due to their negligible impact, both tidal conditions and lens distortion were excluded from consideration in \ref{paper:2} and \ref{paper:5}.

Upon the creation of the homographies, I proposed a method to automatically determine the pixel ($C_x, C_y$) from the masks that most accurately represents the geographic position of the ship.

\begin{figure}[h]
\includegraphics[width=10.5 cm]{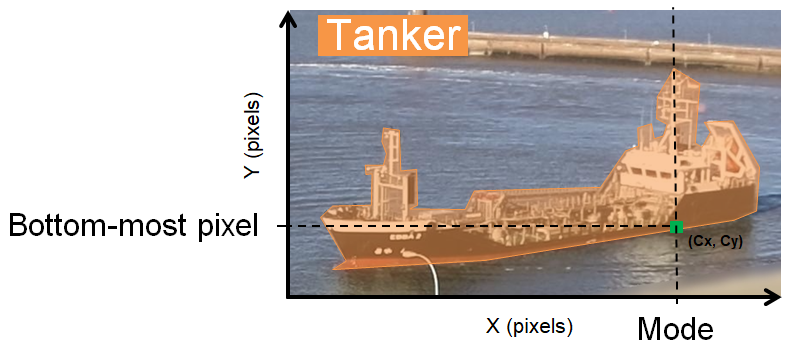}
\centering
\caption[Example of segmented ship mask with calculated pixel to be georeferenced.]{Example of segmented ship mask with calculated pixel to be georeferenced (in red, enlarged for visualization). Reprinted from [BCP-II] (CC BY 4.0).}\label{fig:georef}
\end{figure}  

The pixel to be georeferenced represents the intersection between the ship's hull and the waterline, located beneath the bridge or wheelhouse where the navigation antenna is positioned. 
This is achieved by identifying the bottom-most pixel within the ship mask in the vertical direction (Y) that corresponds to the statistical mode along the horizontal axis (X) (see Fig.~\ref{fig:georef}). 
Then, this pixel is georeferenced using the homography transformation defined in Equation~\ref{eq:p=hp}, facilitating the conversion of image pixel coordinates into real-world latitudes and longitudes.

For the evaluation, \ref{paper:2} uses the resulting masks of DetectoRS, which was the method that provided the best \gls{map} result during the initial study (74.7\%).

The true latitudes and longitudes obtained via AIS ($\phi_{AIS}$, $\lambda_{AIS}$) on the validation set of ShipSG, were quantitatively compared with those georeferenced via homography ($\phi_H$, $\lambda_H$). 
To facilitate this comparison, both sets of latitudes and longitudes were converted from decimal degrees to \gls{utm} coordinates, allowing for all results to be expressed in meters.
Among the metrics employed for the quantitative assessment in~\ref{paper:2}, the \gls{gde} is the one that best represents the accuracy of the method as it directly measures the distance in meters between the actual and estimated positions.
Therefore, the \gls{gde} measures the distance between true (\gls{ais}) and georeferenced (H) positions. 
The haversine equation (Eq.~\ref{eq:haversine}) is used instead of euclidean distance to take into account the radius (R) of the Earth
\begin{equation}
\label{eq:haversine}
GDE=2~\cdot R\cdot \arcsin\sqrt{\sin^2\dfrac{\abs{\phi_{AIS}-\phi_{H}}}{2}+\cos\phi_{AIS}\cdot \cos\phi_H\cdot \sin^2\dfrac{\abs{\lambda_{AIS}-\lambda_{H}}}{2}}
\end{equation}

where $\phi_{AIS}$ and $\phi_H$ represent the longitudes from the ground truth and georeferencing method using homography, respectively, $\lambda_{AIS}$ and $\lambda_H$ correspond to the latitudes, and R is the Earth's radius at Bremerhaven.

\begin{figure}[h]
\includegraphics[width=9 cm]{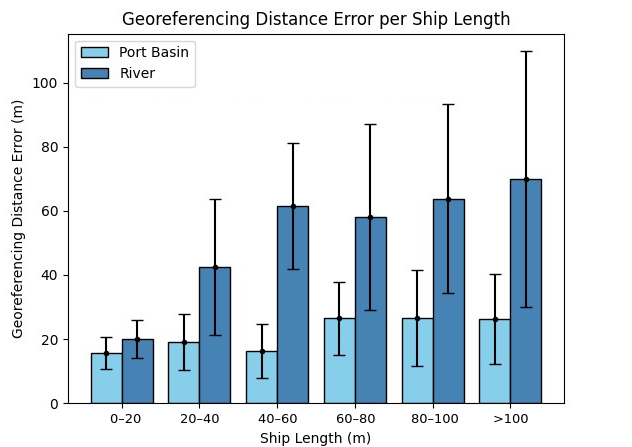}
\centering
\caption[Georeferencing distance error per ship length.]{Georeferencing distance error per ship length. GDEs and their uncertainties fall within the bounds of the ship length. Reprinted from [BCP-II] (CC BY 4.0).}\label{fig:gde}
\end{figure} 

The \gls{gde}, given as $mean~\pm~standard~deviation$ in meters, reaches $22~m~\pm~10~m$ for ranges inside the port basin (up to 400 m to the camera) and $53~m~\pm~24~m$ on the river (from 400 m to 1200 m).
In Chapter~\ref{chap:shipsg} it was described that ship lengths along with the ship
positions from \gls{ais} messages were collected and are used to observe how the \gls{gde} changes with ship length and range. 
As seen in Fig.~\ref{fig:gde}, for the smallest ship lengths (0 to 20 m), the \gls{gde} within the port basin and river are similar. 
This demonstrates that the accuracy of pinpointing the georeferenced pixel of the mask increases with the decrease in ship size, regardless of the distance between the ship and the camera.
Distance to the ship from the camera is the primary factor influencing the \gls{gde}.
For ships longer than 20 meters, a marked rise in \gls{gde} occurs at distances beyond 400 meters (on the river). 
In contrast, at distances shorter than 400 meters (within the port basin), the length of the ship has a smaller effect on the \gls{gde} compared to distances on the river beyond 400 meters.
Despite the challenge of identifying the precise pixel of the mask for georeferencing increasing with the distance to the ship, where each pixel spans a broader geographical area, the \gls{gde} remains consistent within uncertainties for each ship length. 
This suggests that the method provides estimations with a level of accuracy that can be considered contextually appropriate, when the specific operational contexts allow a a deviation of this magnitude within acceptable safety or operational thresholds.

\begin{figure}[h]
\includegraphics[width=1\columnwidth]{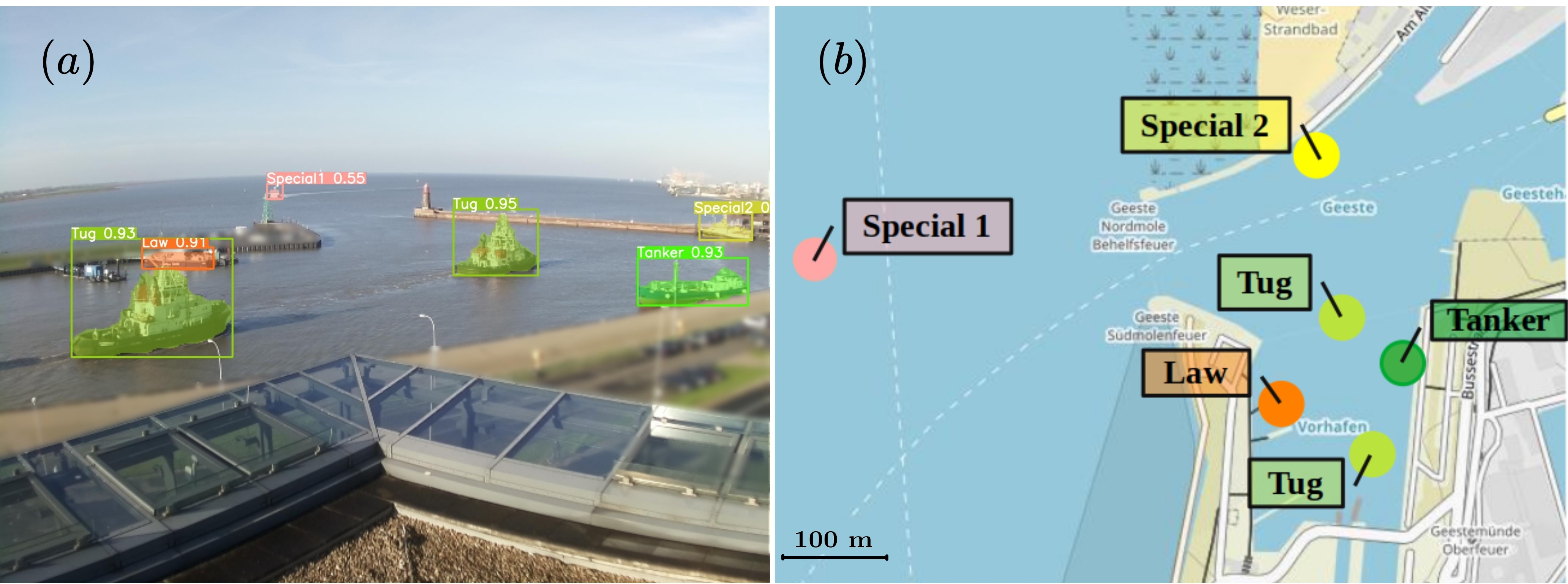}
\caption[Segmented and georeferenced ships using ScatYOLOv8+CBAM and homographies to improve maritime awareness.]{Segmented and georeferenced ships using ScatYOLOv8+CBAM and homographies to improve maritime awareness. (a) ShipSG image with segmented and classified ships using the ScatYOLOv8n+CBAM architecture presented in~\ref{paper:5}. (b) Georeferenced ships displayed on OpenStreetMap~\cite{OpenStreetMap} using the homography-based method of~\ref{paper:2} on segmented masks. Reprinted from~\ref{paper:5}. \textcopyright 2023 IEEE.}
\label{fig:res-georef}
\end{figure}

An additional quantitative georeferencing exploration, was done in~\ref{paper:5} with the resulting masks of the proposed ScatYOLOv8n+CBAM (see Sec.~\ref{sec:scatyolo}).
The georeferencing of the masks resulting from ScatYOLOv8+CBAM, studied in~\ref{paper:5}, follows the same procedure to automatically identify the pixel to be georeferenced, and with the same homographies calculated in the previous evaluation with standard methods.
The \gls{gde} yielded by the masks predicted with ScatYOLOv8+CBAM was of $18~m~\pm~13~m$ within the port basin (up to 400 m range). On the river (range from 400 m to 1200 m), the measured GDE using ScatYOLOv8+CBAM was $44~m~\pm~27~m$.
These results confirm the applicability of the optimized model allowing the benefits described in \ref{chap:adv_ship_rec} without any degradation in georeferencing performance.
As seen in Figure~\ref{fig:res-georef}, when the georeferenced masks are accessed using web services, this work can support real-time decision making, thus improving maritime awareness.

It is important to note the lack of complex operations in the georeferencing method, which allows an easy deployment on an embedded system.
Specifically, these georeferencing steps were measured 0.5 ms on average on the NVIDIA Jetson AGX Xavier, a stark contrast to the inference times of the instance segmentation. 
This shows that the bulk of the computation time is devoted to mask segmentation.
The subsequent pixel searching and coordinate transformation using a homography adds a minimal amount of time to the overall procedure. 

\begin{table}[h]
\caption{Comparison of the proposed method for ship georeferencing accuracy with existing works.}
\label{tab:new_geo_sota}
\resizebox{\textwidth}{!}{%
\begin{tabular}{cccc}
\hline
\textbf{Source} & \textbf{System} & \textbf{Range to Object} & \textbf{Error (m)} \\ \hline
\cite{naus2021assessment} & Radar Antenna + GPS & 1000 m & 6.5 \\ 
\cite{livingstone2014ship} & Synthetic Aperture Radar & 800 km & 13~$\pm$~23 \\
\cite{wei2020geolocation} & Opt. Remote Sensing & 36000 km & 165~$\pm$~109 \\
\cite{helgesen2020low} & Opt. Camera + GPS + IMU & 400 m & 20 \\ 
\ref{paper:2} & Opt. Camera & 400 m & 22$~\pm~$10~ \\
\ref{paper:2} & Opt. Camera & 1200 m & 53$~\pm~$24 \\
\ref{paper:5} & Opt. Camera & 400 m & 18$~\pm~$10 \\
\ref{paper:5}* & Opt. Camera & 1200 m & 44$~\pm~$27 \\\hline
\multicolumn{4}{l}{\footnotesize *Value not reported in \ref{paper:5} but calculated for this table.}
\end{tabular}%
}
\end{table}

Given the absence of directly comparable methodologies that simultaneously address the use of monocular cameras without prior camera pose knowledge for fast ship georeferencing, this approach establishes a benchmark in the literature. A comparison of the obtained results with existing ship georeferencing accuracies that use other technologies can be seen in Table~\ref{tab:new_geo_sota}, which is an updated version of Table~\ref{tab:geo_sota} for ship georeferencing. The most comparable set-up and result is given by \cite{helgesen2020low}, that utilizes prior knowledge of camera calibration, and its application was limited to controlled conditions with a single video sequence of a two small ships. Our method obtains similar positioning error, however providing a much more comprehensive study: results using two views, longer ranges, uncertainties, a large variety of ships of different categories and sizes, and does not need prior camera pose knowledge.

In this section, I expanded upon the georeferencing methodology of Section~\ref{sec:qual_ship_geo} to show quantitative results of the use of homographies for ship georeferencing on ShipSG, based on~\ref{paper:2} and~\ref{paper:5}.
The results prove that the approach provides useful information from the segmented ships to the situational awareness picture. 
This information involves the representation of the ship on a global scale using single images and without prior knowledge of the camera, standing out in the literature.

\section{Summary and Discussion}

Ship georeferencing using maritime footage, as discussed in this chapter, stands as a pivotal element in improving situational awareness. 
We have seen an in-depth exploration of ship georeferencing techniques using single images for enhancing maritime situational awareness, as presented in ~\ref{paper:1},~\ref{paper:2} and~\ref{paper:5}. 

This chapter started with a foundational understanding of homographies, essential for the proposed method and quantitative studies of ship georeferencing. 
Through the utilization of homographies, we could qualitatively and quantitatively explore the mapping of recognized ships from monocular images to geographic coordinates without previous knowledge about the camera. 
This is a significant benefit for its versatility; it can be applied any existing camera setup, as long as there are identifiable reference points on the surface to create the homography. 
This fundamental understanding provided the basis for subsequent discussions on ship detection, heading calculation using optical flow, and the quantitative analysis of ship segmentation and georeferencing using the ShipSG dataset.

We have seen the practical application of ship detection, georeferencing and heading to support abnormal behaviour detection in the maritime domain. 
The process allows, using a created homography from static camera view, to subsequently map ship locations to geographic coordinates and obtain the heading direction for visualization.

Bounding boxes, while useful for detection, introduce unnecessary background noise and inaccuracies, particularly when the center of the bounding box is used for georeferencing. 
Masks, on the other hand, delineate the precise contours of ships, providing a more precise representation for geospatial mapping. 
This precision is crucial for applications requiring exact geolocated data for improved situational awareness, such as assisted navigation, maritime monitoring or maritime safety and security operations.
Therefore, the complete pipeline first segments ships from images, yielding precise ship masks. 
The segmentation is followed by the automatic search of the pixel within the mask that intersects the ship hull and the water below, at the point where the navigation antenna is located on the ship. 
The pixel coordinates are then transformed into real-world geographic coordinates through a homography transformation. 

Quantitative analyses of ship segmentation and georeferencing using ShipSG uncovered promising accuracy, with a positioning error of $18~m~\pm~13~m$ m in range of up to 400 m and $44~m~\pm~27~m$ from 400 m to 1200 m.
The computational time for the pixel search and homography multiplication, averaging 0.5 milliseconds, is remarkably low in comparison to the instance segmentation inference times.
This efficiency contrasts with the time-intensive nature of instance segmentation inference on an embedded system, highlighting that the major computational demand lies in mask segmentation. 
The need for faster and more accurate ship segmentation was addressed by the proposed advanced methodologies for ship segmentation in Chapter~\ref{chap:adv_ship_rec}.

The analyses underscored the reliability and efficiency of the proposed georeferencing methods, laying the ground for their integration into a real-time maritime situational awareness system.
The georeferenced latitudes and longitudes of ships derived from this pipeline can be integrated into web services for real-time decision-making, significantly bolstering maritime situational awareness. 
This integration facilitates the dynamic monitoring of maritime traffic, enabling prompt responses to navigational hazards, environmental risks, and security threats.

In light of the state-of-the-art methodologies in ship georeferencing prior to this research, the proposed method stands out by diverging from traditional reliance on complex camera calibration, high-resolution orthophotos, or Digital Elevation Models (DEMs). The proposed approach provides a scalable, camera-agnostic solution that significantly advances applicability of maritime situational awareness technologies.

Looking ahead, future work in ship georeferencing could delve into further refining accuracy by incorporating additional ship annotations from the ShipSG dataset, such as keypoints and cuboids. 
These annotations could facilitate a more nuanced consideration of ship dimensions in the georeferencing process, potentially reducing error margins by accounting for the ship's length. 
Moreover, the exploration of cuboids in conjunction with ShipSG annotations opens new avenues for calculating ship orientations without relying on optical flow calculations. 
This approach could eliminate the need for sequential image analysis in anomaly detection, which demands higher frame rates, thereby streamlining the process of determining ship headings and orientations.
Future research could also benefit from analyzing sequences of images or videos and different camera perspectives of the same maritime scene, enhancing the depth and accuracy of situational awareness beyond the capabilities demonstrated with the ShipSG dataset.
Moreover, further fusion of the proposed image-based real-time ship georeferences with data and processes from multiple sensors, e.g. infrared imaging, radar and \gls{ais}, will provide a more comprehensive understanding of maritime situations. 
Overall, ship georeferencing represents a critical component in enhancing maritime situational awareness, with significant potential for further development and integration into maritime surveillance systems.

\chapter{Summary and Conclusion}
\label{chap:conclusion}

This compilation thesis addresses the enhancement of maritime situational awareness through advanced ship recognition and georeferencing methodologies. With the creation and utilization of the ShipSG dataset, we have established a benchmark in the field, which facilitates the development of recognition and georeferencing methodologies for maritime applications. This work is driven by the intuition that the integration of deep-learning-based object recognition methods, ship georeferencing and embedded systems can significantly advance the state of real-time maritime monitoring. The chapters within this thesis offer a progression from the creation of a foundational dataset to the implementation and deployment of real-time recognition (detection and segmentation) and georeferencing methods, following strategies to overcome the limitations of the previous existing literature. The key findings from each chapter are summarized below:

\textbf{Chapter \ref{chap:shipsg}} shows the creation of a novel dataset, ShipSG, for ship segmentation and georeferencing with two views of a maritime infrastructure. The dataset contains 3505 images and 11625 ship masks with their corresponding class, position and length. ShipSG marks a significant advancement in the field by setting a new standard for ship segmentation and georeferencing research. Moreover, it plays a crucial role in validating innovative methodologies for maritime situational awareness presented in this thesis. The dataset has facilitated the verification of recognition methods discussed in the chapters dedicated to ship recognition and advanced ship recognition, along with the quantitative validation of the proposed georeferencing methods. The contribution of ShipSG is underscored with its use in publications~\ref{paper:2}\ref{paper:3}\ref{paper:5}\ref{paper:2}, evidencing its importance and utility in pushing the boundaries of maritime situational awareness research.

\textbf{Chapter \ref{chap:ship_rec}} showcased the initial investigation in applying deep learning techniques for ship detection and segmentation within maritime applications to enhance situational awareness, underscoring the pivotal role of these methods in facilitating a range of applications, such as abnormal vessel behavior detection~\ref{paper:1},  camera integrity assessment~\ref{paper:3} and 3D ship reconstruction~\ref{paper:4}. 
Despite the demonstrated potential and success for ship detection in controlled or synthetic settings, the exploration of standard ship segmentation methods~\ref{paper:2} was highlighted for the superiority of instance segmentation over traditional bounding box detection in extracting detailed ship features crucial for applications like georeferencing. Challenges remained, notably in the precision, real-time processing, and the deployment of these technologies on GPU-powered embedded systems. The initial studies also pointed out a decrease in precision when segmenting small and distant ships, emphasizing the need for improved detection methods. Overall, while significant potential for enhancing maritime situational awareness has been revealed in this chapter, this exploration also uncovers the need for further development in real-time ship segmentation, the recognition of small and distant ships, and deployment on embedded systems, with subsequent chapters aiming to address these gaps through custom-tailored solutions for practical deployment in the maritime domain.

\textbf{Chapter \ref{chap:adv_ship_rec}} builds advances in the field real-time ship segmentation with the design of a customized architecture, ScatYOLOv8+CBAM~\ref{paper:5}, and demonstrated its enhanced performance on ShipSG (\gls{map} 75.46\%).
The chapter analyzed an optimization of the architecture~\ref{paper:6}, that removes the upsampling and downsampling from the ScatBlock to save computing time, and deployed it with TensoRT on the Jetson AGX Xavier to measure inference times for real-time applicability, which brought an acceleration of 36.5\% of the inference speed.
Moreover, the chapter proposed a batched-processing \gls{sahi} to increase the segmentation performance of small and distant ships that is able to run on embedded systems, emphasizing on the use of high-resolution images for a better understanding of the maritime situation. The \gls{map} in small ship segmentation compared to the baseline achieved an improvement of 8\% to 11\%, however resulted also in a slowdown in inference speed.
Choosing the optimal ScatYOLOv8+CBAM model size and incorporating \gls{sahi} depends on balancing real-time processing needs with computational capabilities. For critical applications like port surveillance, smaller configurations offer quick, accurate responses. However, larger models improve precision both in general and specifically in small and distant ships. Understanding specific needs and computational trade-offs is key to effective deployment.
The presented advances bridge the transition from standard methods to real-time instance segmentation on embedded systems, and addresses the ability to accurately identify all ships, independent from their size, and within the proximity of the port area.

\textbf{Chapter \ref{chap:ship_geo}} identifies ship georeferencing using maritime footage as a pivotal element for enhancing maritime situational awareness. This is done through the application of the ship georeferencing method developed in~\ref{paper:2} and further validated in~\ref{paper:5}. The chapter begins with a foundational understanding of homographies, crucial for the method, followed by qualitative and quantitative analysis of ship latitude and longitude positions from monocular images to geographic coordinates without prior camera knowledge. 
The qualitative analysis was used in a practical application; abnormal behavior detection in the maritime domain, where georeferences where used for ship positioning and heading direction, leveraging optical flow. 
In the quantitative analysis, the georeferencing, using ScatYOLOv8+CBAM masks, achieved a positioning error of $18~m~\pm~10~m$ for ranges inside the port basin (up to 400 m) and $44~m~\pm~27~m$ on the river (from 400 m to 1200 m), which improves upon existing results in the literature.
Moreover, the measured average time for the georeferencing, 0.5 ms per frame on the Jetson AGX Xavier, is not significant when compared to the instance segmentation timings.
The versatility of this approach and its applicability to any camera setup with identifiable surface reference points, sets a solid groundwork for further work on ship recognition and georeferencing, where the ShipSG dataset can be used as a benchmark. The real-world applicability of georeferenced ship coordinates to support real-time decision-making processes, alone or in combination with other sensors and processes, shows the significant potential of this research to enhance maritime situational awareness.\\

The outcome of this thesis reflects a significant advancement in maritime situational awareness through the deployment of novel methods both for real-time ship segmentation and georeferencing that are able to run on an embedded system. The introduction of the ScatYoloV8+CBAM architecture, optimized for the ShipSG dataset, demonstrates improved performance metrics over existing methodologies. 
Focusing on precise ship recognition and georeferencing regardless of class and size, the research has yielded a framework that, when applied, enhances the situational awareness of maritime stakeholders by displaying accurately located ships on digital maps, thereby consolidating knowledge into a user-friendly format which is easy to interpret and act upon.
The methodological choices have been centered on maximum accuracy and minimal processing times on embedded systems. This balance of speed and precision facilitates the integration with other sensor data and services, ultimately advancing maritime operations to be safer, more secure, and efficient. 
This rationale ensures that the presented work has tangible impacts.
This thesis has thus contributed to the body of knowledge with validated approaches that aim to facilitate monitoring in the vicinity of maritime infrastructures. Moreover, the methods and findings presented here offer a solid foundation for future research directed at refining high-resolution, real-time processing in resource-limited maritime contexts.

\chapter{Future Work}
\label{chap:future}

The outcomes and challenges presented in this thesis suggest several areas that are worth further research. 

With regards to the ShipSG dataset, future iterations of the dataset will focus on expanding its diversity by incorporating images from various cameras, and including higher resolution images, for more detailed analysis. This initiative aims to address the current limitations related to the variability and detail of maritime scenes, including more adverse weather conditions. Moreover, further improvements will involve leveraging \gls{ais} data for annotating ship heading, and enriching the dataset with additional annotations like ship cuboids or keypoints. These advancements will support the development of more sophisticated algorithms for automatically recognizing ship heading and dimensions, significantly benefiting maritime situational awareness research.

Focusing on improvements of ScatYOLOv8+CBAM and the overall performance of ship segmentation, key areas for advancement include refining the architecture to manage computational demands of larger models more effectively. Potential strategies for improvement involve making the ScatBlock learnable by introducing adjustable parameters within the wavelet filters, or enhancing the ScatBlock by adding learnable modules, such as strided convolutions and integration of transformers, to focus on significant features more efficiently. Additionally, optimizing the slicing and merging tasks in batch-inference processes through parallel processing techniques like multi-threading could significantly enhance efficiency when processing high-resolution images for small ship recognition. 

Future work in ship georeferencing should focus on enhancing accuracy by using additional annotations on the ShipSG dataset, specifically keypoints and cuboids, to refine the georeferencing process by considering ship dimensions. Exploring cuboids alongside ShipSG annotations could also streamline ship orientation calculations, potentially eliminating the need for high frame-rate sequential image analysis in anomaly detection pipelines. 

Finally, by integrating the methodologies presented in this thesis with additional processing chains and sensor data, we could contribute to the production of a broad and further enhanced situational awareness picture. This involves creating real-time visualizations of georeferenced ships using web services. For example, a dynamic map where ship positions update constantly, providing a clear picture of maritime situation. Furthermore, by merging diverse data sources such as ship tracking, 3D reconstruction, anomaly detection, and information from other sensors and data sources like thermal imaging, radar or \gls{ais}, we can gain a more comprehensive understanding of the maritime environment. This fusion of data will help authorities identify potential hazards, improve navigation safety, and optimize resource allocation for search and rescue operations.

\chapter{Publications by the Author for this Thesis}
\label{chap:publications}

This chapter presents a chronological list of the publications used for this compilation thesis, together with my shares on the works and a short summary of my contributions to each publication. 

\renewcommand{\thesection}{[BCP-\Roman{section}]}

\section[\\Detection and Geovisualization of Abnormal Vessel Behavior from Video]{}
\label{paper:1}

E. Solano, \textbf{B. Carrillo-Perez}, T. Flenker, Y. Steiniger, and J. Stoppe, \textbf{"Detection and Geovisualization of Abnormal Vessel Behavior from Video,"} in 2021 IEEE International Intelligent Transportation Systems Conference (ITSC), 2021, pp. 2193–2199.\\

My share on this publication is 35\%. Summary of contributions as $2^{nd}$ author:

\begin{itemize}
    \item Training and validation of YOLOv4 using custom dataset.
    \item Implementation of pipeline for ship detection and georeferencing from Sydney harbour video sequence.
    \item Development of a homography for ship georeferencing and adapted it for the pipeline.
    \item Calculation of ship course estimation of detected ships using optical flow and homography.
    \item I advised and participated in the writing of the manuscript.
    \item Outcome: Proof of concept for ship detection and georeferencing to improve maritime situational awareness.
    \item Next step: Quantitative analysis of the georeferencing method was not reported. Achieved in \ref{paper:2} and \ref{paper:5}.
\end{itemize}

\section[\\Ship Segmentation and Georeferencing from Static Oblique View Images]{}
\label{paper:2}

\textbf{B. Carrillo-Perez}, S. Barnes, and M. Stephan, \textbf{"Ship Segmentation and Georeferencing from Static Oblique View Images,"} Sensors, vol. 22, no. 7, p. 2713, Apr. 2022.\\

My share on this publication is 90\%. Summary of contributions as $1^{st}$ author:

\begin{itemize}
    \item Creation and publication of novel dataset for ship recognition and georeferencing (ShipSG).
    \item In-depth study of four state-of-the-art real-time ship segmentation methods, using ShipSG.
    \item Quantitative analysis of the homography-based ship georeferencing method, using ShipSG.
    \item Preparation of the manuscript.
    \item Outcome: Definition of a pipeline for ship segmentation and georeferencing for ship display on situational awareness maps.
    \item Next step: Deployment on embedded system was not reported. Achieved in \ref{paper:4}, \ref{paper:5} and \ref{paper:6}.
\end{itemize}

\section[\\Integrity Assessment of Maritime Object Detection Impacted by Partial Camera Obstruction]{}
\label{paper:3}

F. A. Costa de Oliveira, \textbf{B. Carrillo-Perez}, A. García-Ortiz, and F. Sill-Torres, \textbf{"Integrity Assessment of Maritime Object Detection Impacted by Partial Camera Obstruction,"} in 2023 IEEE International Conference on System Reliability and Safety (ICSRS), Nov. 2023, pp. 474–480.\\

My share on this publication is 20\%. Summary of contributions as $2^{nd}$ author:

\begin{itemize}
    \item Training and validation of Faster R-CNN for ship detection using ShipSG. The study of obstructions impact was also validated using ShipSG.
    \item I advised and participated in the writing of the manuscript.
    \item Outcome: ShipSG dataset impacts other applications for the improvement of maritime situational awareness such as camera integrity assessment.
\end{itemize}

\section[\\Embedded 3D reconstruction of Dynamic Objects in Real Time for Maritime Situational Awareness Pictures]{}
\label{paper:4}
F. Sattler, \textbf{B. Carrillo-Perez}, S. Barnes, K. Stebner, M. Stephan, and G. Lux, \textbf{“Embedded 3D reconstruction of Dynamic Objects in Real Time for Maritime Situational Awareness Pictures,”} The Visual Computer, Springer, pp. 1–14, 2023. \\

My share on this publication is 15\%. Summary of contributions as $2^{nd}$ author:

\begin{itemize}
    \item Training and validation of YOLOv5 for ship detection on synthetic dataset.
    \item Deployed detector on NVIDIA Jetson AGX Xavier, using Pytorch weights.
    \item I advised and participated in the writing of the manuscript.
    \item Outcome: Deployment of ship detector (bounding box) on embedded system for maritime situational awareness applications.
    \item Next step: Deployment of instance segmentation (mask) on embedded system with fast inference speed and high accuracy on real dataset (ShipSG). Achieved in \ref{paper:5} and \ref{paper:6}.
\end{itemize}

\section[\\Improving YOLOv8 with Scattering Transform and Attention for Maritime Awareness]{}
\label{paper:5}
\textbf{B. Carrillo-Perez}, A. Bueno Rodriguez, S. Barnes, and M. Stephan, \textbf{"Improving YOLOv8 with Scattering Transform and Attention for Maritime Awareness,"} in 2023 IEEE International Symposium on Image and Signal Processing and Analysis (ISPA), 2023, pp. 1–6.

My share on this publication is 90\%. Summary of contributions as $1^{st}$ author:

\begin{itemize}
    \item Improved YOLOv8 for ship segmentation with the novel addition of 2D scattering transform and attention mechanism to conform ScatYOLOv8+CBAM. The model was validated on ShipSG.
    \item Analysis of georeferencing results on ShipSG with the novel proposed architecture.
    \item Deployment of instance segmentation on embedded system (NVIDIA Jetson AGX Xavier).
    \item Preparation of the manuscript.
    \item Outcome: Faster and more accurate ship segmentation and georeferencing, deployed on embedded system.
    \item Next step: Optimize architecture further and use higher image resolutions, specially for small or distant ship segmentation. Achieved in \ref{paper:6}.
\end{itemize}

\section[\\Enhanced Small Ship Segmentation with Optimized ScatYOLOv8+CBAM on Embedded Systems]{}
\label{paper:6}

\textbf{B. Carrillo-Perez}, A. Bueno Rodriguez, S. Barnes, and M. Stephan, \textbf{"Enhanced Small Ship Segmentation with Optimized ScatYOLOv8+CBAM on Embedded Systems,"} 2024. IEEE International Conference on Real-time Computing and Robotics (RCAR), 2024, pp. 1–6. (Accepted)

My share on this publication is 90\%. Summary of contributions as $1^{st}$ author:

\begin{itemize}
    \item Optimization of ScatBlock for faster inference with ScatYOLOv8+CBAM.
    \item Comprehensive analysis of the use of ScatYOLOv8+CBAM for all model sizes.
    \item Improvement of the slice prediction method (\gls{sahi}) to perform inference in batches, focusing on small or distant ship segmentation using ShipSG.
    \item Weight serialization with TensorRT for Deployment on NVIDIA Jetson AGX Xavier, allowing faster inference.
    \item Preparation of the manuscript.
    \item Outcome: More efficient deployment of ScatYOLOv8+CBAM on embedded system and improved the small and distant ship segmentation accuracy.
\end{itemize}

\renewcommand{\thesection}{\arabic{section}}

\backmatter
\bibliographystyle{IEEEtran}
\bibliography{bibliography}

\begin{thebibliography}{100}
\providecommand{\url}[1]{#1}
\csname url@samestyle\endcsname
\providecommand{\newblock}{\relax}
\providecommand{\bibinfo}[2]{#2}
\providecommand{\BIBentrySTDinterwordspacing}{\spaceskip=0pt\relax}
\providecommand{\BIBentryALTinterwordstretchfactor}{4}
\providecommand{\BIBentryALTinterwordspacing}{\spaceskip=\fontdimen2\font plus
\BIBentryALTinterwordstretchfactor\fontdimen3\font minus
  \fontdimen4\font\relax}
\providecommand{\BIBforeignlanguage}[2]{{%
\expandafter\ifx\csname l@#1\endcsname\relax
\typeout{** WARNING: IEEEtran.bst: No hyphenation pattern has been}%
\typeout{** loaded for the language `#1'. Using the pattern for}%
\typeout{** the default language instead.}%
\else
\language=\csname l@#1\endcsname
\fi
#2}}
\providecommand{\BIBdecl}{\relax}
\BIBdecl

\bibitem{engler2018resiliencen}
E.~Engler, D.~G{\"o}ge, and S.~Brusch, ``Resiliencen--a multi-dimensional
  challenge for maritime infrastructures,'' \emph{NA{\v{S}}E MORE: znanstveni
  {\v{c}}asopis za more i pomorstvo}, vol.~65, no.~2, pp. 123--129, 2018.

\bibitem{torres2020indicator}
F.~S. Torres, N.~Kulev, B.~Skobiej, M.~Meyer, O.~Eichhorn, and
  J.~Sch{\"a}fer-Frey, ``Indicator-based safety and security assessment of
  offshore wind farms,'' in \emph{2020 Resilience Week (RWS)}.\hskip 1em plus
  0.5em minus 0.4em\relax IEEE, 2020, pp. 26--33.

\bibitem{cetin2013increasing}
F.~T. Cetin, B.~Yilmaz, Y.~Kabak, J.-H. Lee, C.~Erbas, E.~Akagunduz, S.-J. Lee,
  and A.~I.~A. (TURKEY), ``Increasing maritime situational awareness with
  interoperating distributed information sources,'' in \emph{18th Interantional
  Command and Control Research and Technology Symposium}, 2013, pp. 9--22.

\bibitem{ventikos2022risk}
N.~P. Ventikos and K.~Louzis, ``Risk dynamics for marine systems: towards a
  bio-inspired framework for dynamic risk assessment,'' \emph{Transportation
  Safety and Environment}, vol.~4, no.~3, p. tdac018, 2022.

\bibitem{wang2019maritime}
K.~Wang, M.~Liang, Y.~Li, J.~Liu, and R.~W. Liu, ``Maritime traffic data
  visualization: A brief review,'' in \emph{2019 IEEE 4th International
  Conference on Big Data Analytics (ICBDA)}.\hskip 1em plus 0.5em minus
  0.4em\relax IEEE, 2019, pp. 67--72.

\bibitem{belmoukari2023smart}
B.~Belmoukari, J.-F. Audy, and P.~Forget, ``Smart port: a systematic literature
  review,'' \emph{European Transport Research Review}, vol.~15, no.~1, p.~4,
  2023.

\bibitem{imo2015resolutionA110629}
\BIBentryALTinterwordspacing
{International Maritime Organization (IMO)}, ``Revised guidelines for the
  onboard operational use of shipborne automatic identification systems
  (ais),'' Resolution A.1106(29), December 2015, adopted on December 2, 2015,
  under agenda item 10 during the 29th session of the Assembly. [Online].
  Available:
  \url{https://wwwcdn.imo.org/localresources/en/OurWork/Safety/Documents/AIS/Resolution%20A.1106(29).pdf}
\BIBentrySTDinterwordspacing

\bibitem{kim2019adaptive}
\BIBentryALTinterwordspacing
K.-i. Kim and K.~M. Lee, ``Adaptive information visualization for maritime
  traffic stream sensor data with parallel context acquisition and machine
  learning,'' \emph{Sensors}, vol.~19, no. 5273, 2019. [Online]. Available:
  \url{https://www.mdpi.com/1424-8220/19/23/5273}
\BIBentrySTDinterwordspacing

\bibitem{jakovlev2020analysis}
S.~Jakovlev, A.~Daranda, M.~Voznak, A.~Lektauers, T.~Eglynas, and M.~Jusis,
  ``Analysis of the possibility to detect fake vessels in the automatic
  identification system,'' in \emph{2020 61st International Scientific
  Conference on Information Technology and Management Science of Riga Technical
  University (ITMS)}.\hskip 1em plus 0.5em minus 0.4em\relax IEEE, 2020, pp.
  1--5.

\bibitem{struck2021backwards}
M.~C. Struck and J.~Stoppe, ``A backwards compatible approach to authenticate
  automatic identification system messages,'' in \emph{2021 IEEE International
  Conference on Cyber Security and Resilience (CSR)}.\hskip 1em plus 0.5em
  minus 0.4em\relax IEEE, 2021, pp. 524--529.

\bibitem{wimpenny2018public}
G.~Wimpenny, J.~Safar, A.~Grant, M.~Bransby, and N.~Ward, ``Public key
  authentication for ais and the vhf data exchange system (vdes),'' in
  \emph{Proceedings of the 31st International Technical Meeting of the
  Satellite Division of The Institute of Navigation (ION GNSS+ 2018)}, 2018,
  pp. 1841--1851.

\bibitem{alincourt2016methodology}
E.~Alincourt, C.~Ray, P.-M. Ricordel, D.~Dare-Emzivat, and A.~Boudraa,
  ``Methodology for ais signature identification through magnitude and temporal
  characterization,'' in \emph{OCEANS 2016-Shanghai}.\hskip 1em plus 0.5em
  minus 0.4em\relax IEEE, 2016, pp. 1--6.

\bibitem{balduzzi2014security}
M.~Balduzzi, A.~Pasta, and K.~Wilhoit, ``A security evaluation of ais automated
  identification system,'' in \emph{Proceedings of the 30th annual computer
  security applications conference}, 2014, pp. 436--445.

\bibitem{yan2020exploring}
Z.~Yan, Y.~Xiao, L.~Cheng, R.~He, X.~Ruan, X.~Zhou, M.~Li, and R.~Bin,
  ``Exploring ais data for intelligent maritime routes extraction,''
  \emph{Applied Ocean Research}, vol. 101, p. 102271, 2020.

\bibitem{reggiannini2024remote}
M.~Reggiannini, E.~Salerno, C.~Bacciu, A.~D’Errico, A.~Lo~Duca, A.~Marchetti,
  M.~Martinelli, C.~Mercurio, A.~Mistretta, M.~Righi \emph{et~al.}, ``Remote
  sensing for maritime traffic understanding,'' \emph{Remote Sensing}, vol.~16,
  no.~3, p. 557, 2024.

\bibitem{schwarz2015near}
E.~Schwarz, D.~Krause, M.~Berg, H.~Daedelow, and H.~Maass, ``Near real time
  applications for maritime situational awareness,'' \emph{The International
  Archives of the Photogrammetry, Remote Sensing and Spatial Information
  Sciences}, vol.~40, pp. 999--1003, 2015.

\bibitem{prasad2017video}
D.~K. Prasad, D.~Rajan, L.~Rachmawati, E.~Rajabally, and C.~Quek, ``Video
  processing from electro-optical sensors for object detection and tracking in
  a maritime environment: A survey,'' \emph{IEEE Transactions on Intelligent
  Transportation Systems}, vol.~18, no.~8, pp. 1993--2016, 2017.

\bibitem{li2020causal}
F.~Li, C.-H. Chen, G.~Xu, D.~Chang, and L.~P. Khoo, ``Causal factors and
  symptoms of task-related human fatigue in vessel traffic service: A
  task-driven approach,'' \emph{The Journal of Navigation}, vol.~73, no.~6, pp.
  1340--1357, 2020.

\bibitem{flenker2021marlin}
T.~Flenker and J.~Stoppe, ``Marlin: An iot sensor network for improving
  maritime situational awareness,'' \emph{MARESEC 2021}, 2021.

\bibitem{chen2020deep}
Z.~Chen, D.~Chen, Y.~Zhang, X.~Cheng, M.~Zhang, and C.~Wu, ``Deep learning for
  autonomous ship-oriented small ship detection,'' \emph{Safety Science}, vol.
  130, p. 104812, 2020.

\bibitem{rekavandi2022guide}
A.~M. Rekavandi, L.~Xu, F.~Boussaid, A.-K. Seghouane, S.~Hoefs, and
  M.~Bennamoun, ``A guide to image and video based small object detection using
  deep learning: Case study of maritime surveillance,'' \emph{arXiv preprint
  arXiv:2207.12926}, 2022.

\bibitem{Hastings2009}
\BIBentryALTinterwordspacing
J.~T. Hastings and L.~L. Hill, \emph{Georeferencing}.\hskip 1em plus 0.5em
  minus 0.4em\relax Boston, MA: Springer US, 2009, pp. 1246--1249. [Online].
  Available: \url{https://doi.org/10.1007/978-0-387-39940-9_181}
\BIBentrySTDinterwordspacing

\bibitem{wawrzyniak2019vessel}
N.~Wawrzyniak, T.~Hyla, and A.~Popik, ``Vessel detection and tracking method
  based on video surveillance,'' \emph{Sensors}, vol.~19, no.~23, p. 5230,
  2019.

\bibitem{helgesen2020low}
{\O}.~K. Helgesen, E.~F. Brekke, A.~Stahl, and {\O}.~Engelhardtsen, ``Low
  altitude georeferencing for imaging sensors in maritime tracking,''
  \emph{IFAC-PapersOnLine}, vol.~53, no.~2, pp. 14\,476--14\,481, 2020.

\bibitem{mittal2019survey}
S.~Mittal, ``A survey on optimized implementation of deep learning models on
  the nvidia jetson platform,'' \emph{Journal of Systems Architecture},
  vol.~97, pp. 428--442, 2019.

\bibitem{zhao2019embedded}
H.~Zhao, W.~Zhang, H.~Sun, and B.~Xue, ``Embedded deep learning for ship
  detection and recognition,'' \emph{Future Internet}, vol.~11, no.~2, p.~53,
  2019.

\bibitem{ning2020heterogeneous}
H.~Ning, Y.~Li, F.~Shi, and L.~T. Yang, ``Heterogeneous edge computing open
  platforms and tools for internet of things,'' \emph{Future Generation
  Computer Systems}, vol. 106, pp. 67--76, 2020.

\bibitem{sattler2023maritime}
F.~Sattler, S.~Barnes, and M.~Stephan, ``A maritime situational awareness
  framework using dynamic 3d reconstruction in real-time,'' in \emph{2023 27th
  International Conference Information Visualisation (IV)}.\hskip 1em plus
  0.5em minus 0.4em\relax IEEE, 2023, pp. 334--339.

\bibitem{szeliski2022computer}
R.~Szeliski, \emph{Computer vision: algorithms and applications}.\hskip 1em
  plus 0.5em minus 0.4em\relax Springer Nature, 2022.

\bibitem{goodfellow2016deep}
I.~Goodfellow, Y.~Bengio, and A.~Courville, \emph{Deep learning}.\hskip 1em
  plus 0.5em minus 0.4em\relax MIT press, 2016.

\bibitem{talaei2023deep}
T.~Talaei~Khoei, H.~Ould~Slimane, and N.~Kaabouch, ``Deep learning: Systematic
  review, models, challenges, and research directions,'' \emph{Neural Computing
  and Applications}, vol.~35, no.~31, pp. 23\,103--23\,124, 2023.

\bibitem{sammut2011encyclopedia}
C.~Sammut and G.~I. Webb, \emph{Encyclopedia of machine learning}.\hskip 1em
  plus 0.5em minus 0.4em\relax Springer Science \& Business Media, 2011.

\bibitem{chai2021deep}
J.~Chai, H.~Zeng, A.~Li, and E.~W. Ngai, ``Deep learning in computer vision: A
  critical review of emerging techniques and application scenarios,''
  \emph{Machine Learning with Applications}, vol.~6, p. 100134, 2021.

\bibitem{Cunningham2008}
\BIBentryALTinterwordspacing
P.~Cunningham, M.~Cord, and S.~J. Delany, \emph{Supervised Learning}.\hskip 1em
  plus 0.5em minus 0.4em\relax Berlin, Heidelberg: Springer Berlin Heidelberg,
  2008, pp. 21--49. [Online]. Available:
  \url{https://doi.org/10.1007/978-3-540-75171-7_2}
\BIBentrySTDinterwordspacing

\bibitem{rawat2017deep}
W.~Rawat and Z.~Wang, ``Deep convolutional neural networks for image
  classification: A comprehensive review,'' \emph{Neural computation}, vol.~29,
  no.~9, pp. 2352--2449, 2017.

\bibitem{zaidi2022survey}
S.~S.~A. Zaidi, M.~S. Ansari, A.~Aslam, N.~Kanwal, M.~Asghar, and B.~Lee, ``A
  survey of modern deep learning based object detection models,'' \emph{Digital
  Signal Processing}, vol. 126, p. 103514, 2022.

\bibitem{shanmugamani2018deep}
R.~Shanmugamani, \emph{Deep Learning for Computer Vision: Expert techniques to
  train advanced neural networks using TensorFlow and Keras}.\hskip 1em plus
  0.5em minus 0.4em\relax Packt Publishing Ltd, 2018.

\bibitem{lecun2015deep}
Y.~LeCun, Y.~Bengio, and G.~Hinton, ``Deep learning,'' \emph{Nature}, vol. 521,
  no. 7553, pp. 436--444, 2015.

\bibitem{zhang2020lightweight}
J.~Zhang, X.~Li, L.~Li, P.~Sun, X.~Su, T.~Hu, and F.~Chen, ``Lightweight u-net
  for cloud detection of visible and thermal infrared remote sensing images,''
  \emph{Optical and Quantum Electronics}, vol.~52, pp. 1--14, 2020.

\bibitem{he2016deep}
K.~He, X.~Zhang, S.~Ren, and J.~Sun, ``Deep residual learning for image
  recognition,'' in \emph{Proceedings of the IEEE conference on computer vision
  and pattern recognition}, 2016, pp. 770--778.

\bibitem{lin2017feature}
T.-Y. Lin, P.~Doll{\'a}r, R.~Girshick, K.~He, B.~Hariharan, and S.~Belongie,
  ``Feature pyramid networks for object detection,'' in \emph{Proceedings of
  the IEEE conference on computer vision and pattern recognition}, 2017, pp.
  2117--2125.

\bibitem{vaswani2017attention}
A.~Vaswani, N.~Shazeer, N.~Parmar, J.~Uszkoreit, L.~Jones, A.~N. Gomez,
  {\L}.~Kaiser, and I.~Polosukhin, ``Attention is all you need,''
  \emph{Advances in neural information processing systems}, vol.~30, 2017.

\bibitem{bahdanau2014neural}
D.~Bahdanau, K.~Cho, and Y.~Bengio, ``Neural machine translation by jointly
  learning to align and translate,'' \emph{arXiv preprint arXiv:1409.0473},
  2014.

\bibitem{Luong2015}
M.-T. Luong, H.~Pham, and C.~D. Manning, ``Effective approaches to
  attention-based neural machine translation,'' \emph{arXiv preprint
  arXiv:1508.04025}, 2015.

\bibitem{woo2018cbam}
S.~Woo, J.~Park, J.-Y. Lee, and I.~S. Kweon, ``Cbam: Convolutional block
  attention module,'' in \emph{Proceedings of the European conference on
  computer vision (ECCV)}, 2018, pp. 3--19.

\bibitem{guo2022attention}
M.-H. Guo, T.-X. Xu, J.-J. Liu, Z.-N. Liu, P.-T. Jiang, T.-J. Mu, S.-H. Zhang,
  R.~R. Martin, M.-M. Cheng, and S.-M. Hu, ``Attention mechanisms in computer
  vision: A survey,'' \emph{Computational visual media}, vol.~8, no.~3, pp.
  331--368, 2022.

\bibitem{Hu2018}
J.~Hu, L.~Shen, and G.~Sun, ``Squeeze-and-excitation networks,'' in
  \emph{Proceedings of the IEEE Conference on Computer Vision and Pattern
  Recognition}, 2018, pp. 7132--7141.

\bibitem{Yang2016}
Z.~Yang, D.~Yang, C.~Dyer, X.~He, A.~Smola, and E.~Hovy, ``Hierarchical
  attention networks for document classification,'' in \emph{Proceedings of the
  2016 Conference of the North American Chapter of the Association for
  Computational Linguistics: Human Language Technologies}, 2016, pp.
  1480--1489.

\bibitem{Parmar2018}
N.~Parmar, A.~Vaswani, J.~Uszkoreit, L.~Kaiser, N.~Shazeer, A.~Ku, and D.~Tran,
  ``Image transformer,'' in \emph{International Conference on Machine
  Learning}.\hskip 1em plus 0.5em minus 0.4em\relax PMLR, 2018, pp. 4055--4064.

\bibitem{Xu2015}
K.~Xu, J.~Ba, R.~Kiros, K.~Cho, A.~Courville, R.~Salakhutdinov, R.~Zemel, and
  Y.~Bengio, ``Show, attend and tell: Neural image caption generation with
  visual attention,'' \emph{arXiv preprint arXiv:1502.03044}, 2015.

\bibitem{Wang2017}
F.~Wang, M.~Jiang, C.~Qian, S.~Yang, C.~Li, H.~Zhang, X.~Wang, and X.~Tang,
  ``Residual attention network for image classification,'' \emph{arXiv preprint
  arXiv:1704.06904}, 2017.

\bibitem{Chen2017}
L.~Chen, H.~Zhang, J.~Xiao, L.~Nie, J.~Shao, and T.-S. Chua, ``Sca-cnn: Spatial
  and channel-wise attention in convolutional networks for image captioning,''
  in \emph{Proceedings of the IEEE Conference on Computer Vision and Pattern
  Recognition (CVPR)}, 2017.

\bibitem{brauwers2021general}
G.~Brauwers and F.~Frasincar, ``A general survey on attention mechanisms in
  deep learning,'' \emph{IEEE Transactions on Knowledge and Data Engineering},
  2021.

\bibitem{gong2021review}
M.~Gong, D.~Wang, X.~Zhao, H.~Guo, D.~Luo, and M.~Song, ``A review of
  non-maximum suppression algorithms for deep learning target detection,'' in
  \emph{Seventh Symposium on Novel Photoelectronic Detection Technology and
  Applications}, vol. 11763.\hskip 1em plus 0.5em minus 0.4em\relax SPIE, 2021,
  pp. 821--828.

\bibitem{NEURIPS2019_9015}
\BIBentryALTinterwordspacing
A.~Paszke, S.~Gross, F.~Massa, A.~Lerer, J.~Bradbury, G.~Chanan, T.~Killeen,
  Z.~Lin, N.~Gimelshein, L.~Antiga, A.~Desmaison, A.~Kopf, E.~Yang, Z.~DeVito,
  M.~Raison, A.~Tejani, S.~Chilamkurthy, B.~Steiner, L.~Fang, J.~Bai, and
  S.~Chintala, ``Pytorch: An imperative style, high-performance deep learning
  library,'' in \emph{Advances in Neural Information Processing Systems
  32}.\hskip 1em plus 0.5em minus 0.4em\relax Curran Associates, Inc., 2019,
  pp. 8024--8035. [Online]. Available:
  \url{http://papers.neurips.cc/paper/9015-pytorch-an-imperative-style-high-performance-deep-learning-library.pdf}
\BIBentrySTDinterwordspacing

\bibitem{lin2014microsoft}
\BIBentryALTinterwordspacing
T.-Y. Lin, M.~Maire, S.~Belongie, L.~Bourdev, R.~Girshick, J.~Hays, P.~Perona,
  D.~Ramanan, C.~L. Zitnick, and P.~Dollár, ``Microsoft coco: Common objects
  in context,'' 2014. [Online]. Available: \url{http://arxiv.org/abs/1405.0312}
\BIBentrySTDinterwordspacing

\bibitem{sarker2021deep}
I.~H. Sarker, ``Deep learning: a comprehensive overview on techniques,
  taxonomy, applications and research directions,'' \emph{SN Computer Science},
  vol.~2, no.~6, p. 420, 2021.

\bibitem{qiao2021marine}
D.~Qiao, G.~Liu, T.~Lv, W.~Li, and J.~Zhang, ``Marine vision-based situational
  awareness using discriminative deep learning: A survey,'' \emph{Journal of
  Marine Science and Engineering}, vol.~9, no.~4, p. 397, 2021.

\bibitem{everingham2015pascal}
M.~Everingham, S.~A. Eslami, L.~Van~Gool, C.~K. Williams, J.~Winn, and
  A.~Zisserman, ``The pascal visual object classes challenge: A
  retrospective,'' \emph{International journal of computer vision}, vol. 111,
  no.~1, pp. 98--136, 2015.

\bibitem{shao2018seaships}
Z.~Shao, W.~Wu, Z.~Wang, W.~Du, and C.~Li, ``Seaships: A large-scale precisely
  annotated dataset for ship detection,'' \emph{IEEE transactions on
  multimedia}, vol.~20, no.~10, pp. 2593--2604, 2018.

\bibitem{chen2020video}
X.~Chen, L.~Qi, Y.~Yang, Q.~Luo, O.~Postolache, J.~Tang, and H.~Wu,
  ``Video-based detection infrastructure enhancement for automated ship
  recognition and behavior analysis,'' \emph{Journal of Advanced
  Transportation}, vol. 2020, 2020.

\bibitem{ghahremani2019multi}
A.~Ghahremani, Y.~Kong, E.~Bondarev \emph{et~al.}, ``Multi-class detection and
  orientation recognition of vessels in maritime surveillance,''
  \emph{Electronic Imaging}, vol. 2019, no.~11, pp. 266--1, 2019.

\bibitem{nita2020cnn}
C.~Nita and M.~Vandewal, ``Cnn-based object detection and segmentation for
  maritime domain awareness,'' in \emph{Artificial Intelligence and Machine
  Learning in Defense Applications II}, vol. 11543.\hskip 1em plus 0.5em minus
  0.4em\relax International Society for Optics and Photonics, 2020, p. 1154306.

\bibitem{ribeiro2022real}
M.~Ribeiro, B.~Damas, and A.~Bernardino, ``Real-time ship segmentation in
  maritime surveillance videos using automatically annotated synthetic
  datasets,'' \emph{Sensors}, vol.~22, no.~21, p. 8090, 2022.

\bibitem{teixeira2022literature}
E.~Teixeira, B.~Araujo, V.~Costa, S.~Mafra, and F.~Figueiredo, ``Literature
  review on ship localization, classification, and detection methods based on
  optical sensors and neural networks,'' \emph{Sensors}, vol.~22, no.~18, p.
  6879, 2022.

\bibitem{jocheryolov8}
\BIBentryALTinterwordspacing
G.~Jocher, A.~Chaurasia, and J.~Qiu, \emph{{YOLOv8 by Ultralytics}}, Jan. 2023.
  [Online]. Available: \url{https://github.com/ultralytics/ultralytics}
\BIBentrySTDinterwordspacing

\bibitem{wang2021scaled}
C.-Y. Wang, A.~Bochkovskiy, and H.-Y.~M. Liao, ``Scaled-yolov4: Scaling cross
  stage partial network,'' in \emph{Proceedings of the IEEE/cvf conference on
  computer vision and pattern recognition}, 2021, pp. 13\,029--13\,038.

\bibitem{redmon2016you}
J.~Redmon, S.~Divvala, R.~Girshick, and A.~Farhadi, ``You only look once:
  Unified, real-time object detection,'' in \emph{Proceedings of the IEEE
  conference on computer vision and pattern recognition}, 2016, pp. 779--788.

\bibitem{bochkovskiy2020yolov4}
A.~Bochkovskiy, C.-Y. Wang, and H.-Y.~M. Liao, ``Yolov4: Optimal speed and
  accuracy of object detection,'' \emph{arXiv preprint arXiv:2004.10934}, 2020.

\bibitem{wang2020cspnet}
C.-Y. Wang, H.-Y.~M. Liao, Y.-H. Wu, P.-Y. Chen, J.-W. Hsieh, and I.-H. Yeh,
  ``Cspnet: A new backbone that can enhance learning capability of cnn,'' in
  \emph{Proceedings of the IEEE/CVF conference on computer vision and pattern
  recognition workshops}, 2020, pp. 390--391.

\bibitem{ren2015faster}
S.~Ren, K.~He, R.~Girshick, and J.~Sun, ``Faster r-cnn: Towards real-time
  object detection with region proposal networks,'' \emph{Advances in neural
  information processing systems}, vol.~28, pp. 91--99, 2015.

\bibitem{he2017mask}
K.~He, G.~Gkioxari, P.~Dollár, and R.~Girshick, ``Mask r-cnn,'' in \emph{2017
  IEEE International Conference on Computer Vision (ICCV)}, 2017, pp.
  2980--2988.

\bibitem{xie2017aggregated}
S.~Xie, R.~Girshick, P.~Doll{\'a}r, Z.~Tu, and K.~He, ``Aggregated residual
  transformations for deep neural networks,'' in \emph{Proceedings of the IEEE
  conference on computer vision and pattern recognition}, 2017, pp. 1492--1500.

\bibitem{qiao2021detectors}
S.~Qiao, L.-C. Chen, and A.~Yuille, ``Detectors: Detecting objects with
  recursive feature pyramid and switchable atrous convolution,'' in
  \emph{Proceedings of the IEEE/CVF Conference on Computer Vision and Pattern
  Recognition}, 2021, pp. 10\,213--10\,224.

\bibitem{NIPS2014_19de10ad}
\BIBentryALTinterwordspacing
M.~F. Stollenga, J.~Masci, F.~Gomez, and J.~Schmidhuber, ``Deep networks with
  internal selective attention through feedback connections,'' in
  \emph{Advances in Neural Information Processing Systems}, Z.~Ghahramani,
  M.~Welling, C.~Cortes, N.~Lawrence, and K.~Weinberger, Eds., vol.~27.\hskip
  1em plus 0.5em minus 0.4em\relax Curran Associates, Inc., 2014. [Online].
  Available:
  \url{https://proceedings.neurips.cc/paper_files/paper/2014/file/19de10adbaa1b2ee13f77f679fa1483a-Paper.pdf}
\BIBentrySTDinterwordspacing

\bibitem{yu2015multi}
F.~Yu and V.~Koltun, ``Multi-scale context aggregation by dilated
  convolutions,'' \emph{arXiv preprint arXiv:1511.07122}, 2015.

\bibitem{bolya2019yolact}
D.~Bolya, C.~Zhou, F.~Xiao, and Y.~J. Lee, ``Yolact: Real-time instance
  segmentation,'' in \emph{Proceedings of the IEEE/CVF international conference
  on computer vision}, 2019, pp. 9157--9166.

\bibitem{long2015fully}
J.~Long, E.~Shelhamer, and T.~Darrell, ``Fully convolutional networks for
  semantic segmentation,'' in \emph{Proceedings of the IEEE conference on
  computer vision and pattern recognition}, 2015, pp. 3431--3440.

\bibitem{lee2020centermask}
Y.~Lee and J.~Park, ``Centermask: Real-time anchor-free instance
  segmentation,'' in \emph{Proceedings of the IEEE/CVF conference on computer
  vision and pattern recognition}, 2020, pp. 13\,906--13\,915.

\bibitem{guo2020fully}
Y.~Guo, F.~Chen, Q.~Cheng, J.~Wu, B.~Wang, Y.~Wu, and W.~Zhao, ``Fully
  convolutional one-stage circular object detector on medical images,'' in
  \emph{2020 4th International Conference on Advances in Image Processing},
  2020, pp. 21--26.

\bibitem{lee2019energy}
Y.~Lee, J.-w. Hwang, S.~Lee, Y.~Bae, and J.~Park, ``An energy and
  gpu-computation efficient backbone network for real-time object detection,''
  in \emph{Proceedings of the IEEE/CVF Conference on Computer Vision and
  Pattern Recognition Workshops}, 2019, pp. 0--0.

\bibitem{jocheryolov5}
\BIBentryALTinterwordspacing
G.~Jocher, A.~Chaurasia, A.~Stoken, J.~Borovec, NanoCode012, Y.~Kwon,
  K.~Michael, TaoXie, J.~Fang, imyhxy, Lorna, Z.~Yifu, C.~Wong, A.~V,
  D.~Montes, Z.~Wang, C.~Fati, J.~Nadar, Laughing, UnglvKitDe, V.~Sonck,
  tkianai, yxNONG, P.~Skalski, A.~Hogan, D.~Nair, M.~Strobel, and M.~Jain,
  \emph{ultralytics/yolov5: v7.0 - YOLOv5 SOTA Realtime Instance Segmentation},
  Nov. 2022. [Online]. Available: \url{https://doi.org/10.5281/zenodo.7347926}
\BIBentrySTDinterwordspacing

\bibitem{singh2021deep}
P.~Singh, G.~Saha, and M.~Sahidullah, ``Deep scattering network for speech
  emotion recognition,'' in \emph{2021 29th European Signal Processing
  Conference (EUSIPCO)}.\hskip 1em plus 0.5em minus 0.4em\relax IEEE, 2021, pp.
  131--135.

\bibitem{pan2020spatio}
C.~Pan, S.~Chen, and A.~Ortega, ``Spatio-temporal graph scattering transform,''
  \emph{arXiv preprint arXiv:2012.03363}, 2020.

\bibitem{cheng2020new}
S.~Cheng, Y.-S. Ting, B.~M{\'e}nard, and J.~Bruna, ``A new approach to
  observational cosmology using the scattering transform,'' \emph{Monthly
  Notices of the Royal Astronomical Society}, vol. 499, no.~4, pp. 5902--5914,
  2020.

\bibitem{rodriguez2021recurrent}
{\'A}.~B. Rodr{\'\i}guez, R.~Balestriero, S.~De~Angelis, M.~C. Ben{\'\i}tez,
  L.~Zuccarello, R.~Baraniuk, J.~M. Ibanez, and M.~V. de~Hoop, ``Recurrent
  scattering network detects metastable behavior in polyphonic seismo-volcanic
  signals for volcano eruption forecasting,'' \emph{IEEE Transactions on
  Geoscience and Remote Sensing}, vol.~60, pp. 1--23, 2021.

\bibitem{bruna2013invariant}
J.~Bruna and S.~Mallat, ``Invariant scattering convolution networks,''
  \emph{IEEE transactions on pattern analysis and machine intelligence},
  vol.~35, no.~8, pp. 1872--1886, 2013.

\bibitem{oyallon2018compressing}
E.~Oyallon, E.~Belilovsky, S.~Zagoruyko, and M.~Valko, ``Compressing the input
  for cnns with the first-order scattering transform,'' in \emph{Proceedings of
  the European Conference on Computer Vision (ECCV)}, 2018, pp. 301--316.

\bibitem{saponara2021impact}
S.~Saponara and A.~Elhanashi, ``Impact of image resizing on deep learning
  detectors for training time and model performance,'' in \emph{International
  Conference on Applications in Electronics Pervading Industry, Environment and
  Society}.\hskip 1em plus 0.5em minus 0.4em\relax Springer, 2021, pp. 10--17.

\bibitem{mao2016towards}
H.~Mao, S.~Yao, T.~Tang, B.~Li, J.~Yao, and Y.~Wang, ``Towards real-time object
  detection on embedded systems,'' \emph{IEEE Transactions on Emerging Topics
  in Computing}, vol.~6, no.~3, pp. 417--431, 2016.

\bibitem{wang2023uav}
G.~Wang, Y.~Chen, P.~An, H.~Hong, J.~Hu, and T.~Huang, ``Uav-yolov8: a
  small-object-detection model based on improved yolov8 for uav aerial
  photography scenarios,'' \emph{Sensors}, vol.~23, no.~16, p. 7190, 2023.

\bibitem{zhu2023biformer}
L.~Zhu, X.~Wang, Z.~Ke, W.~Zhang, and R.~W. Lau, ``Biformer: Vision transformer
  with bi-level routing attention,'' in \emph{Proceedings of the IEEE/CVF
  Conference on Computer Vision and Pattern Recognition}, 2023, pp.
  10\,323--10\,333.

\bibitem{sahi23akyon}
F.~C. Akyon, S.~Onur~Altinuc, and A.~Temizel, ``Slicing aided hyper inference
  and fine-tuning for small object detection,'' in \emph{2022 IEEE
  International Conference on Image Processing (ICIP)}, 2022, pp. 966--970.

\bibitem{han2011geolocation}
K.~M. Han and G.~N. DeSouza, ``Geolocation of multiple targets from airborne
  video without terrain data,'' \emph{Journal of Intelligent \& Robotic
  Systems}, vol.~62, no.~1, pp. 159--183, 2011.

\bibitem{shami2024geo}
M.~B. Shami, G.~Kiss, T.~A. Haakonsen, and F.~Lindseth, ``Geo-locating road
  objects using inverse haversine formula with nvidia driveworks,'' \emph{arXiv
  preprint arXiv:2401.07582}, 2024.

\bibitem{milosavljevic2017method}
A.~Milosavljevi{\'c}, D.~Ran{\v{c}}i{\'c}, A.~Dimitrijevi{\'c}, B.~Predi{\'c},
  and V.~Mihajlovi{\'c}, ``A method for estimating surveillance video
  georeferences,'' \emph{ISPRS international journal of geo-information},
  vol.~6, no.~7, p. 211, 2017.

\bibitem{naus2021assessment}
K.~Naus, M.~W, P.~Szymak, L.~Gucma, and M.~Gucma, ``Assessment of ship position
  estimation accuracy based on radar navigation mark echoes identified in an
  electronic navigational chart,'' \emph{Measurement}, vol. 169, p. 108630,
  2020.

\bibitem{livingstone2014ship}
C.~E. Livingstone, M.~Dragosevic, S.~Chu, and I.~Sikaneta, \emph{Ship detection
  and measurement of ship motion by multi-aperture Synthetic Aperture
  Radar}.\hskip 1em plus 0.5em minus 0.4em\relax Defence Research and
  Development Canada, 2014.

\bibitem{wei2020geolocation}
Y.~Wei, Z.~Zhang, B.~Mu, Y.~Li, Q.~Wang, and R.~Liu, ``Geolocation accuracy
  evaluation of gf-4 geostationary high-resolution optical images over coastal
  zones and offshore areas,'' \emph{Journal of Coastal Research}, vol. 102,
  no.~SI, pp. 326--333, 2020.

\bibitem{zhang2000flexible}
Z.~Zhang, ``A flexible new technique for camera calibration,'' \emph{IEEE
  Transactions on pattern analysis and machine intelligence}, vol.~22, no.~11,
  pp. 1330--1334, 2000.

\bibitem{chen2019deep}
J.~Chen and X.~Ran, ``Deep learning with edge computing: A review,''
  \emph{Proceedings of the IEEE}, vol. 107, no.~8, pp. 1655--1674, 2019.

\bibitem{NVIDIATensorRT2024}
{NVIDIA Corporation}, ``Nvidia tensorrt developer guide,''
  \url{https://docs.nvidia.com/deeplearning/tensorrt/developer-guide/index.html},
  2024, accessed: 2024-05-13.

\bibitem{stepanenko2019using}
S.~Stepanenko and P.~Yakimov, ``Using high-performance deep learning platform
  to accelerate object detection,'' in \emph{Proceedings of the International
  Conference on Information Technology and Nanotechnology, Samara, Russia},
  2019, pp. 26--29.

\bibitem{onnxruntime}
O.~R. developers, ``Onnx runtime,'' \url{https://onnxruntime.ai/}, 2021,
  version: x.y.z.

\bibitem{heller2022marine}
D.~Heller, M.~Rizk, R.~Douguet, A.~Baghdadi, and J.-P. Diguet, ``Marine objects
  detection using deep learning on embedded edge devices,'' in \emph{2022 IEEE
  International Workshop on Rapid System Prototyping (RSP)}.\hskip 1em plus
  0.5em minus 0.4em\relax IEEE, 2022, pp. 1--7.

\bibitem{panero2021real}
R.~Panero~Martinez, I.~Schiopu, B.~Cornelis, and A.~Munteanu, ``Real-time
  instance segmentation of traffic videos for embedded devices,''
  \emph{Sensors}, vol.~21, no.~1, p. 275, 2021.

\bibitem{OpenStreetMap}
{OpenStreetMap contributors}, ``{Planet dump retrieved from
  https://planet.osm.org },'' \url{ https://www.openstreetmap.org }, 2017.

\bibitem{Bremerhaven2024}
``Port information guide bremerhaven,''
  \url{http://www.hbh.bremen.de/sixcms/media.php/13/PORT-INFORMATION-GUIDE-Bremerhaven.pdf},
  Harbour Master Port of Bremerhaven.

\bibitem{labelme2016}
W.~Kentaro, ``{labelme: Image Polygonal Annotation with Python},''
  \url{https://github.com/wkentaro/labelme}, 2016.

\bibitem{riveiro2018maritime}
M.~Riveiro, G.~Pallotta, and M.~Vespe, ``Maritime anomaly detection: A
  review,'' \emph{Wiley Interdisciplinary Reviews: Data Mining and Knowledge
  Discovery}, vol.~8, no.~5, p. e1266, 2018.

\bibitem{de2022partial}
F.~A.~C. de~Oliveira, A.~Niemi, A.~Garc{\'\i}a-Ortiz, and F.~S. Torres,
  ``Partial camera obstruction detection using single value image metrics and
  data augmentation,'' in \emph{2022 6th International Conference on System
  Reliability and Safety (ICSRS)}.\hskip 1em plus 0.5em minus 0.4em\relax IEEE,
  2022, pp. 292--299.

\bibitem{FraunhoferCML2021}
H.-C. Burmeister, P.~Grundmann, P.~Hohnrath, and A.~Ujkani, ``Increasing
  maritime situational awareness by augmented reality solutions,'' Fraunhofer
  Center for Maritime Logistics and Services CML, White paper, 2021, available
  online:
  \url{https://www.cml.fraunhofer.de/content/dam/cml/en/documents/Studien/Burmeister\%20Grundmann\%20Hohnrath\%20Ujkani\%20\%282021\%29_Increasing-Situational-Awareness-by-Augmented-Reality-Solutions_White\%20Paper.pdf}.

\bibitem{blender3D}
\BIBentryALTinterwordspacing
\emph{Blender - a 3D modelling and rendering package}, Blender Foundation,
  Stichting Blender Foundation, Amsterdam, 2018. [Online]. Available:
  \url{http://www.blender.org}
\BIBentrySTDinterwordspacing

\bibitem{sloss2004arm}
A.~Sloss, D.~Symes, and C.~Wright, \emph{ARM system developer's guide:
  designing and optimizing system software}.\hskip 1em plus 0.5em minus
  0.4em\relax Elsevier, 2004.

\bibitem{guo2022review}
T.~Guo, T.~Zhang, E.~Lim, M.~Lopez-Benitez, F.~Ma, and L.~Yu, ``A review of
  wavelet analysis and its applications: Challenges and opportunities,''
  \emph{IEEE Access}, vol.~10, pp. 58\,869--58\,903, 2022.

\bibitem{cotter_2020}
F.~Cotter, ``Uses of complex wavelets in deep convolutional neural networks,''
  Ph.D. dissertation, University of Cambridge, 2020.

\bibitem{selesnick2005dual}
I.~W. Selesnick, R.~G. Baraniuk, and N.~C. Kingsbury, ``The dual-tree complex
  wavelet transform,'' \emph{IEEE signal processing magazine}, vol.~22, no.~6,
  pp. 123--151, 2005.

\bibitem{andreux2020kymatio}
M.~Andreux, T.~Angles, G.~Exarchakis, R.~Leonarduzzi, G.~Rochette, L.~Thiry,
  J.~Zarka, S.~Mallat, J.~And{\'e}n, E.~Belilovsky \emph{et~al.}, ``Kymatio:
  Scattering transforms in python,'' \emph{Journal of Machine Learning
  Research}, vol.~21, no.~60, pp. 1--6, 2020.

\bibitem{zhu2021tph}
X.~Zhu, S.~Lyu, X.~Wang, and Q.~Zhao, ``Tph-yolov5: Improved yolov5 based on
  transformer prediction head for object detection on drone-captured
  scenarios,'' in \emph{Proceedings of the IEEE/CVF international conference on
  computer vision}, 2021, pp. 2778--2788.

\bibitem{hartley2003multiple}
R.~Hartley and A.~Zisserman, \emph{Multiple view geometry in computer
  vision}.\hskip 1em plus 0.5em minus 0.4em\relax Cambridge university press,
  2003.

\bibitem{xie2017integration}
Y.~Xie, M.~Wang, X.~Liu, and Y.~Wu, ``Integration of gis and moving objects in
  surveillance video,'' \emph{ISPRS International Journal of Geo-Information},
  vol.~6, no.~4, p.~94, 2017.

\bibitem{shao2020accurate}
Z.~Shao, C.~Li, D.~Li, O.~Altan, L.~Zhang, and L.~Ding, ``An accurate matching
  method for projecting vector data into surveillance video to monitor and
  protect cultivated land,'' \emph{ISPRS International Journal of
  Geo-Information}, vol.~9, no.~7, p. 448, 2020.

\bibitem{brox2004high}
T.~Brox, A.~Bruhn, N.~Papenberg, and J.~Weickert, ``High accuracy optical flow
  estimation based on a theory for warping,'' in \emph{Computer Vision-ECCV
  2004: 8th European Conference on Computer Vision, Prague, Czech Republic, May
  11-14, 2004. Proceedings, Part IV 8}.\hskip 1em plus 0.5em minus 0.4em\relax
  Springer, 2004, pp. 25--36.

\bibitem{bell2007nasa}
D.~G. Bell, F.~Kuehnel, C.~Maxwell, R.~Kim, K.~Kasraie, T.~Gaskins, P.~Hogan,
  and J.~Coughlan, ``Nasa world wind: Opensource gis for mission operations,''
  in \emph{2007 IEEE Aerospace Conference}.\hskip 1em plus 0.5em minus
  0.4em\relax IEEE, 2007, pp. 1--9.

\end{thebibliography}

\end{document}